\documentclass[journal]{IEEEtran}
\usepackage{amsfonts}
\usepackage{graphicx}
\usepackage{subfigure}
\usepackage{multirow}
\usepackage{amssymb}
\usepackage{amsmath}
\usepackage{diagbox}
\usepackage{pifont}
\usepackage{makecell}
\usepackage{xcolor}
\usepackage{float}  
\usepackage[citecolor=blue, colorlinks]{hyperref}
\usepackage{subfiles}
\usepackage{booktabs}
\usepackage{soul}
\newcommand{\wltrev}[1]{\textcolor{black}{#1}} 
\newcommand{\wltrevRe}[1]{\textcolor{black}{#1}} 
\newcommand{\wltrevReRe}[1]{\textcolor{black}{#1}}  

\begin{document}

\title{Unsupervised 3D Point Cloud Completion via Multi-view Adversarial Learning}

\author{Lintai Wu, Xianjing Cheng, Yong Xu,~\emph{Senior Member, IEEE}, Huanqiang Zeng,~\emph{Senior Member, IEEE},\\ and Junhui Hou,~\emph{Senior Member, IEEE}
\thanks{This project was supported in part by the NSFC Excellent Young Scientists Fund 62422118, in part by the Hong Kong Research Grants Council under Grant 11219422, Grant 11202320, and Grant 11218121, and in part by the Shenzhen Science and Technology Program under Grant KJZD20230923114600002. \textit{(Corresponding author: Junhui Hou, Yong Xu)}}
\thanks{L. Wu is with the Bio-Computing Research Center, Harbin Institute of Technology, Shenzhen, Shenzhen 518055, Guangdong, China, and also with the Department of Computer Science, City University of Hong Kong, Hong Kong SAR. Email: lintaiwu2-c@my.cityu.edu.hk.}

\thanks{X. Cheng is with the School of Computer Science and Technology, Harbin Institute of Technology, Shenzhen, Shenzhen 518055, Guangdong, China. Email: chengxianjing@hit.edu.cn.}

\thanks{J. Hou is with the Department of Computer Science, City University of Hong Kong, Hong Kong SAR. Email: jh.hou@cityu.edu.hk.}

\thanks{Y. Xu is with the Bio-Computing Research Center, Harbin Institute of Technology, Shenzhen, Shenzhen 518055, Guangdong, China, and also with the Shenzhen Key Laboratory of Visual Object Detection and Recognition, Shenzhen 518055, Guangdong, China. Email:  laterfall@hit.edu.cn.}
\thanks{H. Zeng is with the School of Engineering, Huaqiao University, Quanzhou 362021, China, and also with the School of Information Science and Engineering, Huaqiao University, Xiamen 361021, China (e-mail: zeng0043@hqu.edu.cn).}}

\markboth{Manuscript Accepted to IEEE TVCG}%
{Shell \MakeLowercase{\textit{et al.}}: Bare Demo of IEEEtran.cls for IEEE Journals}

\maketitle

\begin{abstract}
    In real-world scenarios, scanned point clouds are often incomplete due to occlusion issues. The tasks of self-supervised and weakly-supervised point cloud completion involve reconstructing missing regions of these incomplete objects without the supervision of complete ground truth. Current methods either rely on multiple views of partial observations for supervision or overlook the intrinsic geometric similarity that can be identified and utilized from the given partial point clouds. In this paper, we propose MAL-UPC, a framework that effectively leverages both region-level and category-specific geometric similarities to complete missing structures. Our MAL-UPC does not require any 3D complete supervision and only necessitates single-view partial observations in the training set. Specifically, we first introduce a Pattern Retrieval Network to retrieve similar position and curvature patterns between the partial input and the predicted shape, then leverage these similarities to densify and refine the reconstructed results. Additionally,  we render the reconstructed complete shape into multi-view depth maps and design an adversarial learning module to learn the geometry of the target shape from category-specific single-view depth images of the partial point clouds in the training set. To achieve anisotropic rendering, we design a density-aware radius estimation algorithm to improve the quality of the rendered images. \wltrevReRe{Our MAL-UPC outperforms current state-of-the-art self-supervised methods and even some unpaired approaches}.
    We will make the source code publicly available at \url{https://github.com/ltwu6/malspc}.
\end{abstract}


\IEEEpeerreviewmaketitle
\section{Introduction}

\IEEEPARstart{P}{oint} clouds, which represent instances as a collection of points in 3D space, have attracted increasing interest due to the growing popularity of 3D scanning devices such as LiDAR and depth cameras.
They offer a detailed representation of the geometric features and spatial structures of 3D objects and scenes, and are widely applied in various fields, such as autonomous driving \cite{zhang2024text2nerf,wang2022and,cao2022monoscene,li2023voxformer,cui2021deep}, motion analysis \cite{xu2019mo,park20203d}, and augmented reality \cite{han2020live,ren2023geoudf}. Accurate and complete shape representation plays a vital role in understanding 3D point clouds.
However, point clouds scanned in practical scenarios are often incomplete due to occlusion. This incompleteness severely impacts the performance of downstream tasks, such as 3D object detection \cite{zhang2023unleash, qian20223d, mao20233d} and scene understanding \cite{xia2023scpnet,hou2021exploring}. 

To solve this problem, point cloud completion methods have emerged. Point cloud completion aims to infer the missing regions of an object based on the partial observation. Most of the existing methods rely on a large number of complete point clouds as supervision \cite{pcn, pointr, chen2023anchorformer, snowflake, wen2020point, wang2024pointattn, kasten2024point, cai2024orthogonal,huang2020pf,wang2020cascaded}. However, in real-world scenes, clean and complete point clouds are really few since current devices struggle to directly capture such data. 

To overcome these limitations, some research has focused on unsupervised learning approaches for point cloud completion. Among these methods, some employ complete shapes unpaired with the partial input as priors to drive the completion process \cite{pcl2plc,cycle,ganinverse,lsls,cao2023kt,ma2023symmetric,liu2024cloudmix}, and some utilize 2D signals such as RGB images or silhouette maps from multiple viewpoints as supervision \cite{wu2023leveraging, aiello2022cross}. However, the former still requires complete point clouds to acquire shape priors, and the latter relies on multi-view images along with the projection parameters. Recently, self-supervised methods, which only require partial inputs for supervision, have been proposed. These methods estimate the missing parts of the point cloud by utilizing the multi-view consistency of partial scans and the robust learning capabilities of the masked models. However, they face some limitations. First, most methods rely on multi-view partial point clouds of each object as shape constraints or as training data to drive the shape learning process. Nevertheless, obtaining point clouds from multiple viewpoints is also impractical in real-world applications. Second, they fail to utilize the inherent local similarities and category-specific geometric properties of objects.

From our perspective, it is commonly observed that an object frequently exhibits similar or repetitive local structures. For instance, a table typically features four identical cylindrical legs. This observation implies that the existing and missing regions of a point cloud are likely to share similar local geometric patterns. Moreover, objects within the same category generally display consistent structures. For example, most airplanes feature two flat wings. Consequently, these inherent geometric similarities, specific to each category, can be utilized to guide the completion of missing regions.

In this paper, we propose MAL-UPC, a \textbf{M}ulti-view \textbf{A}dversarial \textbf{L}earning framework for \textbf{U}nsupervised \textbf{P}oint cloud \textbf{C}ompletion. One of the core components, the Pattern Retrieval Network (PRN),  reconstructs complete shapes by initially generating a coarse shape and subsequently encoding local geometric patterns from both the coarse shape and the partial input. These patterns manifest in two forms: position encodings, quantified as the average distance between each point and its nearest neighbors, and curvature encodings, defined by the variations in normals between each point and its closest neighbors. Subsequently, a cross-attention-based module is introduced to retrieve correlated local patterns for the coarse shape from the partial shape. These retrieved patterns are further utilized to guide the densification and refinement of the coarse shape, capitalizing on the region-level geometric similarities and repetitiveness inherent in the data.

In addition, we introduce a Multi-view Adversarial Network (MAN) to learn the shared geometry within each category. Objects within the same category commonly exhibit remarkably similar shape distributions because they share many typical geometric structures. This similarity suggests that missing regions of a point cloud are likely to resemble the geometric structure of other objects in the same category. To exploit this similarity, we employ an adversarial learning strategy to learn the distinctive shape characteristics from category-specific data. Specifically, since depth maps provide a more compact and regular shape representation and are easier to acquire than partial point clouds, we render the reconstructed complete point cloud into depth maps from multiple viewpoints to capture the view-specific structures of the shape. \wltrevRe{Meanwhile, we construct an image bank for each category using the depth maps from which the partial point clouds in the training set are derived. Finally, we utilize a 2D CNN-based network to discriminate the rendered depth maps and those from the image bank, thus refining our prediction based on category-specific geometric knowledge.} Additionally, to achieve anisotropic rendering, we propose a density-aware radius estimation algorithm to adaptively adjust the point radius from different viewpoints, thereby improving the quality of rendered images. 

Compared to previous methods, our MAL-UPC offers the following advantages: \wltrevReRe{First, it does not require multi-view partial point clouds for supervision or as additional data to drive the training process.  Instead, it can achieve remarkable performance by employing only single-view partial observations, i.e., the 3D (partial point cloud) and 2D (silhouette map) representations of the input for supervision, along with the category-specific single-view depth maps from which the partial point clouds in the training set are derived for adversarial learning.} Moreover, MAL-UPC effectively leverages the intrinsic geometry similarity from partial data to reconstruct missing structures. In addition, unlike self-supervised and weakly-supervised methods, our approach does not require any data preprocessing processes, such as random region removal, which is time-consuming and computationally intensive.

We conduct extensive experiments on both synthetic datasets and real-world scans. \wltrevReRe{Since there are no unsupervised 3D point cloud completion methods that only rely on single-view partial observations during training (i.e., our settings), except the self-supervised approaches, the main comparison baselines in this paper are the self-supervised methods. In addition, we also compare our method with some unpaired methods and supervised methods. The experimental results demonstrate that our model outperforms self-supervised methods and even some unpaired approaches. Although our model does not outperform the previous unpaired methods in all instances, it only requires incomplete training data. }

In summary, the main contributions of this paper are two-fold:
\begin{itemize}
    \item We propose a Pattern Retrieval Network that leverages the intrinsic similarity and repetitiveness of region-level local geometry to reconstruct complete shapes. This network effectively infers the missing structures by learning correlated geometric patterns, including position and curvature, from the partial shape. 

    \item We propose a Multi-view Adversarial Network that facilitates the completion of missing shapes by leveraging category-specific geometric similarities. To achieve anisotropic rendering and improve the quality of the rendered depth maps, we design a density-aware radius estimation algorithm.

\end{itemize}

The remainder of this paper is organized as follows. 
Section \ref{related_work} briefly reviews existing supervised and unsupervised techniques for point cloud completion and reconstruction. Section \ref{proposed_method} provides a detailed description of our proposed methods. Section \ref{experiments} discusses our experimental design, findings, and analysis. Section \ref{limit_future} presents the limitations of our approach and future work. Finally, Section \ref{conclusion} concludes this paper.

\begin{figure*}[htbp]	
	\centering
	\includegraphics[width=0.9\textwidth]{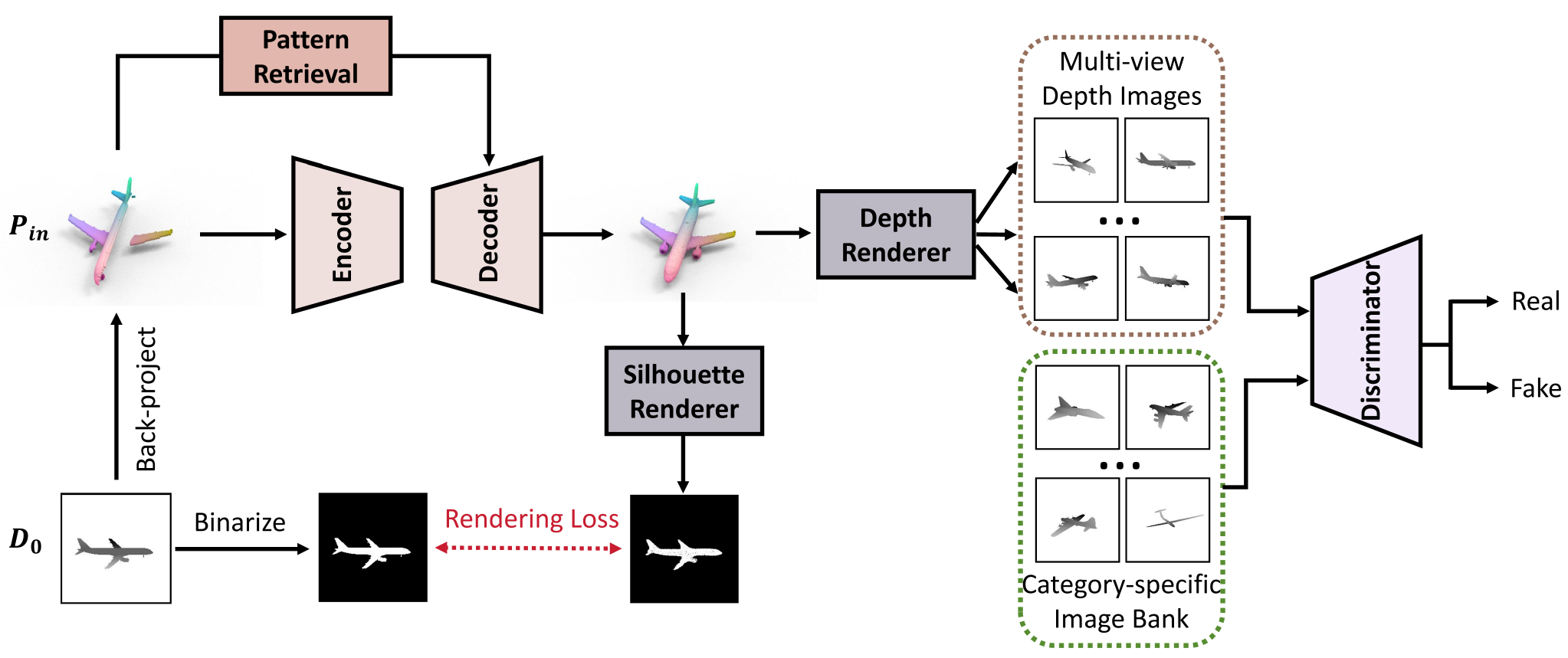}
	\caption{Flowchart of the proposed \textit{unsupervised} point cloud completion method. Given a depth map \(\mathbf{D}_0\), we can obtain an incomplete point cloud \(\mathbf{P}_{in}\) by back-projecting it into 3D space. We first employ an auto-encoder to generate a coarse shape, and then introduce a pattern retrieval module to guide the refinement and densification of the shape based on the position and curvature patterns of \(\mathbf{P}_{in}\). Then, we render the generated point cloud into a silhouette map from the input viewpoint and depth maps from various random viewpoints. \wltrevRe{The silhouette map is supervised by the mask image binarized from \(\mathbf{D}_0\), while the depth maps are used for adversarial learning with the category-specific single-view depth maps of the partial point clouds from the training set.}}
	\label{overall}
\end{figure*}

\section{Related Work}
\label{related_work}

\subsection{Supervised Point Cloud Completion}
The supervised point cloud completion methods utilize complete and clean point clouds paired with partial inputs to guide and drive the completion process. Based on the data representation in the encoding and decoding process, the point cloud completion methods can be classified into voxel-based and point-based approaches. 

The voxel-based methods \cite{3depn,grnet,sun2022patchrd,yan2022shapeformer,chibane2020implicit,mittal2022autosdf,wuwssc} first convert partial point clouds into 3D grids through voxelization and subsequently utilize the 3D Convolutional Neural Networks for learning the complete shapes. Dai et al. \cite{3depn} employed a 3D-Encoder Predictor Network to generate an initial approximation of a shape, followed by a 3D shape synthesis method that refines the coarse output using a shape prior derived from a shape database. Xie et al. \cite{grnet} introduced a Gridding module along with a reverse module for seamless conversion between point clouds and voxels, thereby minimizing structural information loss during the voxelization process. Additionally, they devised a Cubic Feature Sampling module to extract contextual features from neighboring points.
Sun et al. \cite{sun2022patchrd} represented point clouds as occupancy grids and leveraged local patches of partial inputs to infer missing parts by capitalizing on the potential repetition of local patterns within an object. Initially, they employed a network to capture similar local patches of the inputs. Then, these patches underwent deformation and blending processes, ultimately reconstructing into the final outputs. The Voxel-based methods can utilize the powerful feature extraction capabilities of CNNs. However, the computation speed of 3D CNN is relatively slow, and the conversion process between point clouds and voxels inevitably introduces accuracy loss.

The point-based methods \cite{pcn,foldingnet,topnet,li2023proxyformer,wang2022learning,zhang2020detail,pan2021variational} directly encode and decode features within the 3D point space by using the Multi-Layer Perceptron (MLP). To obtain smooth and detailed surfaces, some efforts employed the Folding operation, originally proposed in FoldingNet \cite{foldingnet}, to generate complete point clouds. Yuan et al. \cite{pcn} designed an MLP-based encoder to extract the global feature based on the partial input. Then, they leveraged fully connected (FC) layers to reconstruct a coarse point cloud. Subsequently, the Folding operation was applied to each coarse point. This approach effectively capitalizes on the inherent flexibility of FC layers and the desirable surface-smoothing characteristics offered by the Folding operation. Groueix et al. \cite{atlasnet} employed multiple MLPs to deform 2D grids into distinct local patches, enabling the generation of more intricate shapes. However, this approach resulted in overlapping regions between patches. To address this issue, Liu et al. \cite{msnpcc} proposed an expansion penalty method to minimize the overlap areas. Besides, it is difficult for a fixed 2D grid to capture the intricate topologies of point clouds. To improve the shape representation ability of the seeds, Yang et al. \cite{psgaupcl} designed a model for learning high-dimensional seeds for generating complete shapes. Likewise, Pang et al. \cite{tearingnet} utilized a network to learn the point cloud topology and tore the grid into patches to better represent the object's shape.

To achieve more flexible generation of missing points, some studies directly utilize MLPs to generate the 3D positions of points \cite{topnet,snowflake,vaswani2017attention,pointr,tang2022lake,zhou2022seedformer,wang2022learning,chen2023anchorformer,li2023proxyformer,zhu2023svdformer}. Inspired by the process of tree growth, Tchapmi et al. \cite{topnet} proposed a rooted-tree decoder that splits points layer by layer. Xiang et al. \cite{snowflake} introduced a coarse-to-fine approach where each parent point is divided into multiple child points. To drive the splitting process, they designed a skip-transformer block to capture the relationship between the features of child points and parent points. Drawing upon the remarkable capabilities of the Transformer \cite{vaswani2017attention} architecture in the fields of image processing and natural language processing (NLP), several recent advancements have introduced point cloud completion methods with Transformer as the backbone \cite{zhao2021point,guo2021pct}. Yu et al. \cite{pointr} first considered the process of inferring missing points based on the partial points as a translation task. They proposed a Transformer architecture for encoding the points' features and generating the missing parts of the inputs. Zhou et al. \cite{zhou2022seedformer} introduced Patch Seeds, which can preserve global structures and regional details to represent the shape. Based on the extracted seed features, they designed an Upsample Transformer to generate points. Chen et al. \cite{chen2023anchorformer} proposed to use a set of learned anchors to represent the regional information of objects. Then, they incorporated a modulation scheme and 2D grids to detail the sparse points. Li et al. \cite{li2023proxyformer} leveraged proxies to represent the existing and missing parts of the shapes and devised a missing part sensitive transformer to refine the missing proxies and thus generate high-quality missing structures. Zhu et al. \cite{zhu2023svdformer} first proposed a network that utilizes multi-view depth images to reconstruct a coarse shape. Then, they leveraged shape priors and geometric self-similarities to detail and refine the shapes.

In addition to partial point clouds, some methods incorporate view images, including RGB images and depth maps,  to assist in point cloud completion \cite{du2024cdpnet,vipc,zhu2023csdn,hu2019render4completion,hu20203d}. Zhang et al. \cite{vipc} leveraged a single-view RGB image to reconstruct a coarse point cloud and then devised an offset predictor model to refine it. Zhu et al. \cite{zhu2023csdn} predicted the coarse shape by transferring the image shape feature to the 3D point cloud domain and then proposed a dual-refinement module to refine the coarse shape. Hu et al. \cite{hu2019render4completion} rendered depth maps of a partial point cloud from different viewpoints and completed the shape by performing depth image completion for each view. To further constrain the geometric consistency among multi-view depth maps of a shape,  Hu et al. \cite{hu20203d} proposed a consistency loss to preserve the similarity of the shapes generated by multi-view depth maps. 

\subsection{Unsupervised Point Cloud Completion}
Due to the challenges in acquiring complete point clouds in real-world scenarios, some studies have proposed unsupervised methods for point cloud completion. Unsupervised point cloud completion methods do not require complete 3D ground truth or partial-complete pairs to drive the completion process. Existing unsupervised methods can be divided into three categories according to the supervision signal, i.e., unpaired methods, 2D-supervised methods, and self-supervised methods.

The unpaired point cloud completion methods generate complete point clouds without paired ground truth as supervision during the training process. Instead, they leverage complete and clean point clouds from other domains to capture 3D shape priors. Then, they used the learned shape priors to guide the completion of the partial point cloud from the source domain \cite{pcl2plc,cycle,ganinverse,lsls,cao2023kt,ma2023symmetric,liu2024cloudmix,wu2020multimodal,cui2022energy}. The first unpaired method is PCL2PCL, proposed by Chen et al. \cite{pcl2plc}. PCL2PCL first used partial point clouds from the source domain to pre-train a partial-shape reconstruction auto-encoder and then trained a complete-shape reconstruction auto-encoder by utilizing complete point clouds from another domain. Subsequently, they employed a GAN-based model \cite{gan_goodfellow} to transform the partial global features to the same distribution with the complete shapes' global features. To enhance the learning ability of the generation network, Wen et al. \cite{cycle} proposed a cycle-based model to learn the transformation between partial and complete point clouds. \wltrev{Cui et al. \cite{cui2022energy} proposed a latent-space energy-based model to learn a conditional distribution of the complete shape encoding given the partial encoding. Then, they designed a residual sampling strategy to select a valid encoding and decode it into a complete shape.} Cao et al. \cite{cao2023kt} proposed a knowledge transfer model to learn the shape information of partial inputs. They designed a teacher-assistant-student network to transfer the knowledge learned from complete shapes to partial shapes. In addition to the shape priors provided by complete point clouds, Ma et al. \cite{ma2023symmetric} proposed to leverage the symmetry of objects to reconstruct the missing regions. \wltrevReRe{The unpaired methods rely on additional complete point clouds from other domains to learn shape priors. By contrast, our MAL-UPC only requires single-view partial observations in the training set.}

The 2D-supervised methods employ 2D images along with projection parameters to guide the completion of 3D shapes. Wu et al. \cite{wu2023leveraging} projected reconstructed point clouds into the 2D plane and then used 2D points within the silhouette maps to supervise the view-specific shape. They also proposed a view-based calibrator to reduce noises in the predicted complete shapes. Aiello et al. \cite{aiello2022cross} enhanced the 3D point clouds' features with 2D RGB images by attention modules. They introduced a framework that rendered predicted point clouds from certain view angles and used 2D silhouette maps to supervise the rendered images. \wltrevReRe{However, the 2D-supervised methods utilize images from multiple viewpoints of an object to provide direct and powerful supervision. Thus, \cite{aiello2022cross} also refers to these methods as weakly-supervised point cloud completion methods. Besides, the existing 2D-supervised methods are also referred to as cross-modal point cloud completion methods because they leverage the features of RGB images to assist in point cloud completion. These configurations differ significantly from the settings of our method. Therefore, the 2D-supervised methods are not included in the comparison baselines of our experiments.}

The self-supervised methods only leverage the partial point cloud as a supervision signal to guide the completion process. Mittal et al. \cite{mittal2021self} proposed the first learning-based self-supervised method, PointPnCNet. PointPnCNet first randomly removed a small region from the input partial point cloud and used a network to predict the missing regions, including the manually removed parts and naturally missing parts. It employed the partial matching loss to preserve the existing parts in the predicted shapes. \wltrevRe{However, PointPnCNet relies on multiple partial point clouds for supervision, and thus it is also referred to as a weakly-supervised method. By contrast, our approach requires only single-view partial observations}. Inspired by closed-loop systems in control theory, Hong et al. \cite{hong2023acl} first predicted a coarse point cloud and used it to synthesize partial point clouds, and then input them to the same model as the initial input. By utilizing consistency shape loss among predictions generated by multiple partial inputs, the model could generate complete shapes. \wltrev{Nevertheless, this method requires the same training on the testing objects}. Cui et al. \cite{cui2023p2c} divided partial point clouds into three groups and then proposed a masked auto-encoder to reconstruct complete shapes. Rather than directly using partial Chamfer Distance loss, they designed a Region-aware Chamfer Distance loss to reduce the noises of the predictions. Kim et al. \cite{kim2023learning} proposed a pose-aware approach for self-supervised point cloud completion. They projected the shape features onto various viewpoints and reconstructed the corresponding incomplete point clouds. Then, they leveraged the shape consistency among multiple incomplete point clouds of the same object to constrain the reconstruction process.   

\section{Proposed Method}
\label{proposed_method}

Denote by \(\mathbf{D}_0 \in\mathbb{R}^{H\times W}\) the raw depth map scanned from an object, \(\mathbf{P}_{in}\in\mathbb{R}^{N\times3}\) the partial point cloud back-projected by \(\mathbf{D}_0\), \(\mathbf{P}_{com}\in\mathbb{R}^{M\times3}\) the complete shape of \(\mathbf{P}_{in}\), where \(H\) and \(W\) represent the height and width of the depth image, \(N\) and \(M\) are the numbers of points. Our goal is to find a mapping to make the predicted complete shape \(\mathbf{P}_{out}\) as close as possible to \(\mathbf{P}_{com}\) without \(\mathbf{P}_{com}\) as supervision.

As depicted in Figure \ref{overall}, our framework, namely MAL-UPC, first leverages a Pattern Retrieval Network to generate complete point clouds. Then, we render the complete prediction into a silhouette map and multi-view depth maps. \wltrevRe{The rendered depth maps are used to perform adversarial learning with the single-view depth maps of the partial point clouds from the training set.} In what follows, we will provide a detailed explanation of our approach.

\subsection{Pattern Retrieval Network}
\label{sec_prn}
\begin{figure}[htbp]	
	\centering
	\includegraphics[width=0.48\textwidth]{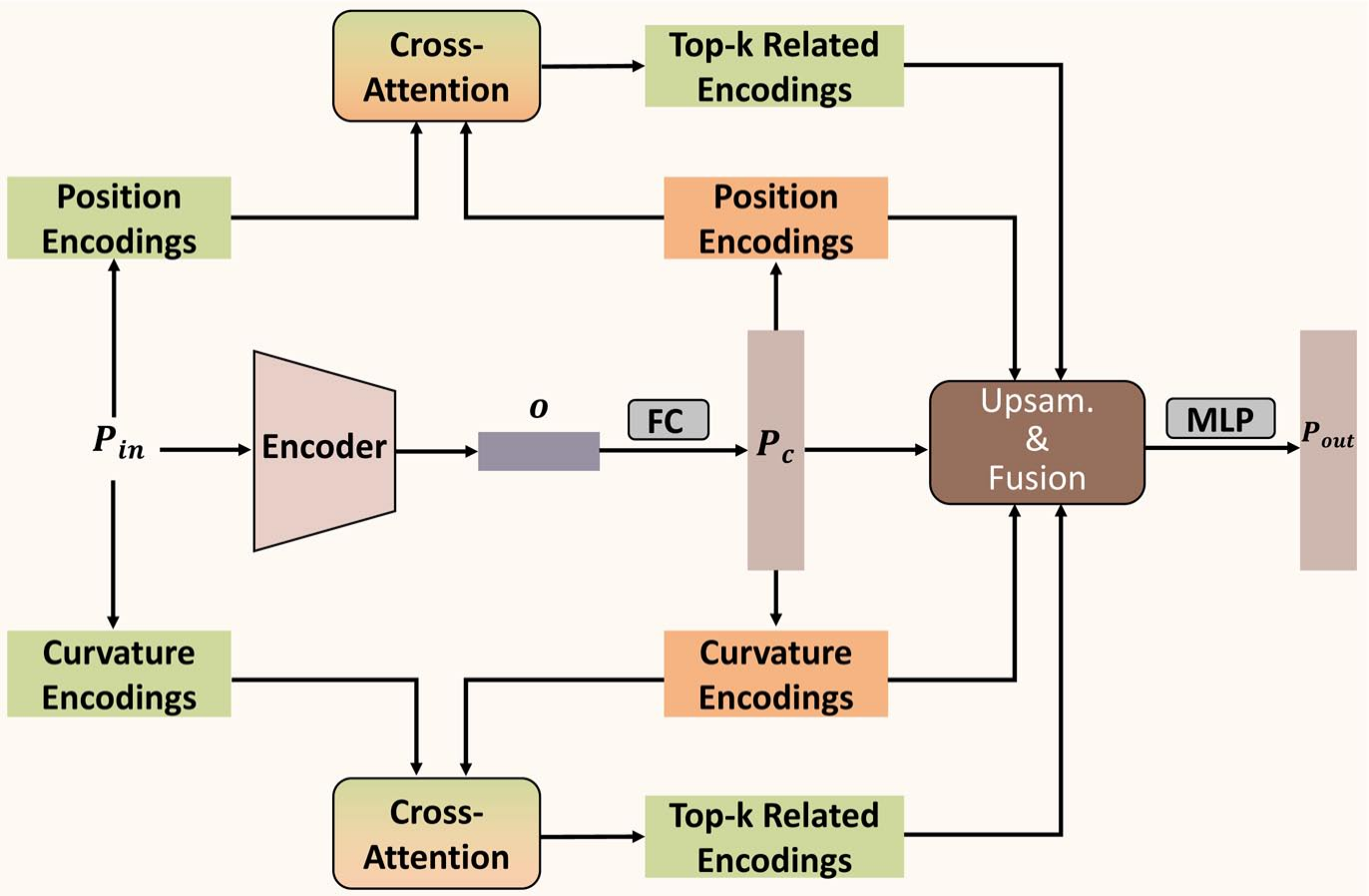}
	\caption{The flowchart of our Pattern Retrieval Network. ``FC" and ``MLP" denote the Fully-Connected layer and the Multi-Layer Perceptron layer. Given a partial point cloud \( \mathbf{P}_{in}\), we first utilize an auto-encoder to generate a coarse shape \( \mathbf{P}_c\). Next, we separately extract position encodings and curvature encodings for both \( \mathbf{P}_{in}\) and \( \mathbf{P}_c\), and then leverage a cross-attention-based module to identify and retrieve the most related encodings for each point in \( \mathbf{P}_c\) from \( \mathbf{P}_{in}\). Subsequently, \( \mathbf{P}_c\) and the selected encodings are upsampled and fused to refine the shape.}
	\label{prnet}
\end{figure}

The Pattern Retrieval Network leverages objects' region-level geometric similarity to infer the missing structures.
As shown in Figure \ref{prnet}, consuming as input a partial point cloud \(\mathbf{P}_{in}\), we first use a 3D encoder derived from PCN \cite{pcn} to learn a global feature \(\mathbf{o}\) that encapsulates the object's shape information. Then we employ three fully-connected layers to reconstruct a coarse yet complete point cloud \(\mathbf{P}_{c}\in\mathbb{R}^{N\times3}\).
Next, we extract the position encodings and curvature encodings from the partial input and coarse output, respectively. The position encodings represent the positional relationship of local points, which are formulated as the average distance between each point and its nearest neighbors. Specifically, for each point \(\mathbf{p}_i\in\mathbf{P}_{in}\), we first find out its \(K\) nearest neighbors \(\mathcal{N}_K(\mathbf{p}_i)=\{\mathbf{q}_i^j\}_{j=1}^K \subset \mathbf{P}_{in}\) and then compute the average Euclidean distance between \(\mathbf{p}_i\) and \(\mathcal{N}_K(\mathbf{p}_i)\) to obtain its position encoding $\mathbf{f}_{pos}^{in}(i)$, i.e.,

\begin{equation}
	\mathbf{f}_{pos}^{in}(i) = \frac{1}{K} \sum_{j=1}^K ||\mathbf{p}_i-\mathbf{q}_i^j ||_2.
\end{equation}

The curvature encodings capture the curvature variations across the local surfaces of the point cloud. We employ the normals of individual points to characterize surface curvature. For each point $\mathbf{p}_i$, we first compute the covariance matrix $\mathbf{C}_i$ based on its \(K\) nearest neighbors \(\mathcal{N}_K(\mathbf{p}_i)\) determined previously. The covariance matrix captures the local surface geometry and is defined as follows:

\begin{equation}
    \mathbf{C}_i = \frac{1}{K} \sum_{\mathbf{q}_i^j \in \mathcal{N}_K(\mathbf{p}_i)} (\mathbf{q}_i^j - \hat{\mathbf{p}_i})(\mathbf{q}_i^j - \hat{\mathbf{p}}_i)^T, \hat{\mathbf{p}}_i = \frac{1}{K} \sum_{j=1}^K \mathbf{q}_i^j.
\end{equation}

The estimated normal vector $\mathbf{n}_i$ for each point $\mathbf{p}_i$ is obtained by computing the eigenvector corresponding to the smallest eigenvalue of the covariance matrix $\mathbf{C}_i$. 
Besides,  \cite{tombari2010unique} is employed to perform the sign disambiguation of the normals. 

 To simulate the curvature variations of the surface, inspired by \cite{cui2023p2c}, we calculated the cosine similarity between the normal vector of each point and its neighboring \(K\) points.

\begin{equation}
    v(i,j) = 1-\texttt{cos}(\mathbf{n}_i,\mathbf{n}_j) = 1-\frac{\mathbf{n}_i \cdot \mathbf{n}_j}{\texttt{max}(||\mathbf{n}_i|| \cdot ||\mathbf{n}_j||, \epsilon)},
\end{equation}
\begin{equation}
    \mathbf{f}_{cur}^{in}(i) = \sqrt{\frac{1}{K} \sum_{j=1}^K (v(i,j)-\frac{1}{K} \sum_{j=1}^K v(i,j))^2},
\end{equation}
where $\mathbf{f}_{cur}^{in}(i)$ is the curvature encoding of $\mathbf{p}_i$, and \(\epsilon\) is a small constant to prevent the denominator from being 0.

In this way, we obtain the position encoding \(\mathbf{f}_{pos}^{in}\in\mathbb{R}^{N\times1}\) and the curvature encoding \(\mathbf{f}_{cur}^{in}\in\mathbb{R}^{N\times1}\)  from the partial input.
Meanwhile, we compute the position encoding \(\mathbf{f}_{pos}^{c}\in\mathbb{R}^{N\times1}\) and the curvature encoding  \(\mathbf{f}_{cur}^{c}\in\mathbb{R}^{N\times1}\) from the coarse shape using the same method. Following this, we introduce a cross-attention-based pattern retrieval module to capture correlated local patterns for coarse predictions from partial inputs. Specifically, for the position encoding, we utilize MLPs to project \(\mathbf{f}_{pos}^{in}\) and \(\mathbf{f}_{pos}^{c}\) to get queries \(\mathbf{g}_{pos}^{c} \in \mathbb{R}^{N\times1}\) and keys \(\mathbf{h}_{pos}^{in} \in \mathbb{R}^{N\times1}\), respectively.
Then we employ the algorithm in cross-attention \cite{vaswani2017attention} to calculate the weight maps between \(\mathbf{h}_{pos}^{in} \) and \(\mathbf{g}_{pos}^{c}\), representing the correlation of the position encodings between each point in \(\mathbf{P}_{in}\) and \(\mathbf{P}_{c}\):

\begin{equation}
    \mathbf{W}_{pos} = \texttt{Softmax} (\mathbf{h}_{pos}^{in} \cdot \mathbf{g}_{pos}^{c}).
\end{equation}

Next, based on \(\mathbf{W}_{pos}\), we retrieve the \(L\) most correlated position encodings from \(\mathbf{f}_{pos}^{in}\) for each point in \(\mathbf{P}_{c}\), denoted as \(\mathbf{e}_{pos} \in \mathbb{R}^{N\times L}\). Meanwhile, the related curvature encodings \(\mathbf{e}_{cur} \in \mathbb{R}^{N\times L}\)  for each point in \(\mathbf{P}_{c}\) are computed in the same manner. Then we upsample \(\mathbf{P}_{c}\) along with \(\mathbf{e}_{pos}\), \(\mathbf{e}_{cur}\), \(\mathbf{f}_{pos}^{c}\), \(\mathbf{f}_{cur}^{c}\), \(\mathbf{o}\), and fuse them in a concatenation manner. Finally, we employ MLPs to infer the point-wised offsets for the upsampled \(\mathbf{P}_{c}\) and add them to obtain the final outputs \(\mathbf{P}_{out}\).

\begin{figure}[tbhp]
	\centering  
	\subfigbottomskip=1pt 
	\subfigure{
		\includegraphics[width=0.23\linewidth]{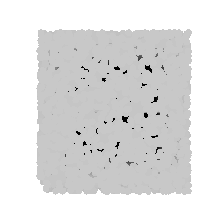}}
	\subfigure{
		\includegraphics[width=0.23\linewidth]{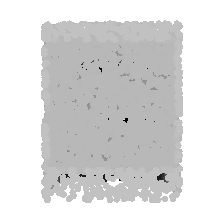}}
        \subfigure{
		\includegraphics[width=0.23\linewidth]{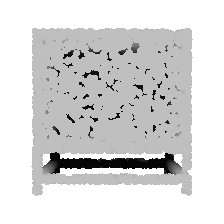}}
	\\
	\setcounter{subfigure}{0}
	\subfigure{
		\includegraphics[width=0.23\linewidth]{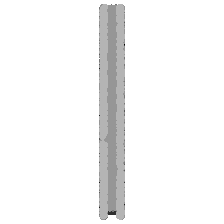}}
	\subfigure{
		\includegraphics[width=0.23\linewidth]{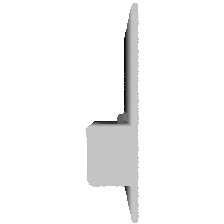}}
        \subfigure{
		\includegraphics[width=0.23\linewidth]{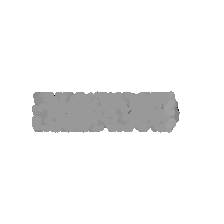}}
	\caption{\wltrevRe{Depth maps rendered from point clouds with an equal point number. The rendering process employs a fixed and identical point radius. With a small foreground area, the visible surface can be well represented. However, when the foreground area is large, a fixed point radius cannot cover the entire visible surface, resulting in some foreground pixels being covered by invisible points.}}
	\label{fixr_render}
\end{figure}

\subsection{Multi-view Adversarial Learning}
Objects within the same category share many similar structures, and therefore, their shape distributions are close. To make the generated complete point cloud as realistic and reasonable as possible, we develop an adversarial learning module to learn the category-specific shape knowledge from the real depth maps. 

\wltrevReRe{Assuming the training set includes $U$ categories, each category comprises $G_u$ single-view partial point clouds to be completed and $G_u$ corresponding depth maps since each partial point cloud is generated by back-projecting a raw depth map into 3D space, where $u=1,2,...,U$. For each category, we gather the raw depth maps of these partial point clouds to construct a category-specific depth image bank. In other words, the depth maps in the image bank are the projections of the incomplete point clouds solely from the perspectives in which they were captured, with no additional samples included.}

Next, we employ a silhouette renderer to render \(\mathbf{P}_{out}\) and \(\mathbf{P}_{c}\) into two silhouette maps \(\mathbf{S}_c\) and \(\mathbf{S}_{out}\) from the perspective in which \(\mathbf{D}_0\) is captured. \(\mathbf{S}_c\) and \(\mathbf{S}_{out}\) are supervised by a mask image binarized from \(\mathbf{D}_0\). Then we leverage a depth renderer to render \(\mathbf{P}_{out}\) into depth maps from different viewpoints. These depth maps and those from the category-specific depth image bank are fed to a discriminator, which consists of five 2D convolutional layers, to discriminate whether they are generated or real samples. In practice, for each predicted result, we render a depth map from a random viewpoint at each epoch, ensuring that the viewpoints vary across all epochs to achieve multi-view rendering while saving training time.

Our differentiable rendering pipeline is derived from \cite{p3ddocu}. Given the intrinsic and extrinsic camera parameters, \cite{p3ddocu} first transform the point cloud from world to camera coordinates and then project it into the 2D image plane. During rasterization, each point is rendered as a circular splat with a predefined radius in screen space, where the opacity of the splat decreases away from its center. The value of each pixel in the silhouette maps is computed by blending the information from the K-nearest points along the z-axis whose splat regions overlap that pixel. For depth maps, each pixel's value is the smallest z-coordinate of the K-nearest points.

However, \cite{p3ddocu} assigns a fixed radius to point splats under all viewpoints. In fact, the density of 2D points projected by a point cloud with a fixed point number varies across different viewpoints. For example, as shown in Figure \ref{fixr_render}, 
when there is a large foreground area in the rendered image, the density of the projected points is small (we call these low-density depth maps). In such cases, a small radius cannot cover all areas of the visible surface and thus many ``hole pixels" appear. These ``hole pixels" are inaccurately represented by the depth values of the points that should have been occluded. In contrast, in scenarios with few foreground pixels, the density of points is large (we call these high-density depth maps). In this case, a small radius suffices to cover the entire visible surface, and thus, the rendered depth map is more realistic and smooth. Therefore, a fixed point radius is inadequate for rendering under different viewpoints. Intuitively, using a larger point radius might produce smoother depth maps in both high point density and low point density scenarios. However, in this way, the projected points cover a larger area on the 2D plane, potentially extending the edges of the foreground pixels beyond the original boundaries of the object.

To address this issue, we introduce a density-aware radius estimation (DARE) algorithm to adaptively determine a suitable radius for points from each viewpoint. First, we predefine an initial point radius for the reconstructed shape and then render it into a coarse depth map. We then calculate the number of foreground pixels on this depth map, denoted as \(A\), and subsequently, the density of the projected points on the depth map is defined as 
\begin{equation}
\label{dare_d}
    d = M/A.
\end{equation}

Then, we model the relationship between the actual point radius and density with an inversely proportional function and set a coefficient \(\eta \) such that
\begin{equation}
\label{dare_r}
    r = \eta / d,
\end{equation}
where \(r\) denotes the point radius. 
Note that the parameter \(\eta \) can be pre-estimated based on the initial point radii. Specifically, for each category, the density of projected points could be approximated using the average foreground pixel number of all the depth maps in the dataset, as specified in Formula \ref{dare_d}. Following this, \(\eta \) can be computed according to Formula \ref{dare_r}. 
In this way, the radius of the point cloud dynamically adjusts according to the density of projected points, yielding a smoother depth map. 

\subsection{Loss Functions}
To preserve the existing regions in the reconstructed point cloud, we employ a partial matching loss \cite{cycle}, which computes the Unidirectional Chamfer Distance (UCD) from output to input:
\begin{equation}
\label{ucd_formula}
\begin{split}
	{\rm \texttt{UCD}}(\mathbf{Q}_1, \mathbf{Q}_2) = \frac{1}{|\mathbf{Q}_1|}\sum_{\mathbf{q}_1\in \mathbf{Q}_1} \mathop{\texttt{min}}\limits_{\mathbf{q}_2\in \mathbf{Q}_2}||\mathbf{q}_1-\mathbf{q}_2||,
\end{split}
\end{equation}

\begin{equation}
    L_{part} = {\rm \texttt{UCD}}(\mathbf{P}_{in}, \mathbf{P}_c)+{\rm \texttt{UCD}}(\mathbf{P}_{in},\mathbf{P}_{out}).
\end{equation}

Furthermore, we use a rendering loss to supervise the predicted point cloud in 2D space. First, we binarize the depth image \(\mathbf{D}_0\) to obtain the 2D silhouette ground truth \(\mathbf{S}_{0}\). Subsequently, we compute the Mean Squared Error (MSE) loss between \(\mathbf{S}_0\) and \(\mathbf{S}_{out}\). Note that due to the sparsity of the coarse point cloud \(\mathbf{P}_c\), the foreground pixels of \(\mathbf{S}_c\) are also sparse and discrete, whereas the foreground of \(\mathbf{S}_0\) is continuous and dense. To solve this problem, we propose a Masked MSE loss, which only calculates the MSE loss on the foreground pixels of the rendered image. This trick confines the projected points of the rendered image to be accurately positioned.

\begin{equation}
\begin{split}
    L_{rend} &= \frac{1}{H\cdot W} \sum_{i=1}^H \sum_{j=1}^W (\mathbf{S}_0(i,j)-\mathbf{S}_{out}(i,j))^2  \\ 
    &+\frac{1}{H\cdot W} \sum_{i=1}^H \sum_{j=1}^W (\mathbf{S}_0(i,j)-\mathbf{S}_{c}(i,j))^2\cdot \mathbf{B}(i,j),
\end{split}
\end{equation}
where \(\mathbf{B}\) is the binary mask for \(\mathbf{S}_{c}\) that indicates whether each pixel of \(\mathbf{S}_{c}\) is foreground or not.

To generate a more uniform point cloud, following \cite{ma2023symmetric}, we incorporate a density loss \(L_{dens}\) to constrain the densities of all local regions in the point cloud to be as close as possible. 

In addition, the optimization of our generation model is also driven by the adversarial loss proposed by \cite{mao2017least}, which is defined as:

\begin{equation}
    L_{gen} = \mathbb{E}_{d_r\sim D_r}[D(d_r)-1]^2
\end{equation}

\begin{table*}[htbp]
	\centering
	\renewcommand\arraystretch{1.25}
	\caption{\wltrevRe{Results on our synthetic dataset in terms of L2 CD\(\downarrow\) (scaled by \(10^{4}\)). ``Ours-U", ``Ours-M" and ``Ours-S" denote our model trained in an unpaired manner,  trained with eight categories together and trained for each category, respectively. ``P2C-X" is P2C augmented with our rendering loss and MAN. ``\ding{51}" and ``\ding{55}" indicate whether the corresponding supervision was used or not. ``PC" is ``point cloud". The \textbf{bold} numbers represent the best results between our method and self-supervised models.}}
	\label{synthetic_results}
	\begin{tabular}{l|c|c|c|c|c|c|c|c|c|c|c}
		\toprule[1.2pt]
		\multirow{2}*{Methods}             & \multicolumn{2}{c|}{Supervision} & \multirow{2}*{Average} & \multirow{2}*{Plane} & \multirow{2}*{Cabinet} & \multirow{2}*{Car} & \multirow{2}*{Chair} & \multirow{2}*{Lamp}	& \multirow{2}*{Couch}	& \multirow{2}*{Table}	& \multirow{2}*{Watercraft} \\ \cline{2-3}
            ~ & \makebox[0.075\textwidth][c]{Paired} & \makebox[0.075\textwidth][c]{Complete PC} & ~ & ~ & ~  & ~  & ~  & ~  & ~  & ~  & ~   \\ \hline 
		FoldingNet \cite{foldingnet}&  \ding{51} &  \ding{51}  & 6.41 & 1.86  & 9.13  & 3.19 & 5.72 &  12.72	& 5.42	& 7.29 	& 6.00    \\ 
		PCN \cite{pcn}      & \ding{51} &  \ding{51} & 5.42  & 2.22  & 5.32  &  3.33    &  7.25   & 8.44 & 5.91 &  6.48	&  4.43  \\
		TopNet \cite{topnet}      & \ding{51} &  \ding{51} & 4.49  & 1.65 & 4.34 & 3.26 &  5.25 & 7.97 & 5.41 & 4.49	& 3.56 \\ 
		AnchorFormer \cite{chen2023anchorformer}      & \ding{51} &  \ding{51} & 3.57  & 1.06  & 4.80 & 2.51 & 4.23 & 4.22 	& 4.44	& 4.64 & 2.68     \\ \hline 
            PCL2PCL \cite{pcl2plc}   & \ding{55} &  \ding{51} & 16.53 & 4.68 & 28.08 & 6.80 &  17.01  & 20.58  & 14.56 & 28.88 & 11.71  \\
		KTNet \cite{cao2023kt}    & \ding{55} &  \ding{51} & 6.62 & 1.41  & 5.58 & 4.51 & 8.95 & 8.65 & 8.77 & 9.86 & 5.29 \\
            LTUPCN \cite{cui2022energy} & \ding{55} &  \ding{51} & 7.35 & 1.67 & 5.78 & 4.21 & 10.35 & 12.55 & 8.84 & 10.49 & 4.96 \\
            USSPA \cite{ma2023symmetric}    & \ding{55} &  \ding{51} & 14.08 &  4.01 & 6.94 &6.81& 27.36 &  25.39 & 12.13 & 19.31 & 10.70 \\
            Ours-U  & \ding{55} &  \ding{51} & 10.20	& 2.58  & 8.65 & 4.49 & 12.74 & 19.70 & 9.61 & 16.13 & 7.71  \\ \hline 
             ACL-SPC \cite{hong2023acl}   & \ding{55} &  \ding{55} & 15.26 & 3.38 & 21.72 & 5.87 &  11.38  & 34.73 & 13.03 & 20.24  & 11.73 \\
		P2C \cite{cui2023p2c}    & \ding{55} &  \ding{55} & 12.28 & \textbf{1.85} & 12.73 & \textbf{4.93} & 12.99 & 19.43 & 24.82 & 14.95 & 6.57 \\  
            P2C-X      & \ding{55} &  \ding{55} & 11.08  & 2.78  & 11.83  & 6.07  &  10.40 &  18.10 & 19.48  & 13.48  &  \textbf{6.53} \\ \hline
            Ours-M          & \ding{55} &  \ding{55}	& 11.47	& 4.08 & 15.96  & 7.41 & 11.86 & 21.25 & \textbf{9.51}  & 13.49  & 8.22  \\
		Ours-S          & \ding{55} &  \ding{55}	& \textbf{9.08}	& 2.35 & \textbf{10.64}  & 5.92 & \textbf{10.14} & \textbf{14.51} & 10.15  & \textbf{11.65}  & 7.28  \\ 
        \bottomrule[1.2pt]
	\end{tabular}
\end{table*}

\begin{table*}[htbp]
	\centering
	\renewcommand\arraystretch{1.25}
	\caption{\wltrev{Results on our synthetic dataset in terms of Precision\(\downarrow\) (left) and Coverage\(\downarrow\) (right). The \textbf{bold} numbers represent the best results.}}
	\label{pre_cov}
	\begin{tabular}{l|c|c|c|c|c|c|c|c|c}
		\toprule[1.2pt]
		Methods             & Average & Plane & Cabinet & Car & Chair & Lamp	& Couch	& Table	& Watercraft \\ \hline 
             ACL-SPC \cite{hong2023acl}   &13.43/\textbf{1.81} & 2.75/0.63 & 19.66/\textbf{2.06} & 4.05/1.82 & 8.88/\textbf{2.49} & 33.14/1.59 & 10.00/3.03 & 18.44/\textbf{1.79} & 10.59/\textbf{1.14} \\
		P2C \cite{cui2023p2c}   & 9.28/3.00 & \textbf{1.33/0.52} & 10.02/2.71 &\textbf{2.97}/1.96 &7.82/5.17 & 18.10/\textbf{1.33} & 17.05/7.77 & 11.97/2.98 &\textbf{5.01}/1.56 \\ \hline
		Ours          & \textbf{6.33}/2.74 & 1.43/0.92 & \textbf{8.08}/2.56 & 4.28/\textbf{1.63} & \textbf{6.27}/3.87 & \textbf{10.52}/3.99 & \textbf{7.71/2.44} & \textbf{7.23}/4.42 & 5.17/2.10 \\ \bottomrule[1.2pt]
	\end{tabular}
\end{table*}

Overall, the generation process of our model is optimized by the generation loss \(L_G \):

\begin{equation}
    L_G = \alpha_1\cdot L_{part} + \alpha_2\cdot L_{rend} + \alpha_3\cdot L_{dens} + \alpha_4\cdot L_{gen}, 
\end{equation}
where \(\alpha_1,\alpha_2,\alpha_3,\alpha_4\) are the weights for balancing these loss terms.
Meanwhile, the discriminator is optimized by \(L_D\):

\begin{equation}
    L_D = \mathbb{E}_{d_s\sim D_s}[D(d_s)-1]^2 + \mathbb{E}_{d_r\sim D_r}[D(d_r)]^2
\end{equation}

\section{Experiments} \label{experiments}

\subsection{Dataset and Evaluation Metric}
To comprehensively evaluate our method, we conduct experiments using both synthetic data and real-world data. \wltrev{Since existing datasets lack the initial depth maps for partial point clouds, we generate a new synthetic dataset from ShapeNet \cite{chang2015shapenet}.}
Specifically, we render a single depth map for each object and then back-project it into 3D space to obtain the partial point cloud. \wltrevRe{The depth map size is $224\times 224$}. \wltrevRe{To better align with human perspectives, we set the elevation angle for scanning shapes within the range of -30 to 30 degrees and randomly sample the azimuth angle between 0 and 360 degrees}. The complete point clouds for evaluation are uniformly sampled from the objects' surfaces. \wltrevRe{Following PCN \cite{pcn}, we select 30974 objects across 8 categories. For each category, we employ 100/150 objects for validation/testing, and the rest is used for training. Concretely, the object categories and their corresponding numbers of partial point clouds in the training set are as follows: airplane (3795), cabinet (1322), car (5677), chair (5750), lamp (2068), couch (2923), table (5750), watercraft (1689). Besides, we construct an image bank for each category using the raw depth maps from which the partial point clouds in the training set are derived. In total, there are 8 image banks, each containing the same number of depth maps as the partial point clouds in their corresponding category from the training set.}
We set the point numbers of partial inputs and complete shapes to 2048 and 16384, respectively. \wltrev{Besides, we evaluate our model on widely used benchmarks, including PCN \cite{pcn}, 3D-EPN \cite{3depn} and CRN \cite{wang2020cascaded}.}
The evaluation metric for synthetic data is CD. \wltrev{Additionally, following previous self-supervised approaches \cite{mittal2021self,hong2023acl}, we report each component of CD, i.e., Precision and Coverage.}

For the real-world dataset, we use the data sourced from ScanNet \cite{dai2017scannet} and KITTI \cite{geiger2012we}. ScanNet contains incomplete objects in indoor scenes, and we extracted 548 chairs and 551 tables from the ScanNet dataset as indoor test data.
KITTI provides partial scans of instances in outdoor scenes.
We extract 473 partial point clouds from car objects in the Velodyne data, each containing at least 100 points.  
Due to the lack of complete ground truth, following previous works \cite{pcn,pcl2plc,ganinverse, pointr, wu2023leveraging}, we adopt the UCD and Unidirectional Hausdorff Distance (UHD) as the evaluation metrics for assessing our model's performance on the real-world dataset. 
The UCD is introduced in Formula \ref{ucd_formula}, and the UHD is formulated as:
\begin{equation}
    \begin{split}
	{\rm \texttt{UHD}}(\mathbf{Q}_1, \mathbf{Q}_2) = \mathop{\texttt{max}}\limits_{q_1 \in \mathbf{Q}_1} \mathop{\texttt{min}}\limits_{q_2 \in \mathbf{Q}_2} \| q_1 - q_2 \|_2
    \end{split}    
\end{equation}

\subsection{Implementation Details}
The point numbers of $\mathbf{P}_{in}$, $\mathbf{P}_c$, and $\mathbf{P}_{out}$ are 2048, 2048 and 16384, respectively. \wltrevRe{The height $H$ and width $W$ of all depth maps are 224.} We employ 16 and 24 neighbors for calculating position and curvature encodings, respectively. \(L\) is 16 and \(\epsilon\) is \(1e-8\). Initial point radii for the DARE algorithm are set to 0.012 for the plane category and 0.03 for other categories. 
The loss weights are all set to 1. The initial learning rate is 0.0001, and it decays by 0.5 for every 200 epochs. The training process takes 600 epochs to converge. The experiments are conducted on an NVIDIA RTX 3090 GPU with Intel(R) Xeon(R) CPUs.
\subsection{Comparison with State-of-the-Art Methods}

\subsubsection{Results on synthetic data}
\label{section_comparison}
We first conduct experiments on the synthetic dataset and compare the results with state-of-the-art methods, including fully supervised methods (i.e., FoldingNet \cite{foldingnet}, PCN \cite{pcn}, TopNet \cite{topnet}, AnchorFormer \cite{chen2023anchorformer}), unpaired methods (i.e., PCL2PCL \cite{pcl2plc}, KTNet \cite{cao2023kt}, LTUPCN \cite{cui2022energy}, USSPA \cite{ma2023symmetric}) and self-supervised methods (i.e., ACL-SPC \cite{hong2023acl}, P2C \cite{cui2023p2c}). \wltrevReRe{Note that although our MAL-UPC leverages category-specific geometric knowledge to infer complete shapes and is therefore classified as an unsupervised method, we believe it is rational to compare our method against self-supervised approaches based on the following facts. First, there are no unsupervised 3D point cloud completion methods that only rely on single-view partial observations during training (i.e., our settings), except ACL-SPC, P2C, and \cite{kim2023learning}. However, the codes of \cite{kim2023learning} have not been made public yet. Regarding the weakly-supervised methods, such as \cite{mittal2021self} and \cite{gu2020weakly}, they use multi-view partial point clouds from complete objects for supervision. Since combining multi-view partial point clouds can nearly reconstruct a complete object \cite{hong2023acl}, this supervision signal is direct and powerful, which significantly differs from our settings. Therefore, it is the most appropriate choice to compare our method with ACL-SPC and P2C. Second, it is common that common/similar geometric structures or objects exist in real-world training datasets. Thus, we propose a Multi-view Adversarial Network (MAN) to fully leverage this prior within each category, which is just our innovation. Unfortunately, ACL-SPC and P2C fail to explore this prior information. Moreover, this prior information is provided by the category-specific single-view depth maps which are 2D representations of the incomplete point clouds present in the training set. In other words, an input partial point cloud learns shape priors from depth maps of other partial point clouds to be completed in the training set, and these depth maps are from the same category as this input partial point cloud. Therefore, we do not employ any additional information beyond the training data. Finally, in addition to the self-supervised approaches, we have compared our method with other state-of-the-art unsupervised methods, i.e., unpaired methods, such as PCL2PCL, KTNet, LTUPCN and USSPA. Actually, these unsupervised methods under comparison require unpaired complete point clouds for training, and this kind of prior is stronger than that of our method.}

\wltrevReRe{Additionally, to provide a more comprehensive comparison, we introduce a new approach that incorporates our rendering loss and MAN into the P2C model. The weights for the rendering loss and adversarial loss are set to 1 and 0.1, respectively, while the original loss terms and weights of the P2C model remain unchanged. This experiment bridges the category-specific prior gap between the self-supervised method and our approach.} \wltrevRe{Regarding our method, we present three models trained in different manners. The first model, denoted as Ours-S, involves training our MAL-UPC framework separately for each category. The second model, referred to as Ours-M, entails training MAL-UPC on all eight categories together while each input employs its category's depth image bank for adversarial learning. The third model, Ours-U, is trained separately for each category in an unpaired manner. Specifically, the MAN module in our model is replaced with the MLP-based discriminator from USSPA, while all other components remain unchanged. For a fair comparison, the inputs to the discriminator in our unpaired model and all unpaired baselines are the predicted point cloud and a complete point cloud from our dataset that is unpaired with the partial input.}
Note that all compared methods are only trained separately for each category.

\begin{table*}[htbp]
	\centering
	\renewcommand\arraystretch{1.25}
	\caption{\wltrev{Results on PCN, 3D-EPN and CRN dataset in terms of L2 CD\(\downarrow\) (scaled by \(10^{4}\)). The \textbf{bold} numbers represent the best results.}}
	\label{other_results}
	\begin{tabular}{l|c|c|c|c|c|c|c|c|c|c}
		\toprule[1.2pt]
		Datasets & Methods & Average & Plane & Cabinet & Car & Chair & Lamp	& Couch	& Table	& Watercraft \\ \hline 
             \multirow{3}*{PCN} & ACL-SPC \cite{hong2023acl}   & 15.81 & 4.29 & 22.76 & 5.80 & 12.06 & 34.21 & 14.13 & 21.52 & 11.75 \\
		~ & P2C \cite{cui2023p2c}    & 17.14 & \textbf{2.18} & 20.64 & \textbf{4.77} & 13.76 & 19.06 & 55.18 & 14.54 & \textbf{7.00} \\ 
		~ & Ours          & \textbf{9.77} & 3.38 & \textbf{11.20} & 6.32 & \textbf{10.20} & \textbf{14.52} & \textbf{14.12} & \textbf{10.71} & 7.72\\ \hline
            \multirow{3}*{3D-EPN} & ACL-SPC \cite{hong2023acl}   & 20.75 & 4.57 & 29.34 & 9.00 & 17.83 & 43.79 & \textbf{17.28} & 30.61 & 13.63 \\
		~ & P2C \cite{cui2023p2c}    & \textbf{14.54} & \textbf{2.58} & \textbf{19.35} & \textbf{8.00} & \textbf{13.39} & \textbf{22.67} & 22.11 & \textbf{19.76} & \textbf{8.51} \\ 
		~ & Ours          & 18.69 & 3.26 & 30.23 & 10.35 & 24.00 & 24.76 & 18.18 & 28.90 & 9.86 \\ \hline
            \multirow{3}*{CRN} & ACL-SPC \cite{hong2023acl}   & 16.81 & 4.56 &  24.30 &  7.03 & 13.11 & 34.71 & 15.53 & 22.86 & 12.40 \\
		~ & P2C \cite{cui2023p2c}    & 18.18 & \textbf{2.47} & 22.54 &  \textbf{5.91} & 14.88 & 19.70 & 56.85 & 15.48 & \textbf{7.63} \\ 
		~ & Ours          & \textbf{10.72} & 3.70 & \textbf{12.80} & 7.42 & \textbf{11.10} & \textbf{15.20} & \textbf{15.45} & \textbf{11.72} & 8.37 \\ 
  
  \bottomrule[1.2pt]
	\end{tabular}
\end{table*}

\begin{table*}[htbp]
    \centering
    \renewcommand\arraystretch{1.25}
    \caption{Results on real-world dataset in terms of UCD\(\downarrow\) (scaled by \(10^{4}\)) and UHD\(\downarrow\) (scaled by \(10^{2}\)). The \textbf{bold} numbers represent the best results between our method and self-supervised models.}
    \label{real_table}
    \begin{tabular}{c|cc|cc|cc|cc|cc}
        \toprule[1.2pt]
        Methods & \multicolumn{2}{c|}{PCN \cite{pcn}}& \multicolumn{2}{c|}{PCL2PCL \cite{pcl2plc}} & \multicolumn{2}{c|}{ACL-SPC \cite{hong2023acl}} & \multicolumn{2}{c|}{P2C \cite{cui2023p2c}} & \multicolumn{2}{c}{Ours} \\
        \hline
        Metrics & UCD & UHD & UCD & UHD & UCD & UHD & UCD & UHD & UCD & UHD \\
        \hline
        chair & 5.49 & 9.19  & 9.65 & 10.07  & 7.74 & 11.56   & \textbf{7.26} & 12.21 &8.06 &  \textbf{9.79} \\
        car & 2.43 & 5.22 & 3.62 & 5.57  & 1.54 & 5.17 & 1.76 & 5.02 & \textbf{1.32} & \textbf{4.17} \\
        table & 11.07 & 10.80 & 19.95 & 13.80 & 16.55 & 14.29 & 9.47 & 11.03 & \textbf{8.77} & \textbf{10.63} \\
        \bottomrule[1.2pt]
    \end{tabular}
\end{table*}

The quantitative results presented in Table \ref{synthetic_results} suggest that our method achieves an impressive average CD of 9.08, which outperforms self-supervised methods and two unpaired approaches (i.e., PCL2PCL and USSPA)  by a large margin, and even demonstrates competitive results compared to some fully supervised methods. Regarding the category-specific results, our method significantly outperforms two self-supervised methods across five categories. Specifically, the CD value of MAL-UPC under the lamp category surpasses the results of P2C and ACL-SPC by nearly 5 and 20, respectively. Besides, the CD values of MAL-UPC are lower than those of self-supervised benchmarks by 2-4 under cabinet, couch, and table categories. Furthermore, even with the prior knowledge of complete 3D point clouds, the results of the two unpaired methods, PCL2PCL and USSPA, lag behind our method across multiple categories. In addition, compared to the supervised method FoldingNet, our model achieves a CD difference of less than 3.

\begin{figure*}[htbp]
	\centering  
	\subfigbottomskip=1pt 
	\subfigure{
		\includegraphics[width=0.087\linewidth]{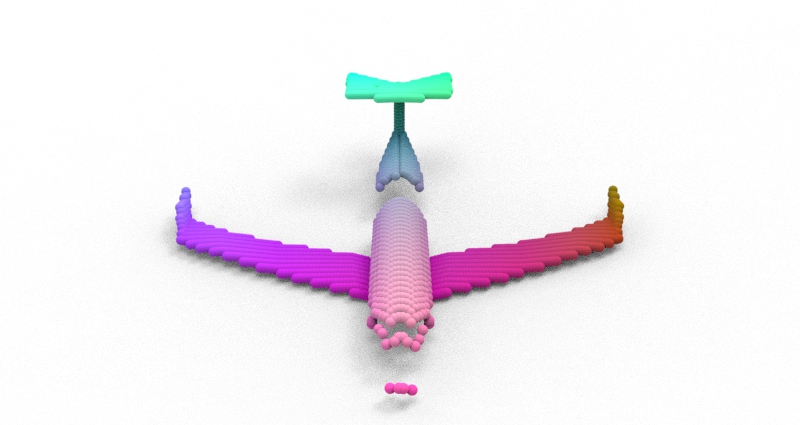}}
	\subfigure{
		\includegraphics[width=0.087\linewidth]{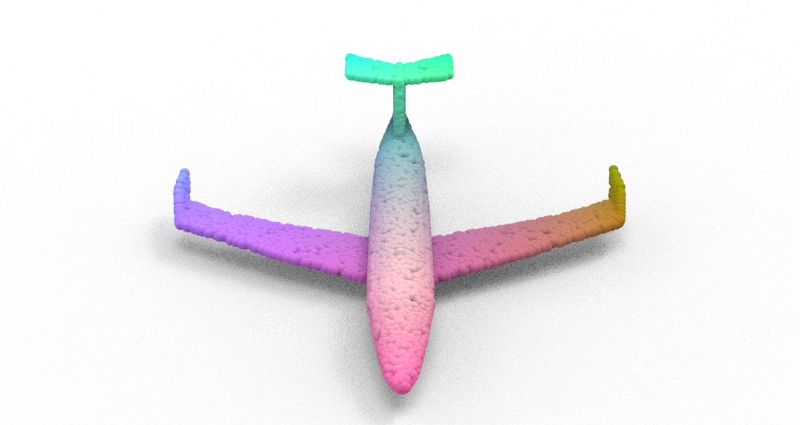}}
	\subfigure{
		\includegraphics[width=0.087\linewidth]{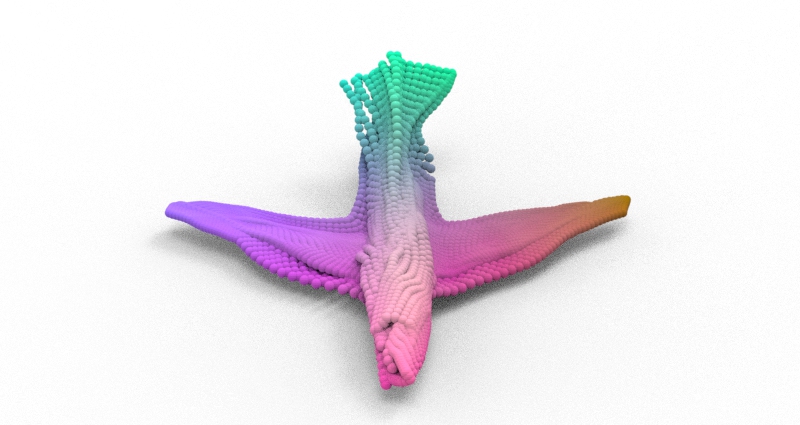}}
	\subfigure{
		\includegraphics[width=0.087\linewidth]{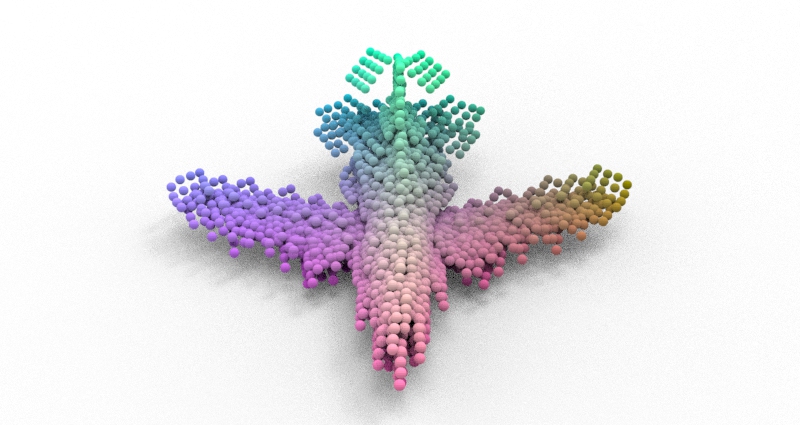}}
	\subfigure{
		\includegraphics[width=0.087\linewidth]{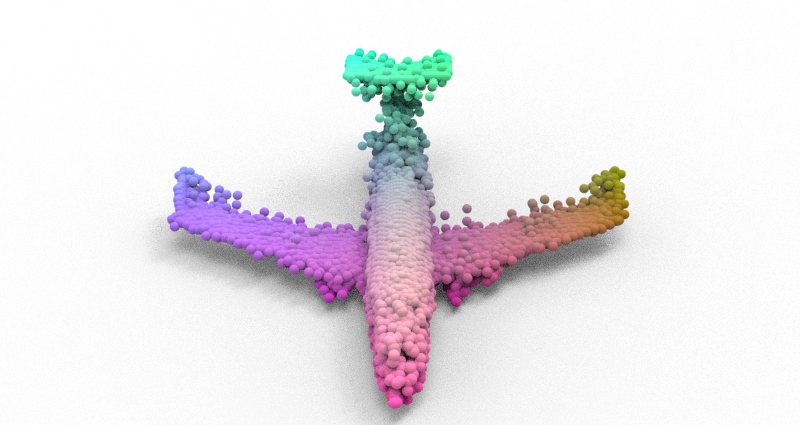}}
	\subfigure{
		\includegraphics[width=0.087\linewidth]{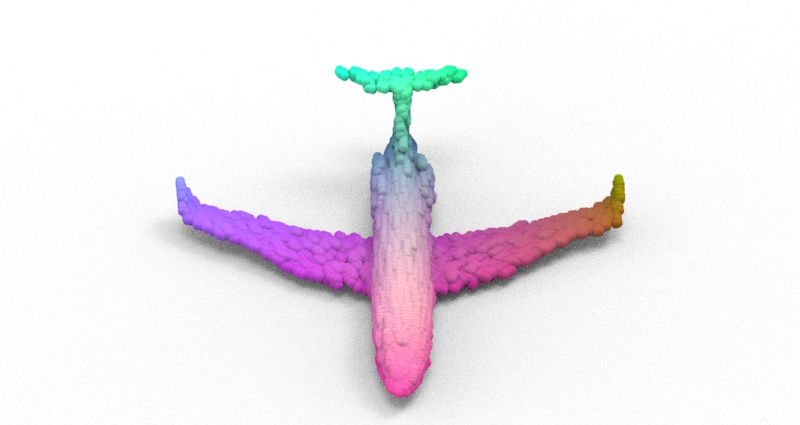}}
	\subfigure{
		\includegraphics[width=0.087\linewidth]{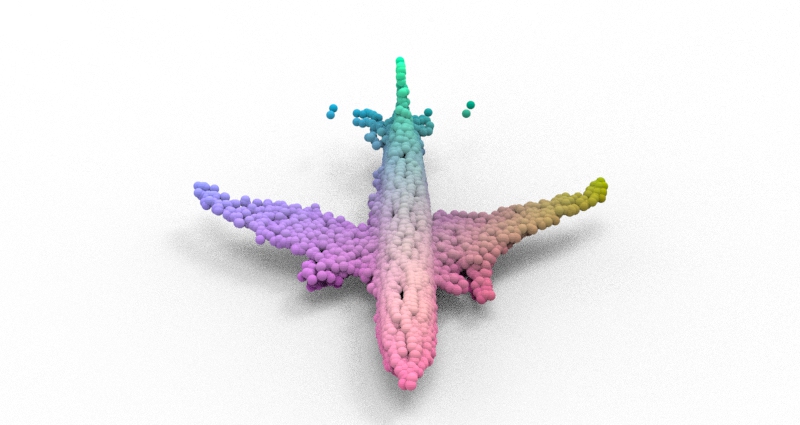}}
	\subfigure{
		\includegraphics[width=0.087\linewidth]{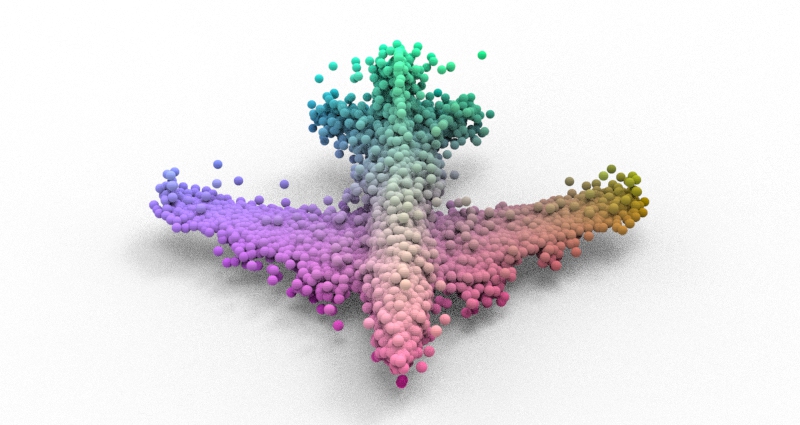}}
        \subfigure{
            \includegraphics[width=0.087\linewidth]{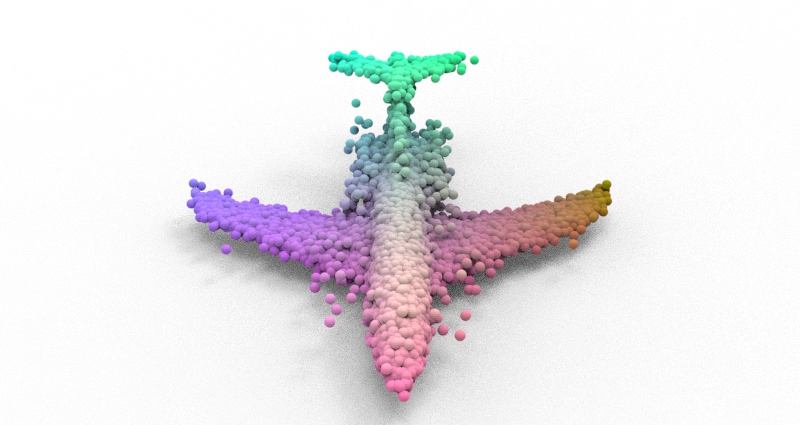}}
        \subfigure{
            \includegraphics[width=0.087\linewidth]{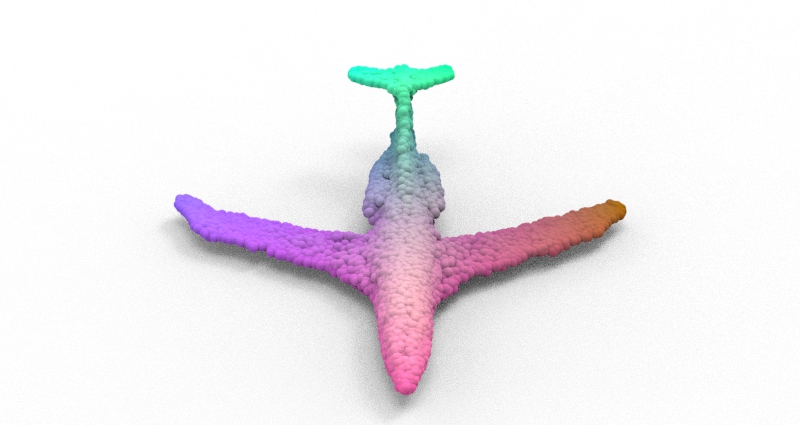}}
        \vspace{3mm}
        \\
	\subfigure{
		\includegraphics[width=0.087\linewidth]{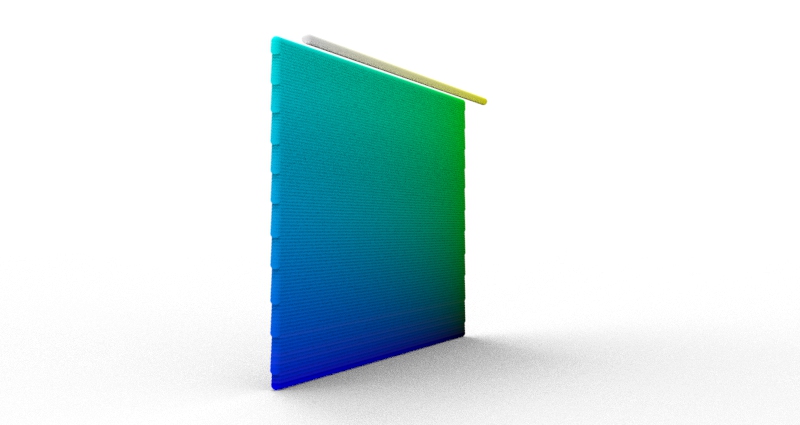}}
	\subfigure{
		\includegraphics[width=0.087\linewidth]{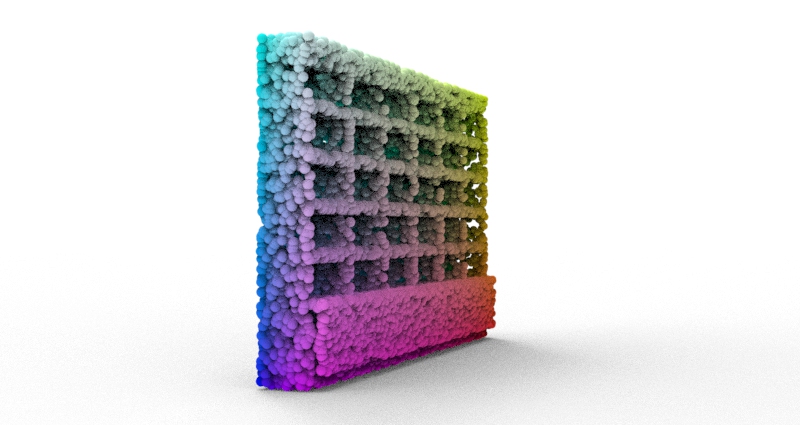}}
	\subfigure{
		\includegraphics[width=0.087\linewidth]{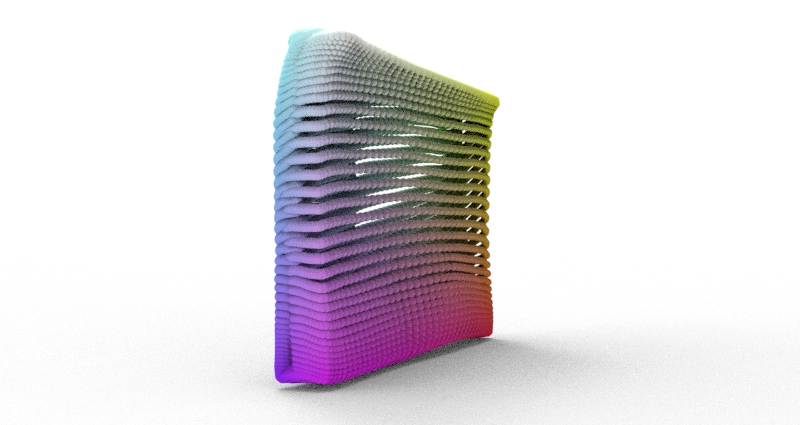}}
	\subfigure{
		\includegraphics[width=0.087\linewidth]{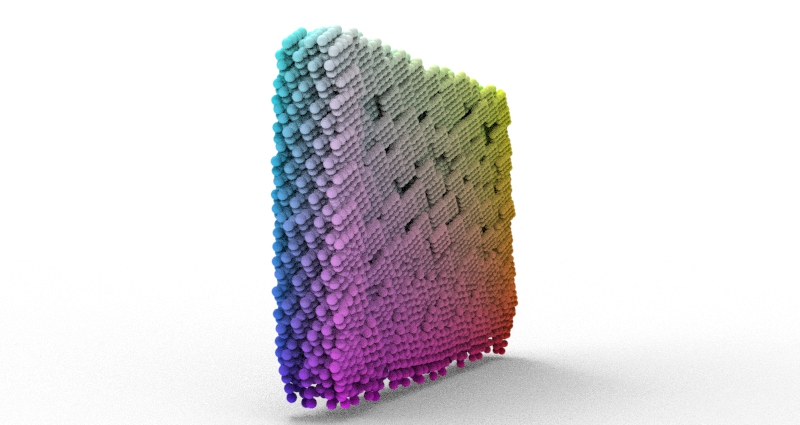}}
	\subfigure{
		\includegraphics[width=0.087\linewidth]{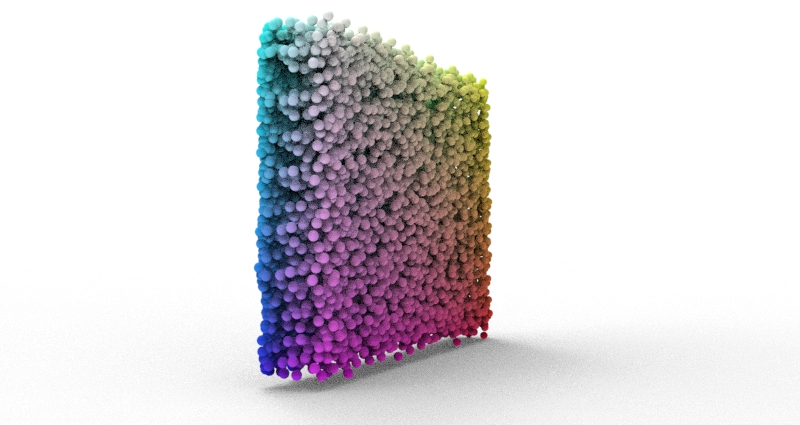}}
	\subfigure{
		\includegraphics[width=0.087\linewidth]{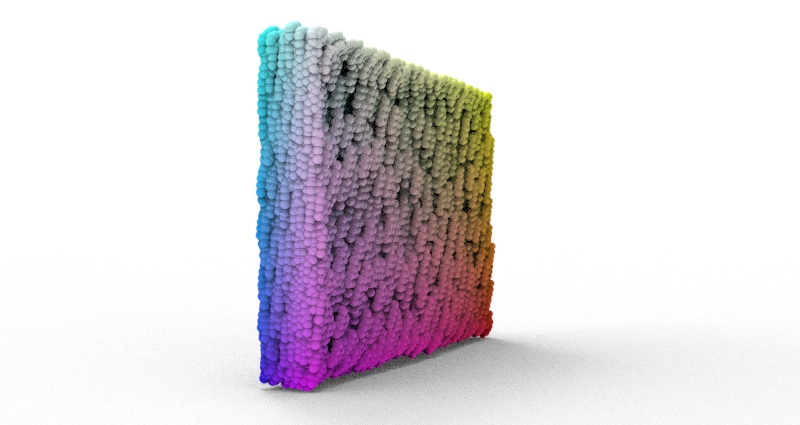}}
	\subfigure{
		\includegraphics[width=0.087\linewidth]{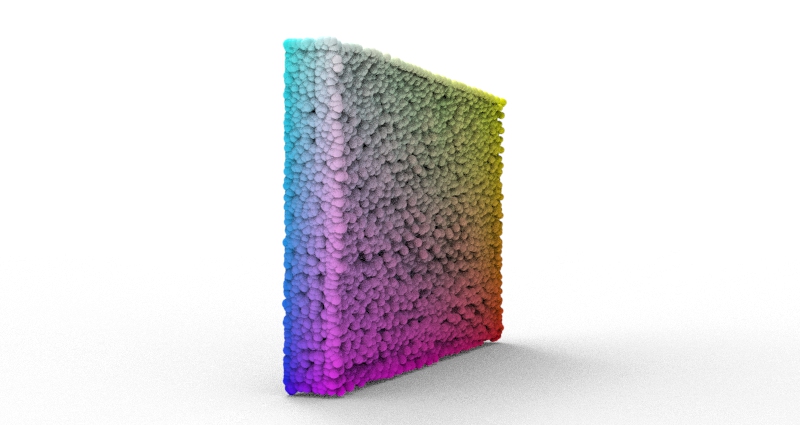}}
	\subfigure{
		\includegraphics[width=0.087\linewidth]{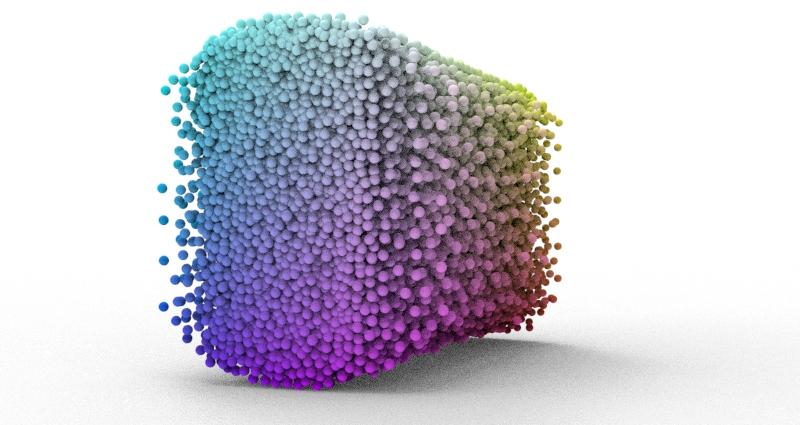}}
	\subfigure{
		\includegraphics[width=0.087\linewidth]{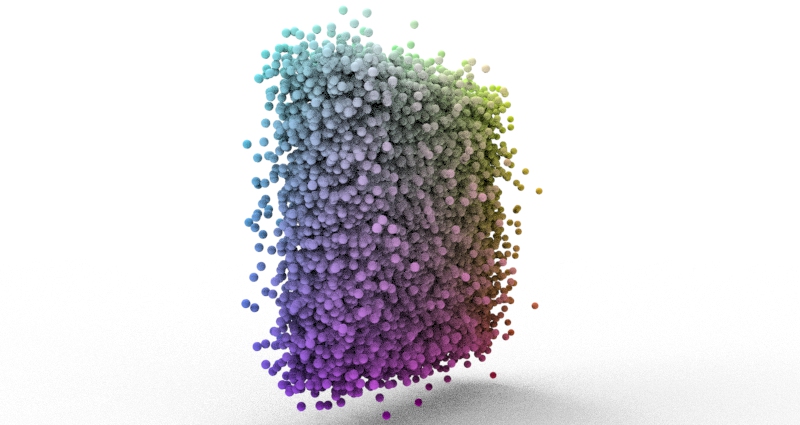}}
        \subfigure{
		\includegraphics[width=0.087\linewidth]{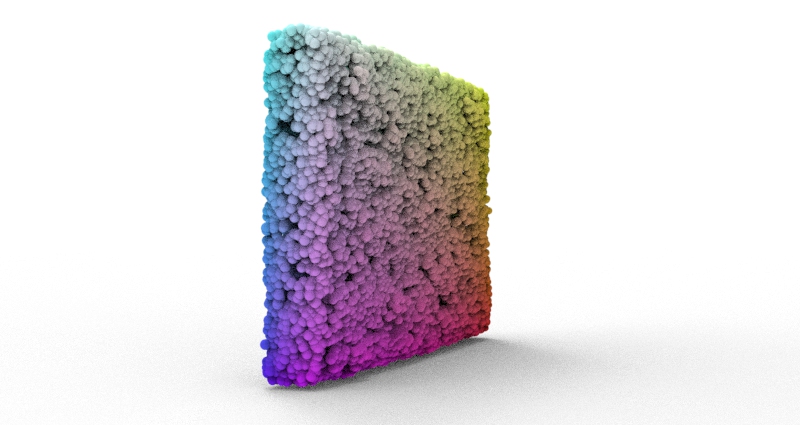}}
	\vspace{3mm}
        \\
	\subfigure{
		\includegraphics[width=0.087\linewidth]{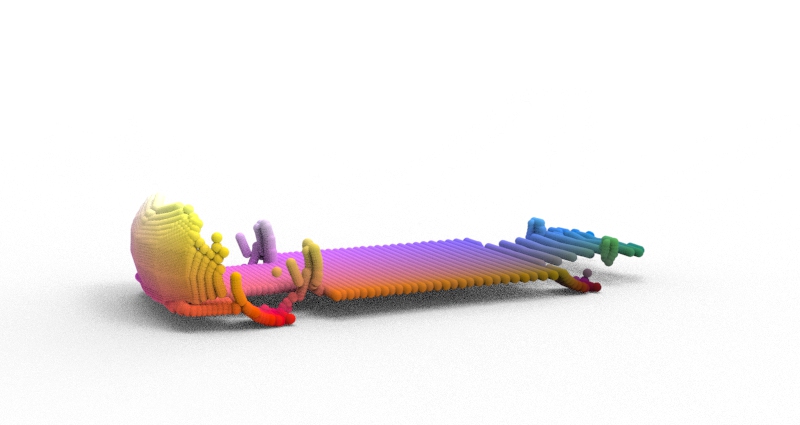}}
	\subfigure{
		\includegraphics[width=0.087\linewidth]{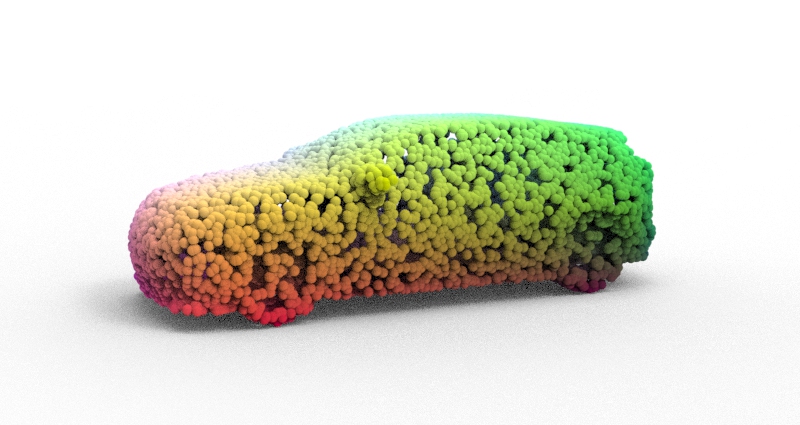}}
	\subfigure{
		\includegraphics[width=0.087\linewidth]{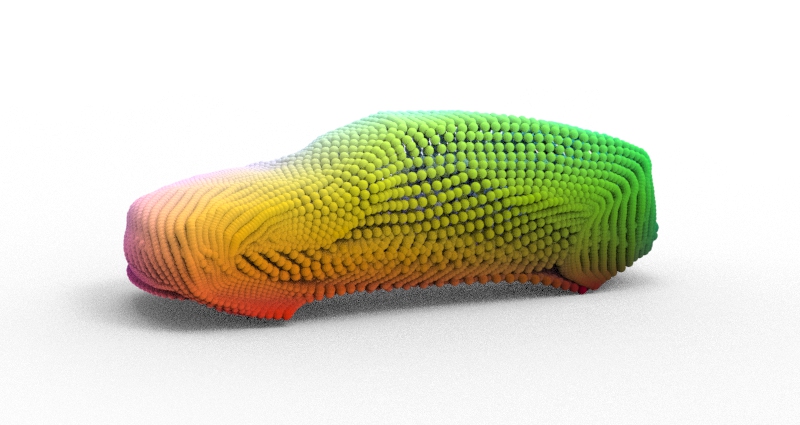}}
	\subfigure{
		\includegraphics[width=0.087\linewidth]{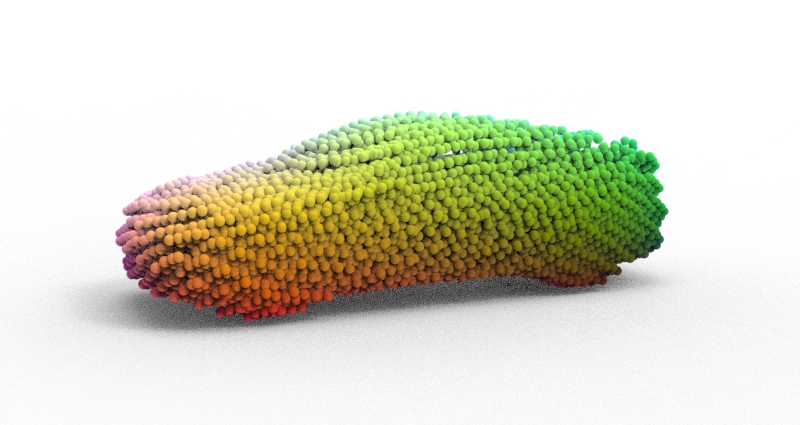}}
	\subfigure{
		\includegraphics[width=0.087\linewidth]{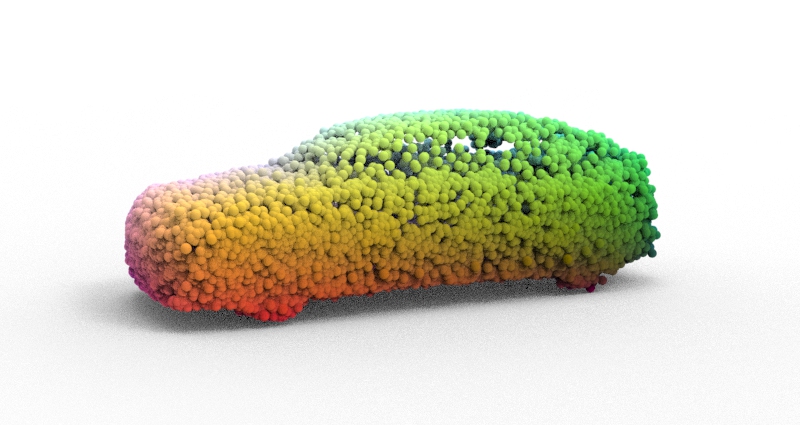}}
	\subfigure{
		\includegraphics[width=0.087\linewidth]{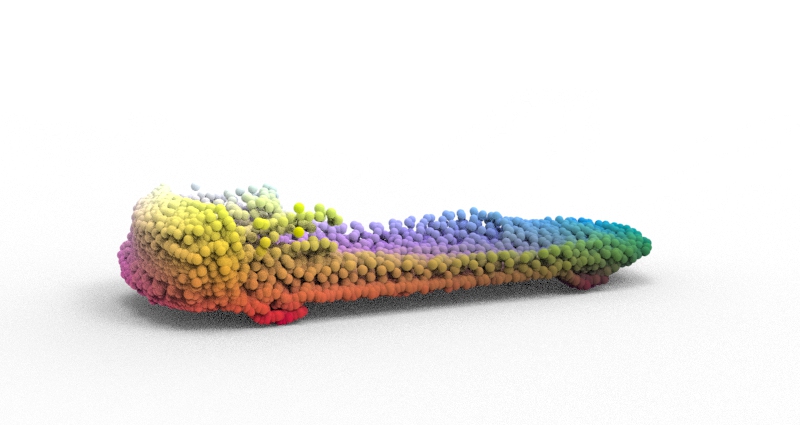}}
	\subfigure{
		\includegraphics[width=0.087\linewidth]{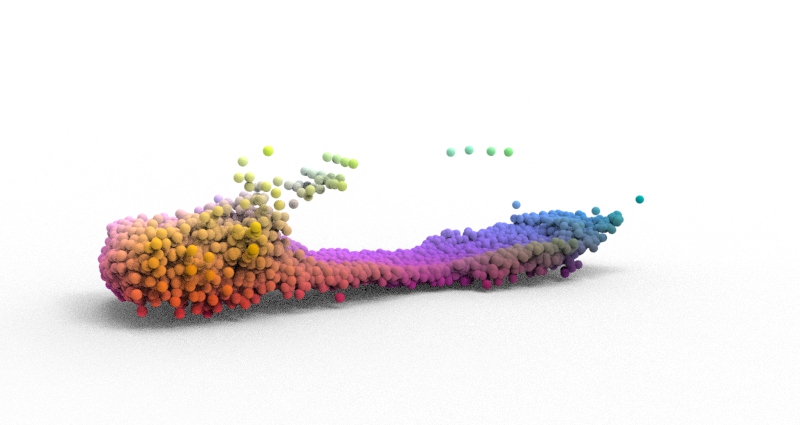}}
	\subfigure{
		\includegraphics[width=0.087\linewidth]{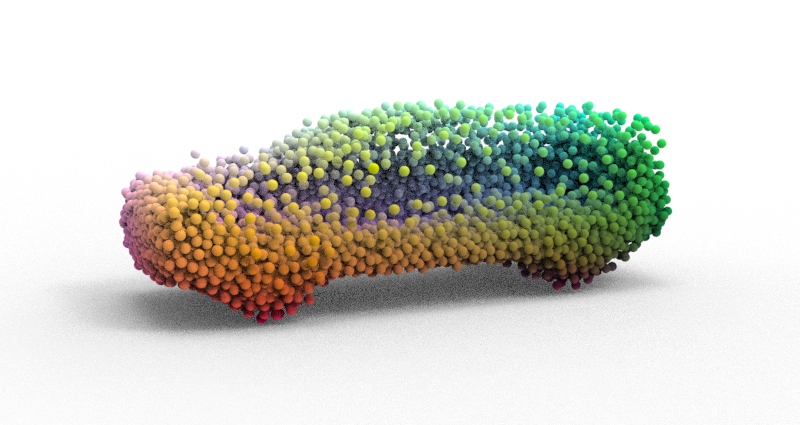}}
	\subfigure{
		\includegraphics[width=0.087\linewidth]{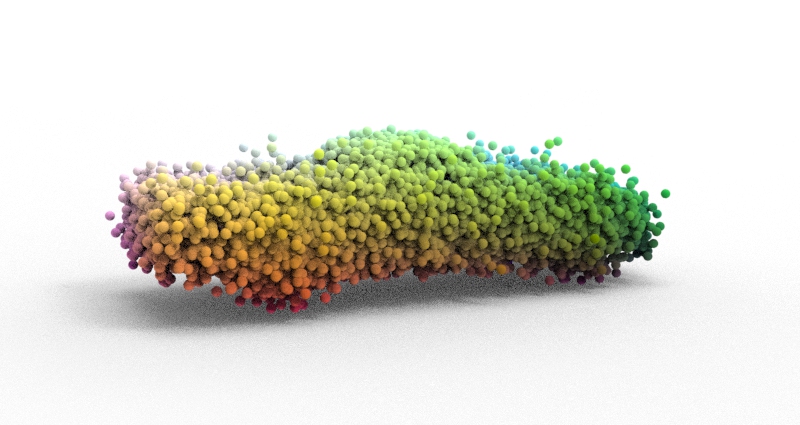}}
        \subfigure{
		\includegraphics[width=0.087\linewidth]{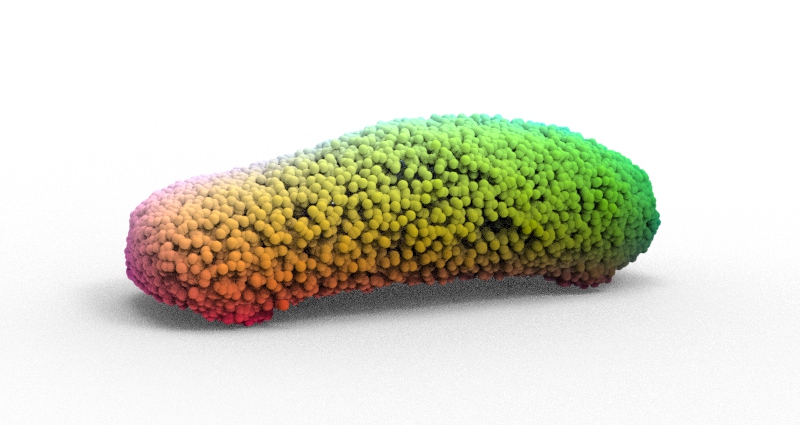}}
	\vspace{3mm}
        \\
	\subfigure{
		\includegraphics[width=0.087\linewidth]{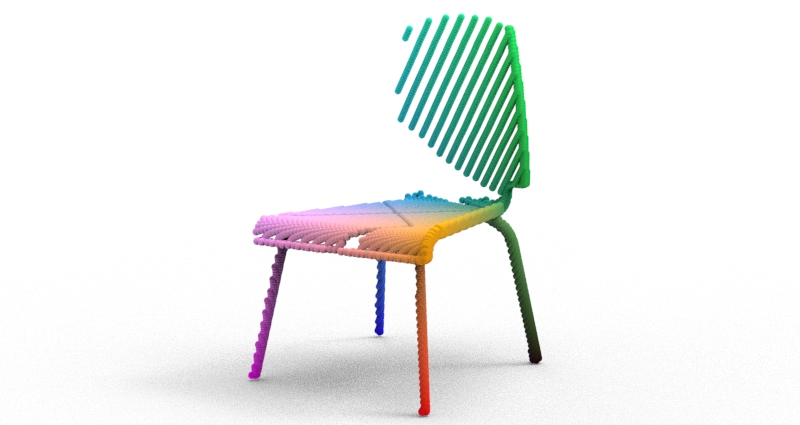}}
	\subfigure{
		\includegraphics[width=0.087\linewidth]{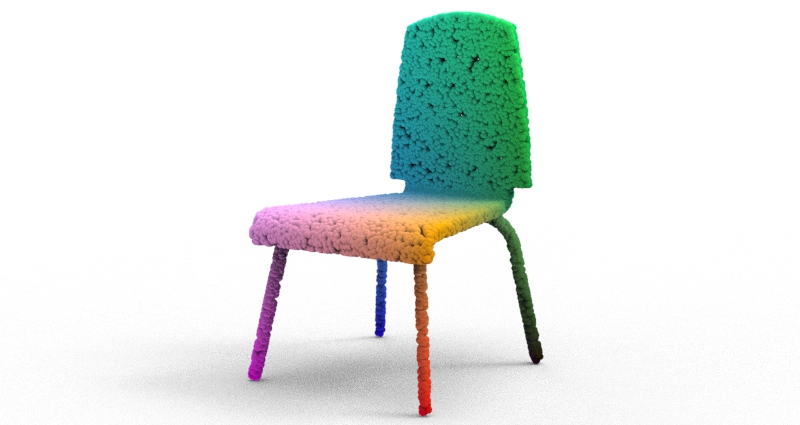}}
	\subfigure{
		\includegraphics[width=0.087\linewidth]{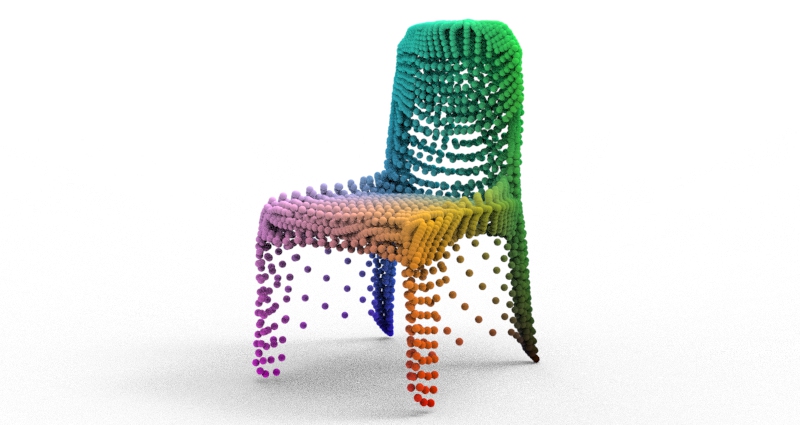}}
	\subfigure{
		\includegraphics[width=0.087\linewidth]{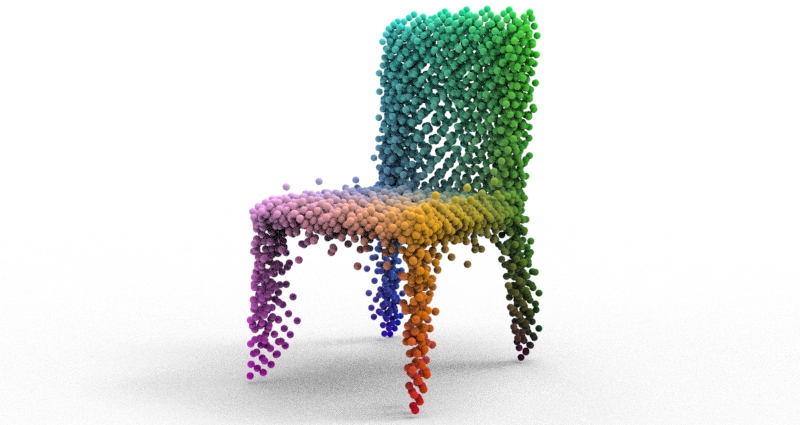}}
	\subfigure{
		\includegraphics[width=0.087\linewidth]{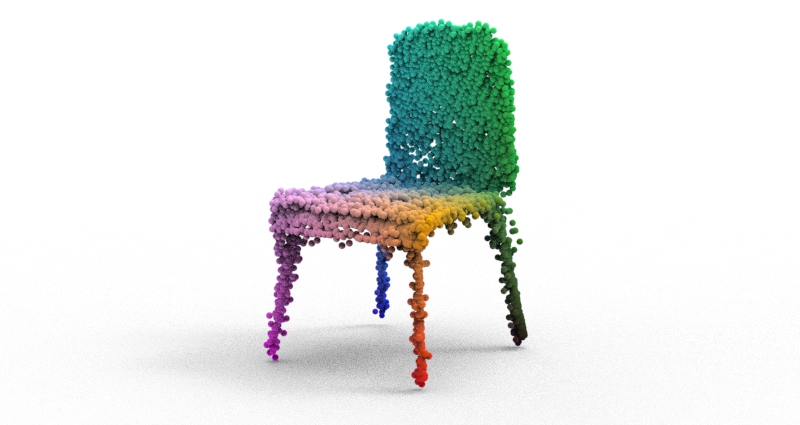}}
	\subfigure{
		\includegraphics[width=0.087\linewidth]{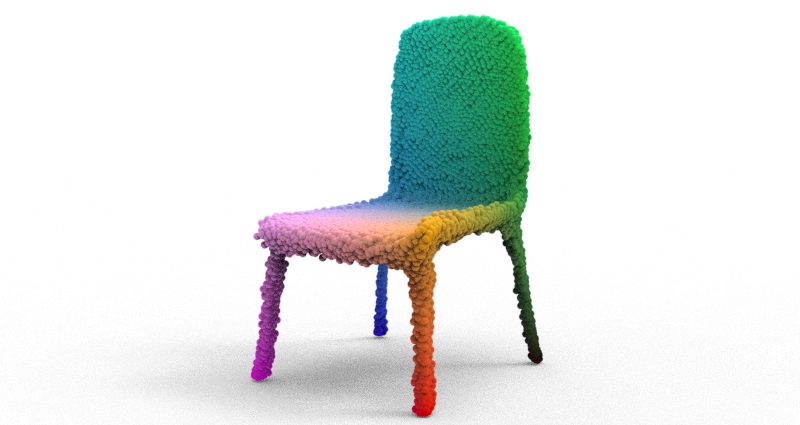}}
	\subfigure{
		\includegraphics[width=0.087\linewidth]{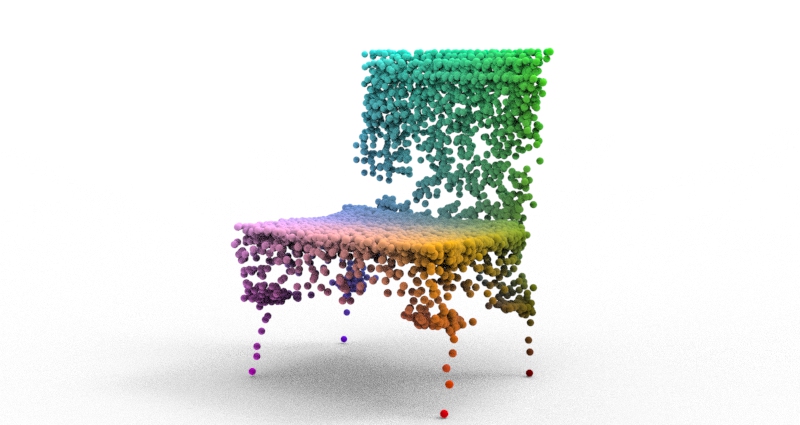}}
	\subfigure{
		\includegraphics[width=0.087\linewidth]{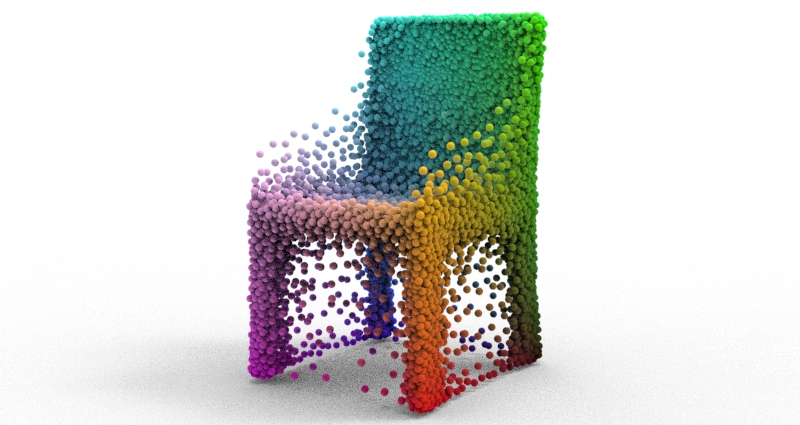}}
	\subfigure{
		\includegraphics[width=0.087\linewidth]{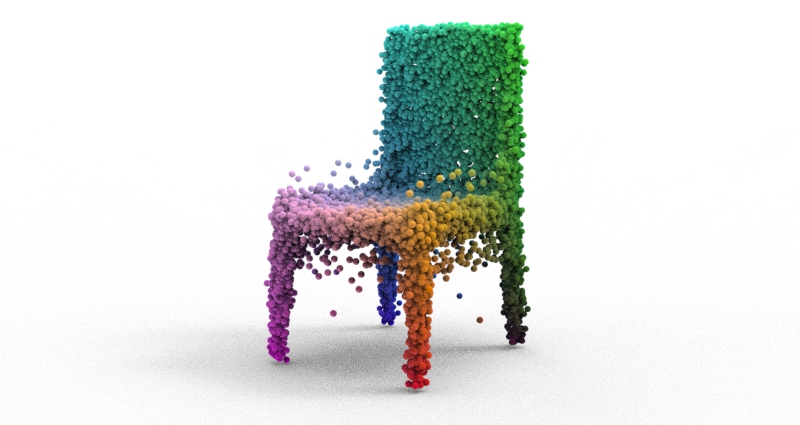}}
        \subfigure{
		\includegraphics[width=0.087\linewidth]{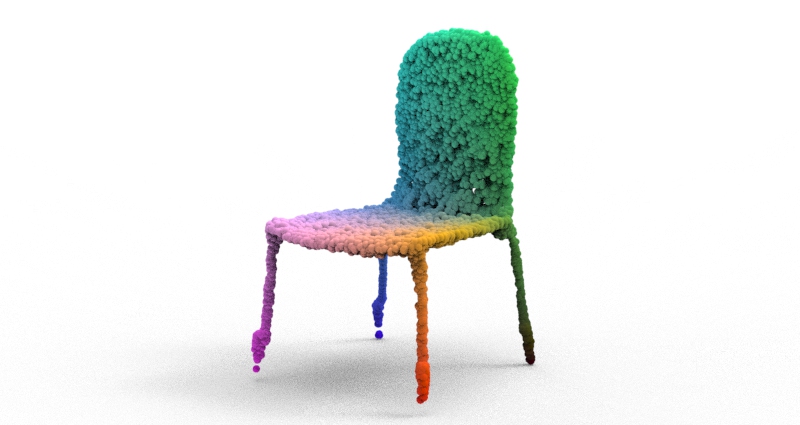}}
	\vspace{3.5mm}
        \\
	\subfigure{
		\includegraphics[width=0.087\linewidth]{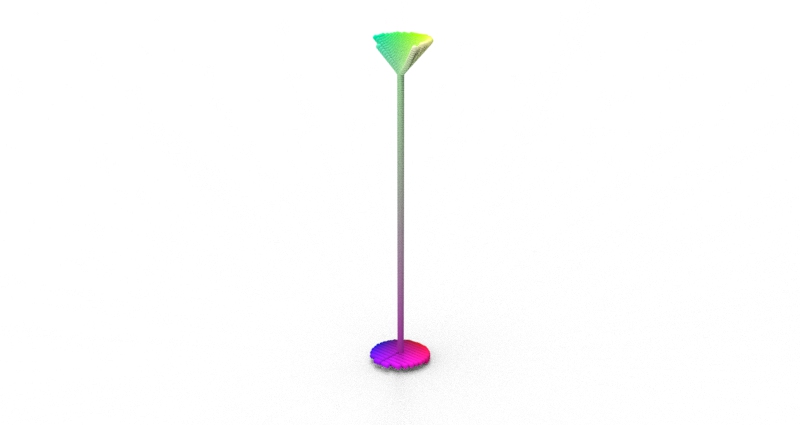}}
	\subfigure{
		\includegraphics[width=0.087\linewidth]{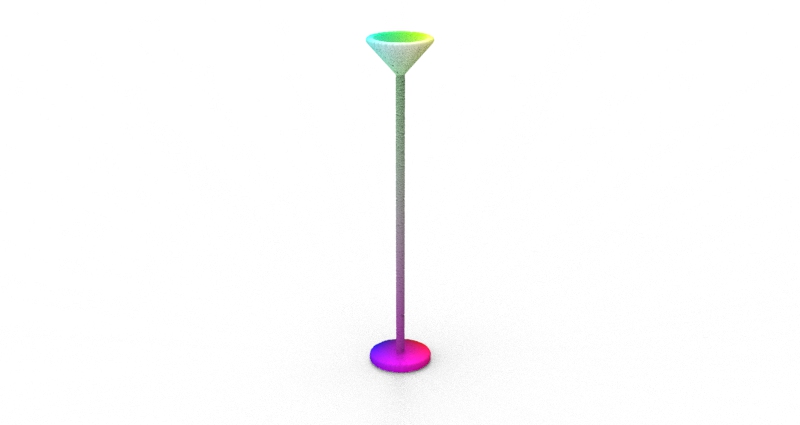}}
	\subfigure{
		\includegraphics[width=0.087\linewidth]{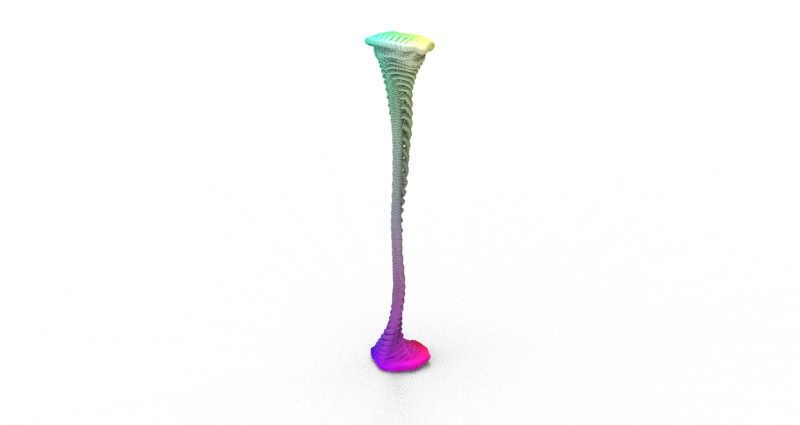}}
	\subfigure{
		\includegraphics[width=0.087\linewidth]{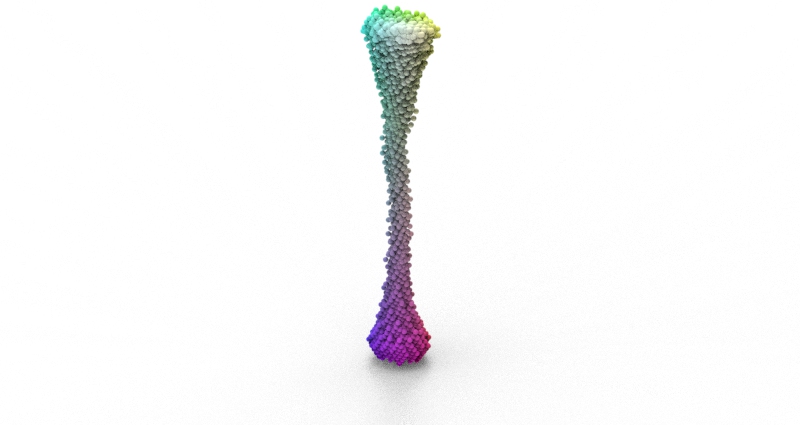}}
	\subfigure{
		\includegraphics[width=0.087\linewidth]{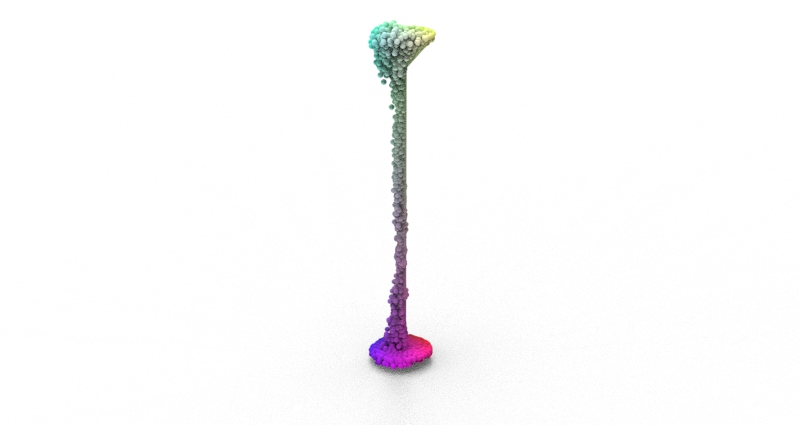}}
	\subfigure{
		\includegraphics[width=0.087\linewidth]{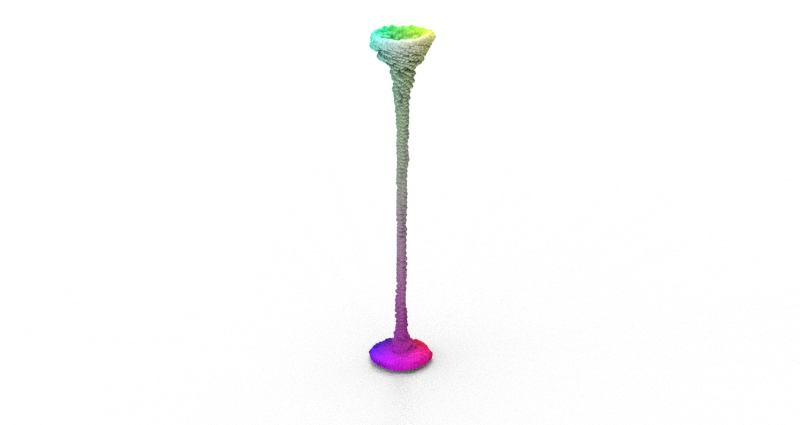}}
	\subfigure{
		\includegraphics[width=0.087\linewidth]{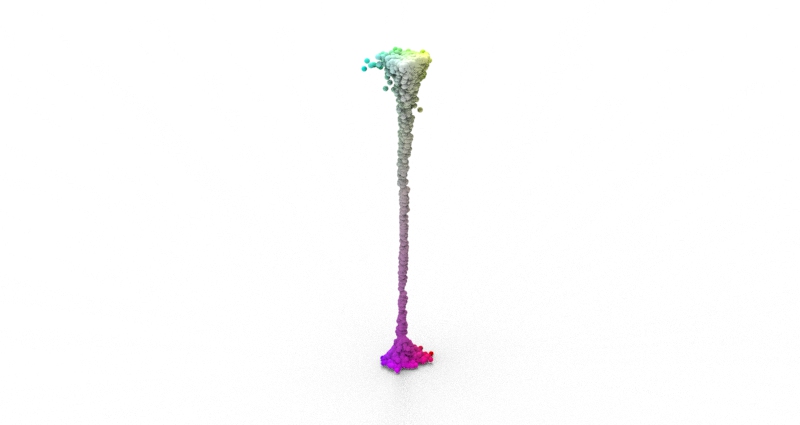}}
	\subfigure{
		\includegraphics[width=0.087\linewidth]{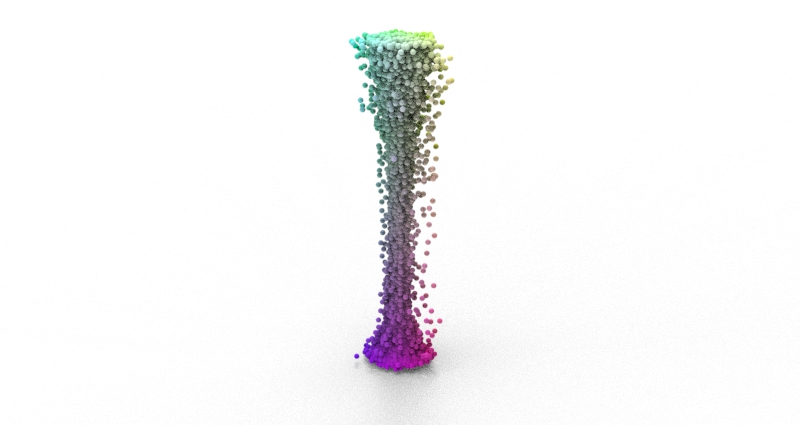}}
	\subfigure{
		\includegraphics[width=0.087\linewidth]{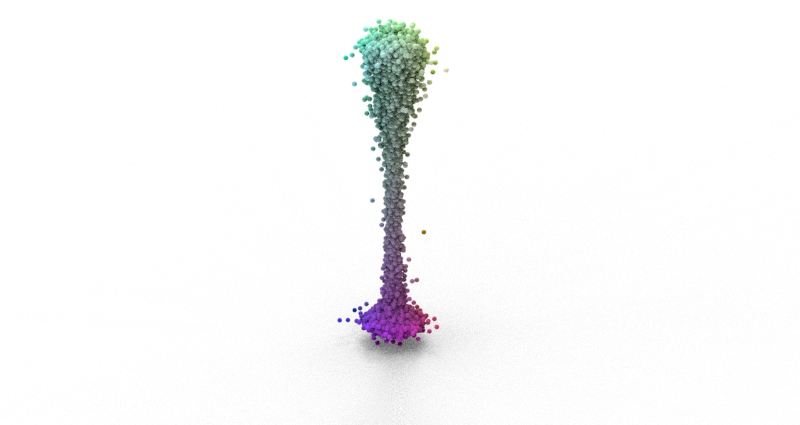}}
        \subfigure{
		\includegraphics[width=0.087\linewidth]{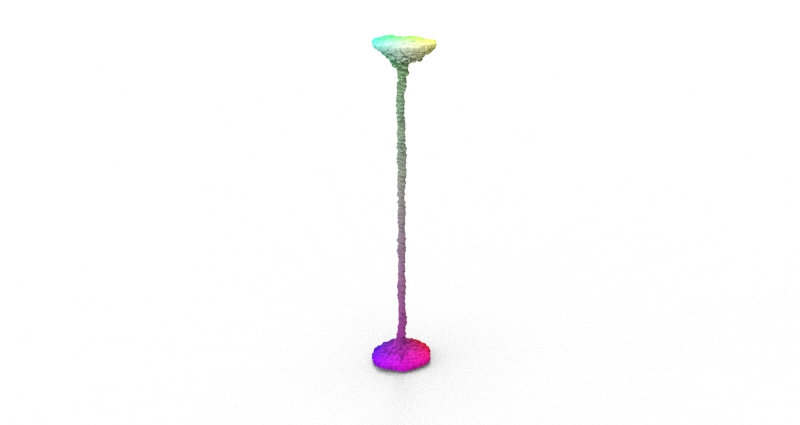}}
	\vspace{2.5mm}
        \\
	\subfigure{
		\includegraphics[width=0.087\linewidth]{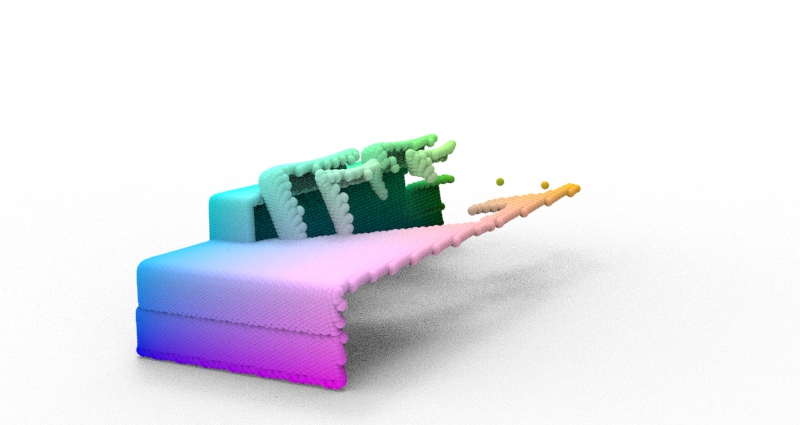}}
	\subfigure{
		\includegraphics[width=0.087\linewidth]{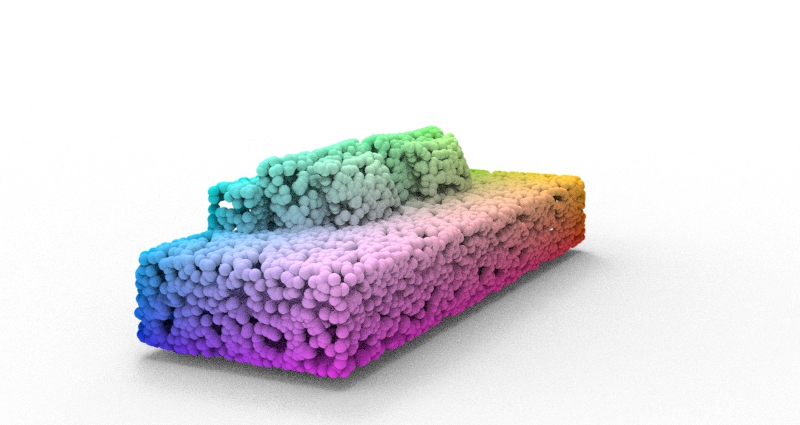}}
	\subfigure{
		\includegraphics[width=0.087\linewidth]{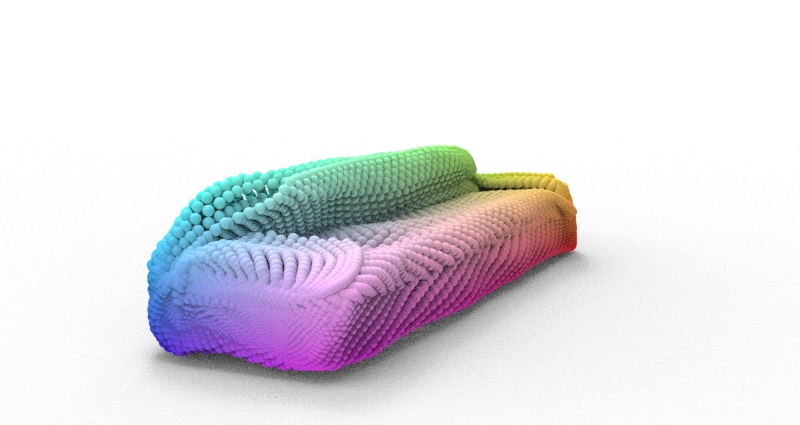}}
	\subfigure{
		\includegraphics[width=0.087\linewidth]{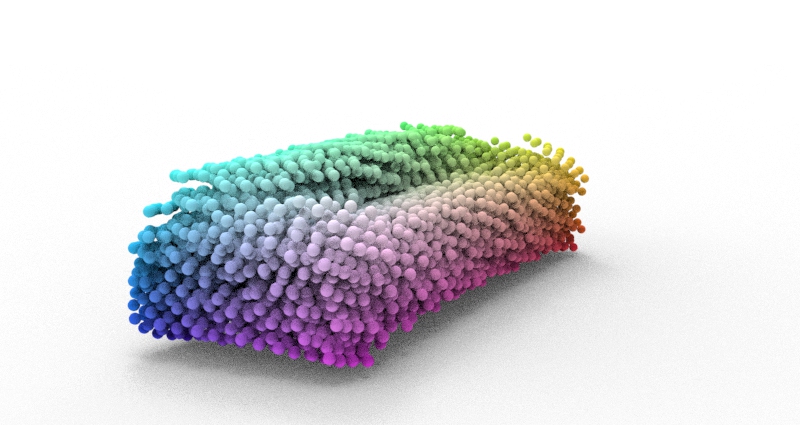}}
	\subfigure{
		\includegraphics[width=0.087\linewidth]{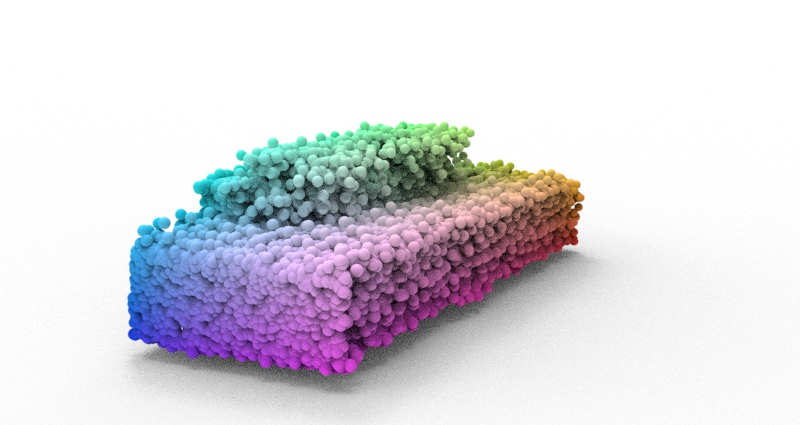}}
	\subfigure{
		\includegraphics[width=0.087\linewidth]{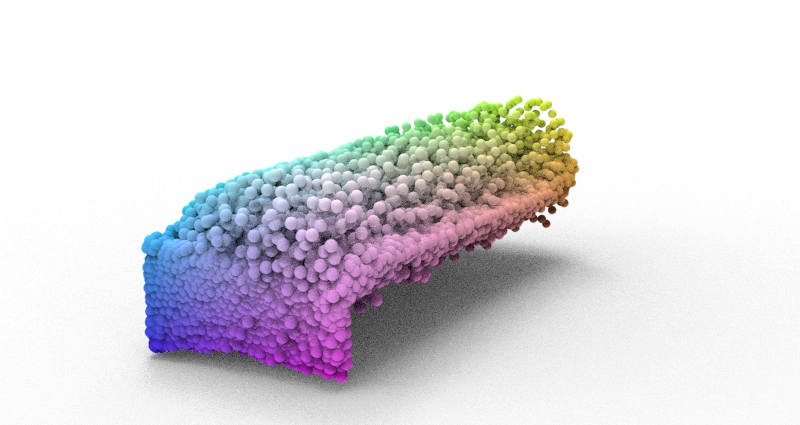}}
	\subfigure{
		\includegraphics[width=0.087\linewidth]{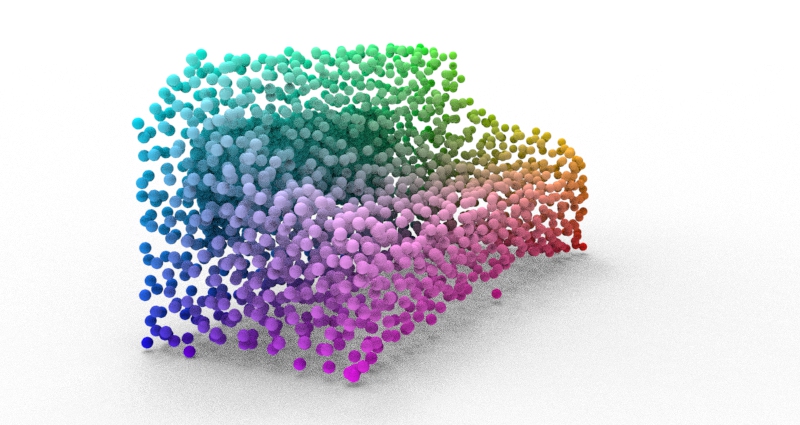}}
	\subfigure{
		\includegraphics[width=0.087\linewidth]{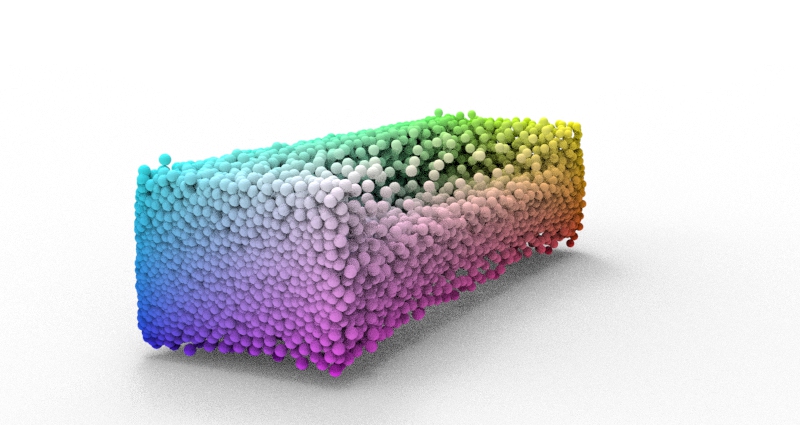}}
	\subfigure{
		\includegraphics[width=0.087\linewidth]{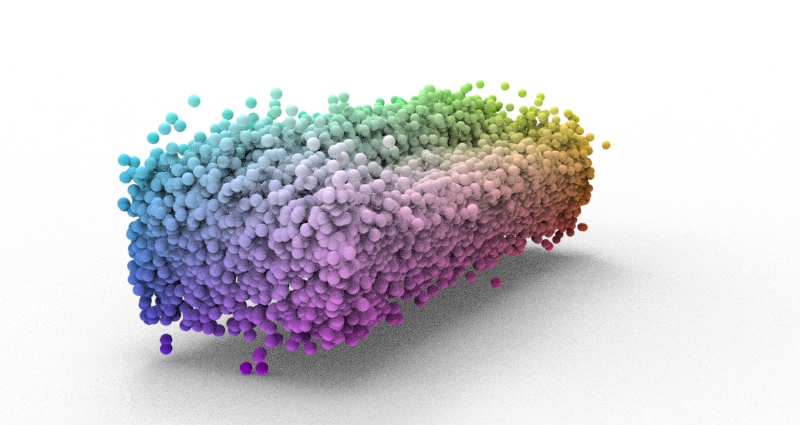}}
        \subfigure{
		\includegraphics[width=0.087\linewidth]{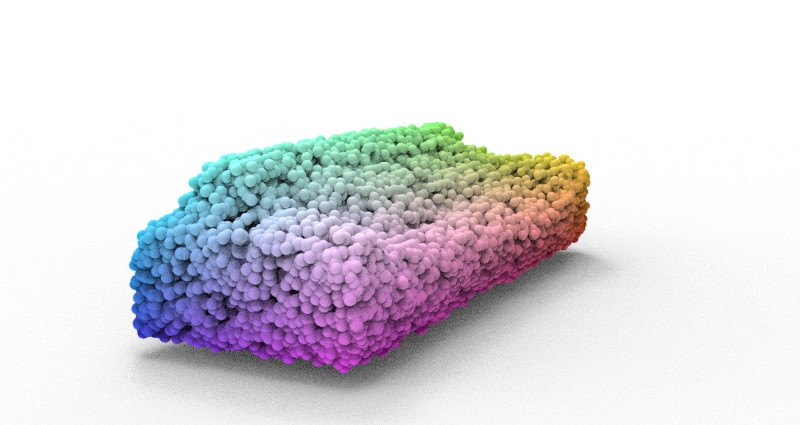}}
	\vspace{4mm}
        \\
	\subfigure{
		\includegraphics[width=0.087\linewidth]{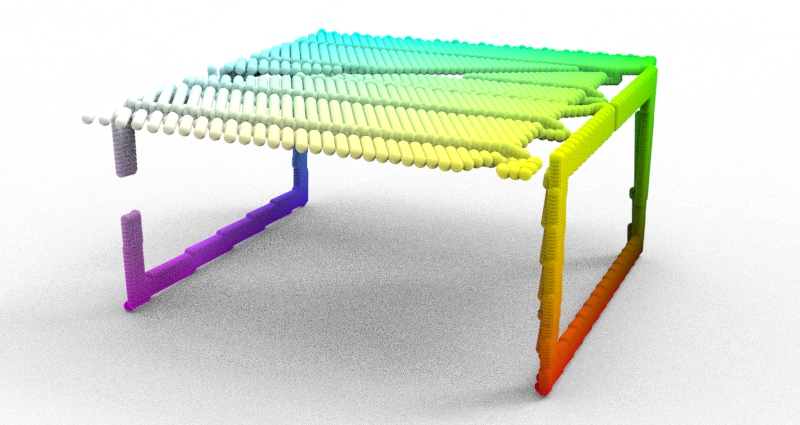}}
	\subfigure{
		\includegraphics[width=0.087\linewidth]{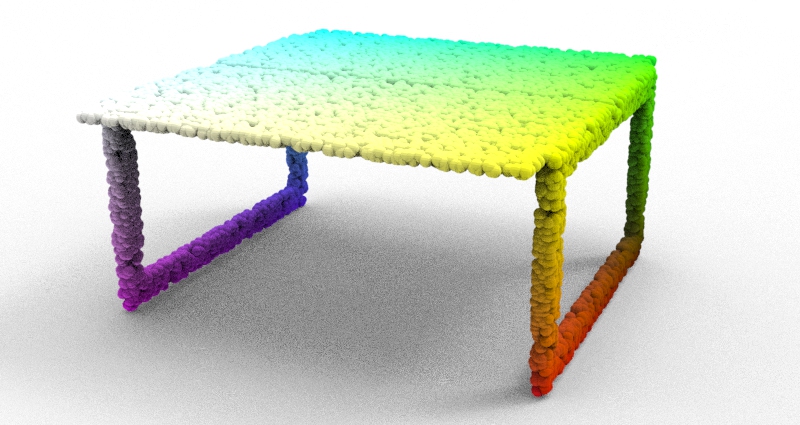}}
	\subfigure{
		\includegraphics[width=0.087\linewidth]{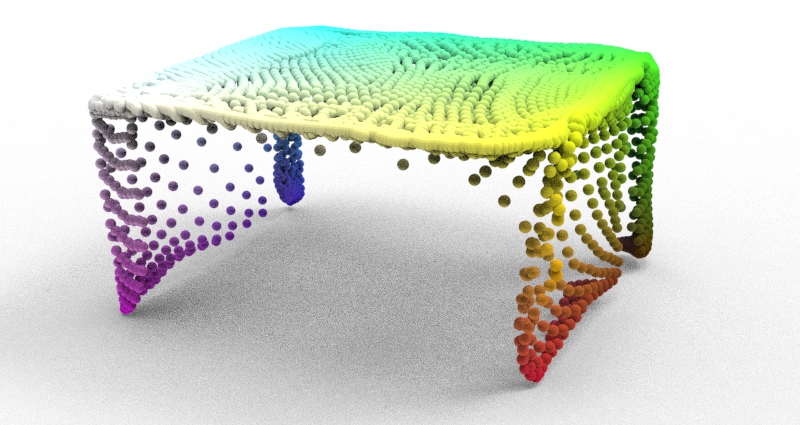}}
	\subfigure{
		\includegraphics[width=0.087\linewidth]{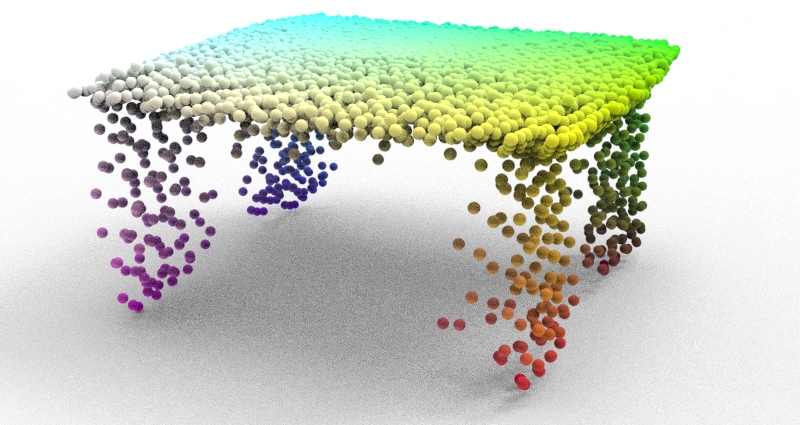}}
	\subfigure{
		\includegraphics[width=0.087\linewidth]{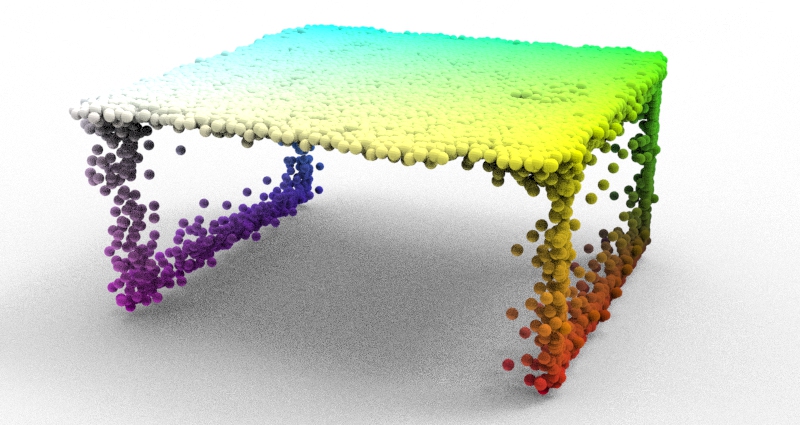}}
	\subfigure{
		\includegraphics[width=0.087\linewidth]{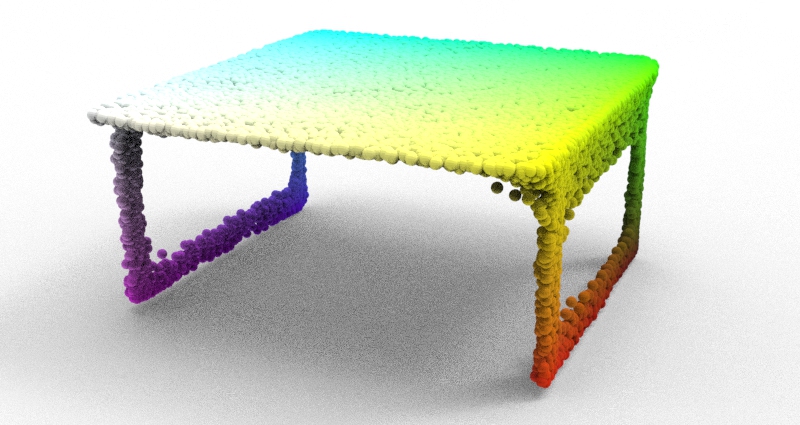}}
	\subfigure{
		\includegraphics[width=0.087\linewidth]{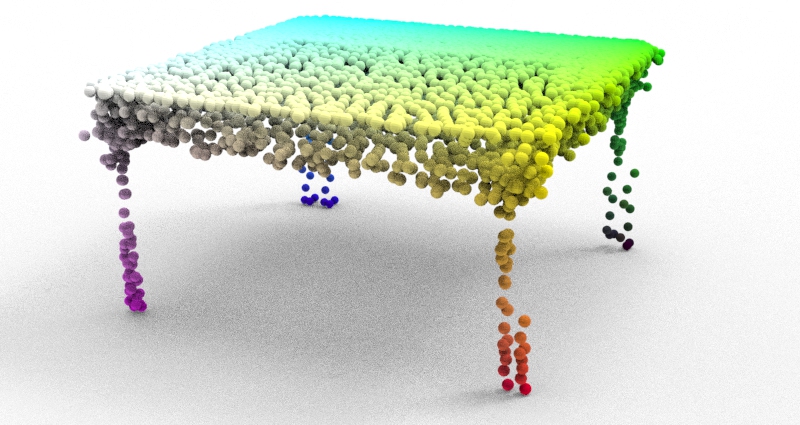}}
	\subfigure{
		\includegraphics[width=0.087\linewidth]{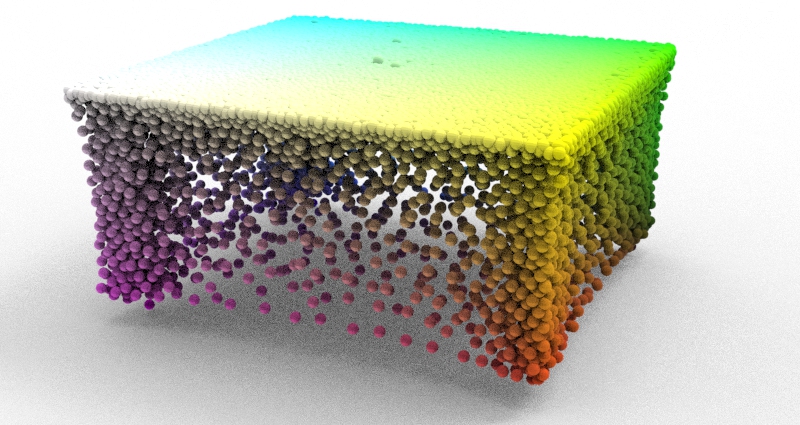}}
	\subfigure{
		\includegraphics[width=0.087\linewidth]{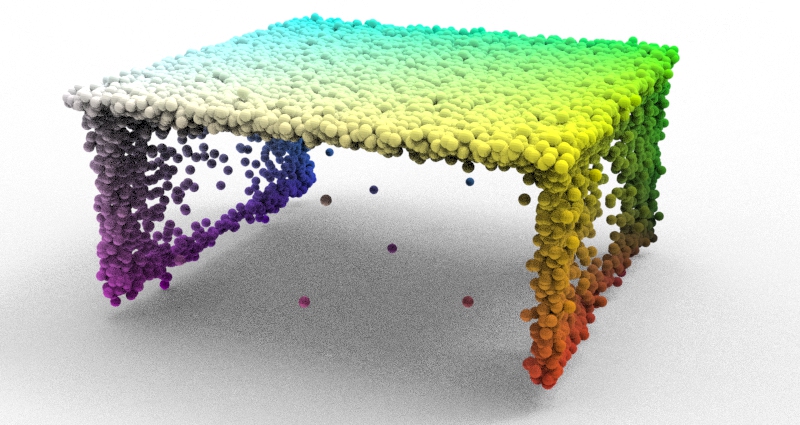}}
        \subfigure{
		\includegraphics[width=0.087\linewidth]{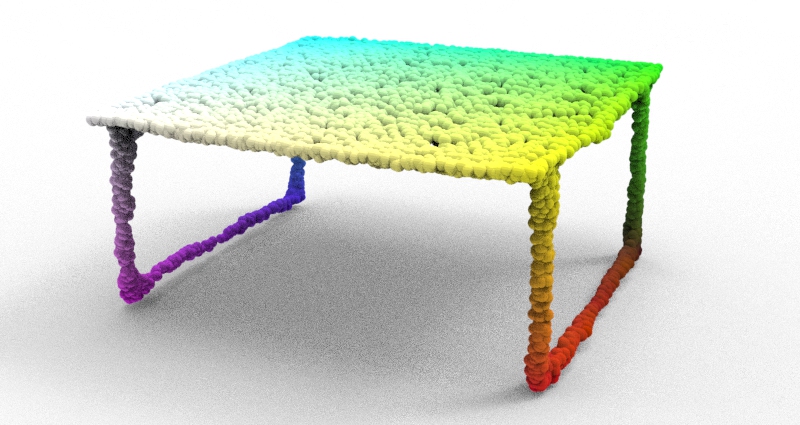}}
	\vspace{3.5mm}
        \\
	\setcounter{subfigure}{0}
	\subfigure[]{
		\includegraphics[width=0.087\linewidth]{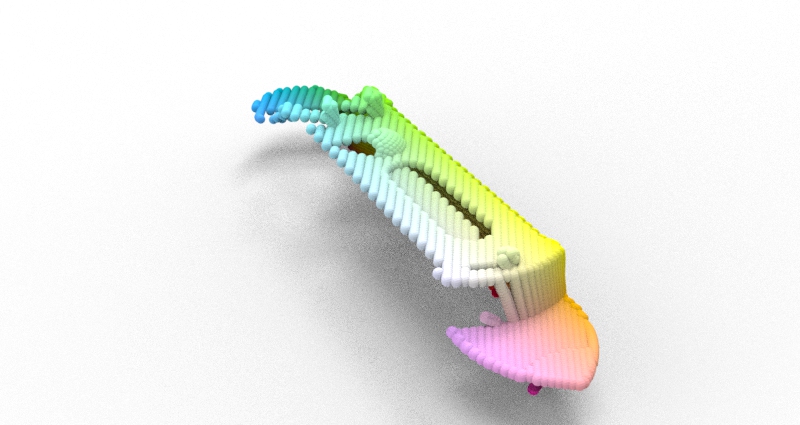}}
	\subfigure[]{
		\includegraphics[width=0.087\linewidth]{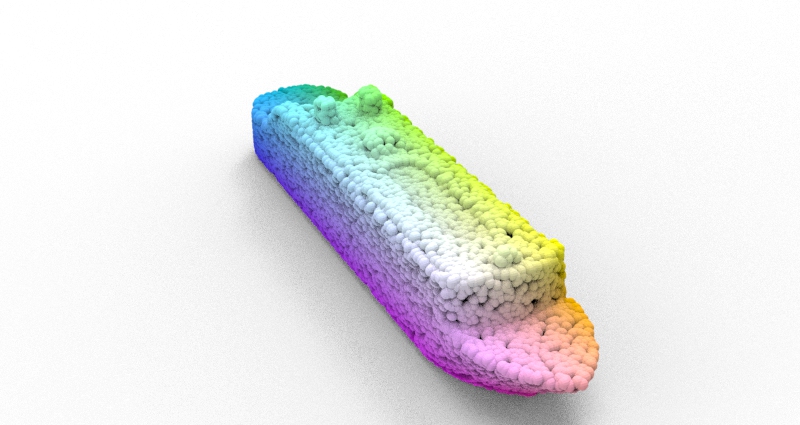}}
	\subfigure[]{
		\includegraphics[width=0.087\linewidth]{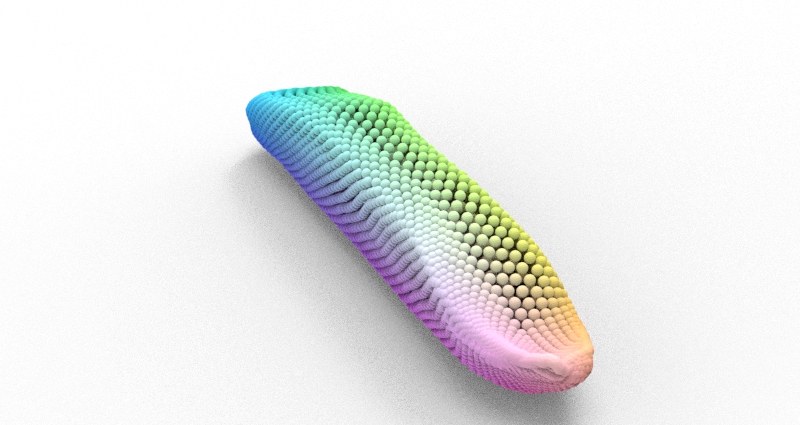}}
	\subfigure[]{
		\includegraphics[width=0.087\linewidth]{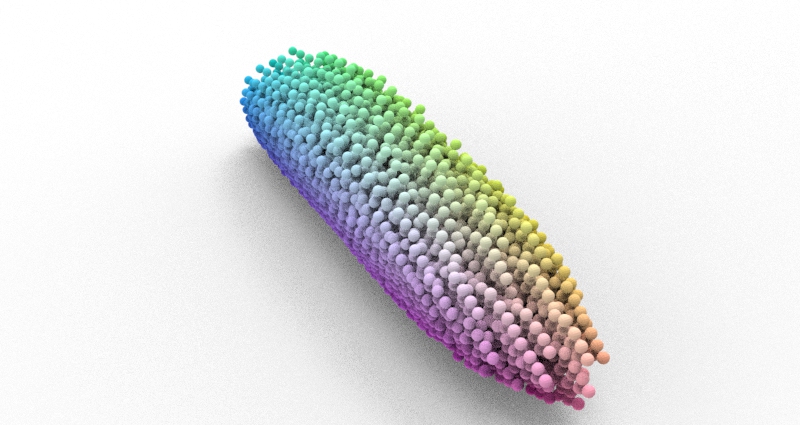}}
	\subfigure[]{
		\includegraphics[width=0.087\linewidth]{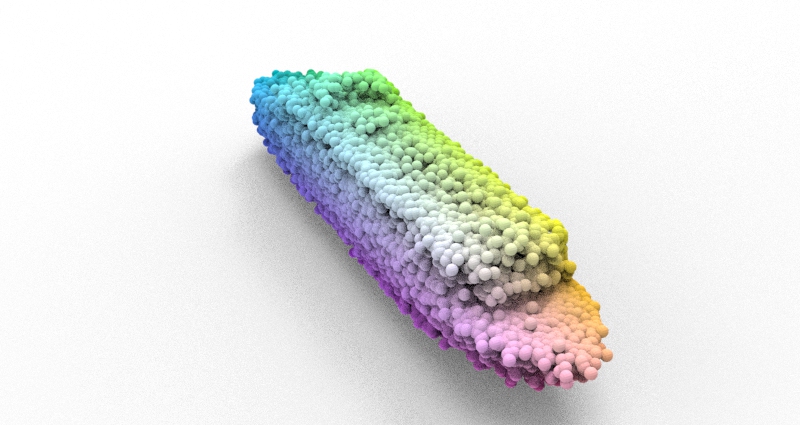}}
	\subfigure[]{
		\includegraphics[width=0.087\linewidth]{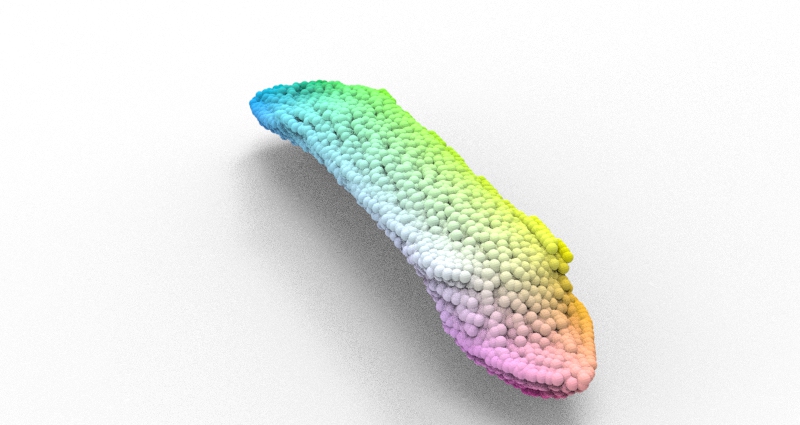}}
	\subfigure[]{
		\includegraphics[width=0.087\linewidth]{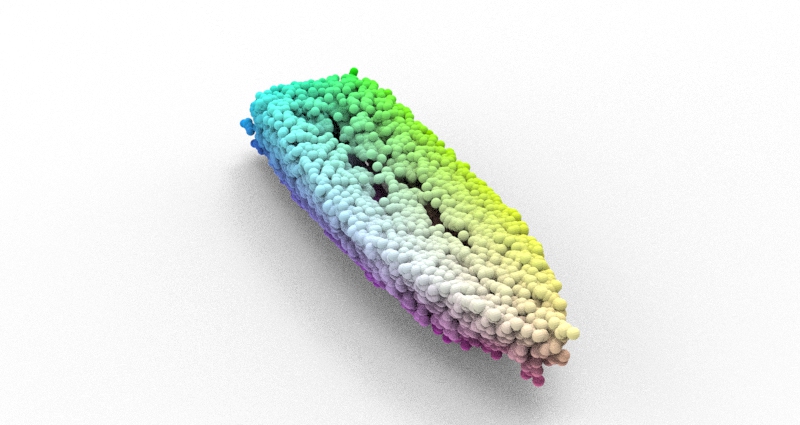}}
	\subfigure[]{
		\includegraphics[width=0.087\linewidth]{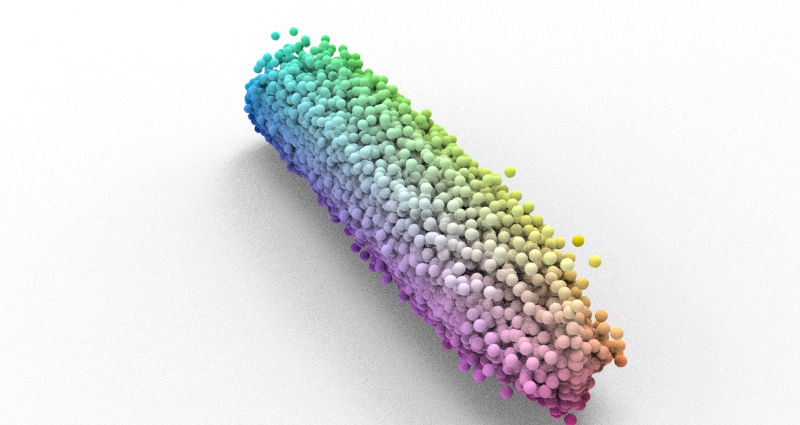}}
	\subfigure[]{
		\includegraphics[width=0.087\linewidth]{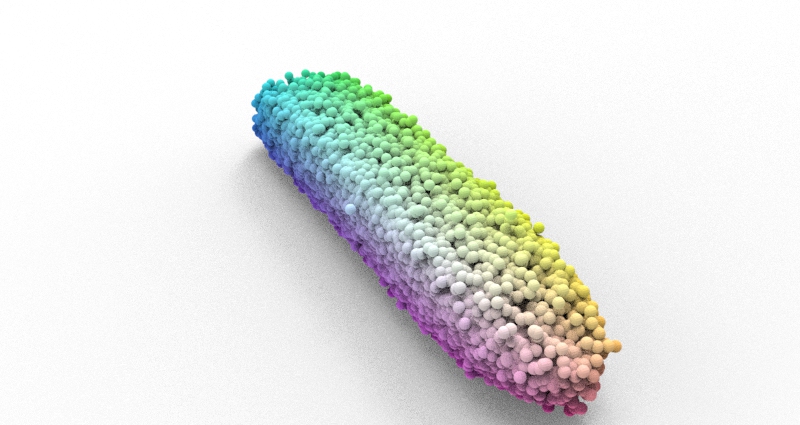}}
        \subfigure[]{
		\includegraphics[width=0.087\linewidth]{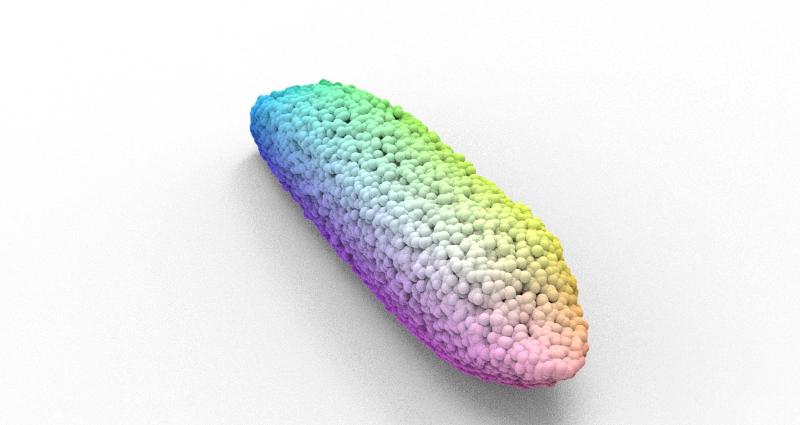}}
	\caption{Qualitative results on the synthetic dataset. (a) Partial points. (b) Ground truth. (c) FoldingNet \cite{foldingnet}. (d) PCN \cite{pcn}. (e) AnchorFormer \cite{chen2023anchorformer}. (f) KTNet \cite{cao2023kt}. (g) USSPA \cite{ma2023symmetric}. (h) ACL-SPC \cite{hong2023acl}. (i) P2C \cite{cui2023p2c}. (j) Ours.}
	\label{visual_comparison}
\end{figure*}

\begin{figure*}[htbp]
	\centering  
	\subfigbottomskip=1pt 
	\subfigure{
		\includegraphics[width=0.095\linewidth]{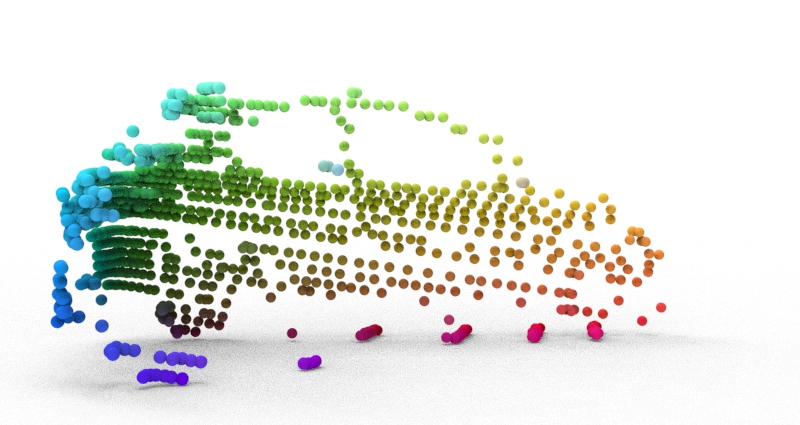}}
	\subfigure{
		\includegraphics[width=0.095\linewidth]{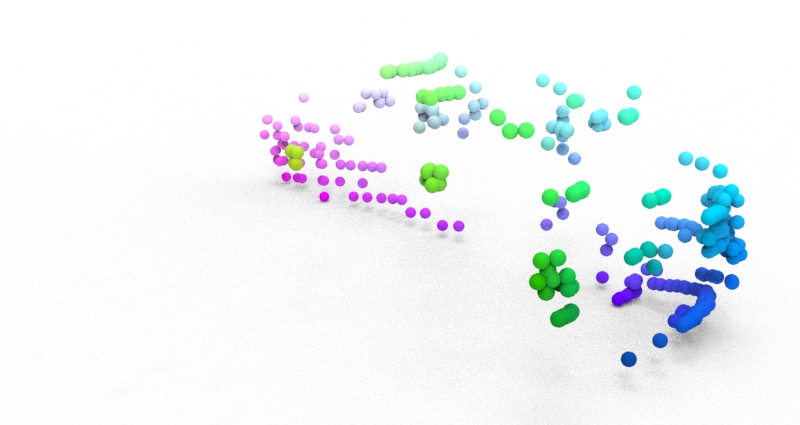}}
        \subfigure{
		\includegraphics[width=0.095\linewidth]{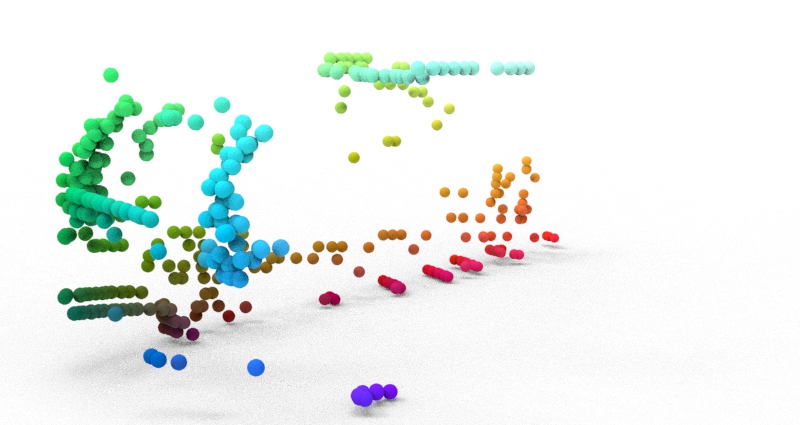}}
	\subfigure{
		\includegraphics[width=0.095\linewidth]{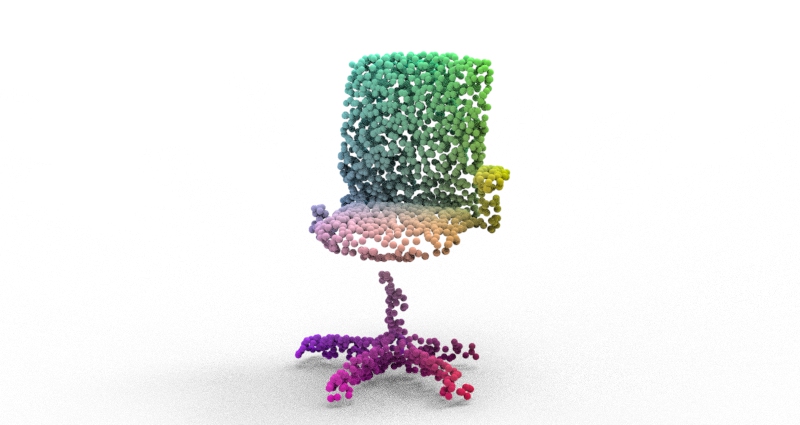}}
	\subfigure{
		\includegraphics[width=0.095\linewidth]{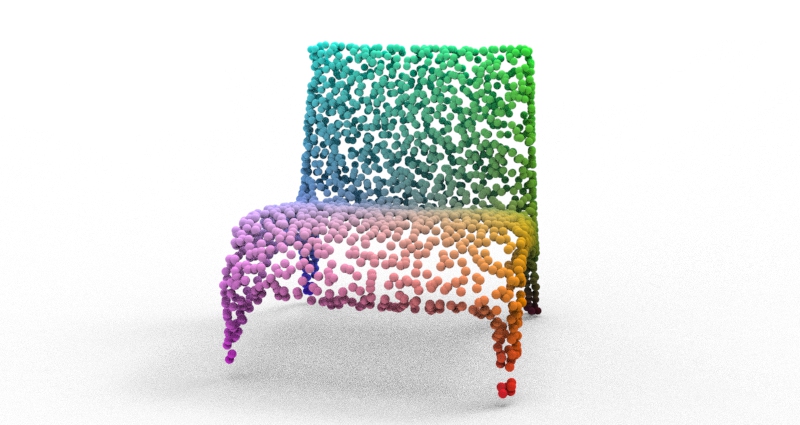}}
	\subfigure{
		\includegraphics[width=0.095\linewidth]{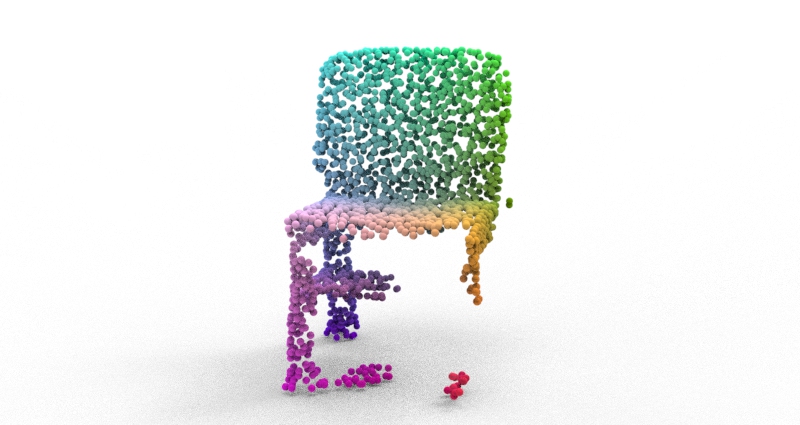}}
	\subfigure{
		\includegraphics[width=0.095\linewidth]{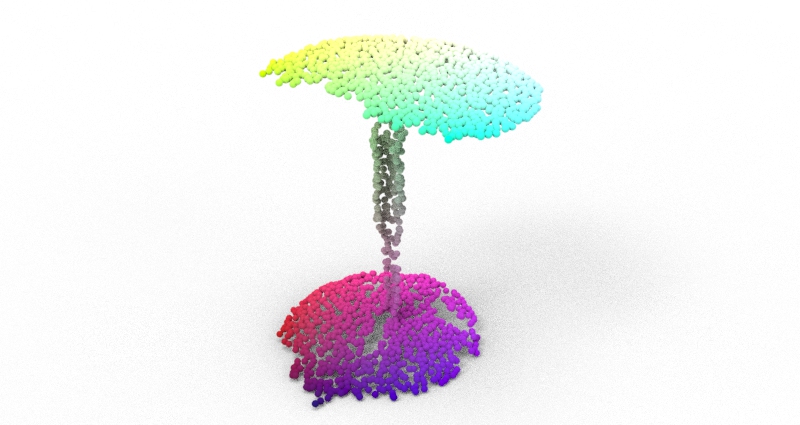}}
	\subfigure{
		\includegraphics[width=0.095\linewidth]{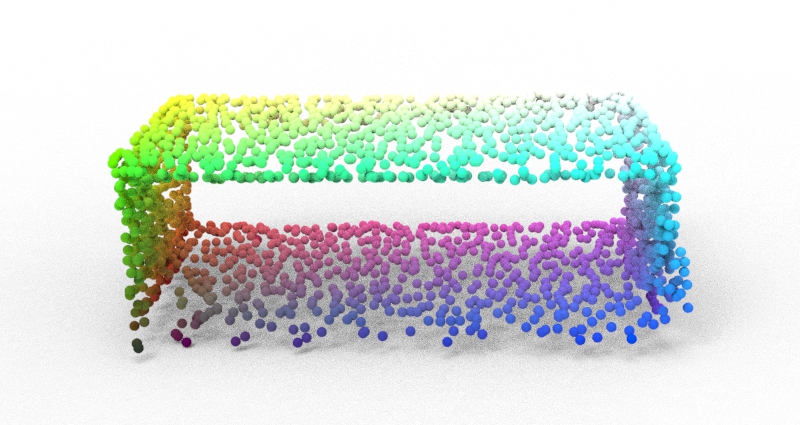}}
        \subfigure{
            \includegraphics[width=0.095\linewidth]{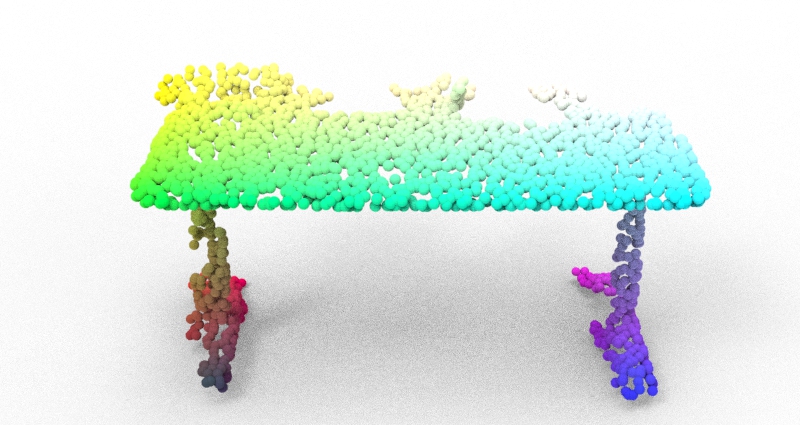}}
        \vspace{5mm}
        \\
	\subfigure{
		\includegraphics[width=0.095\linewidth]{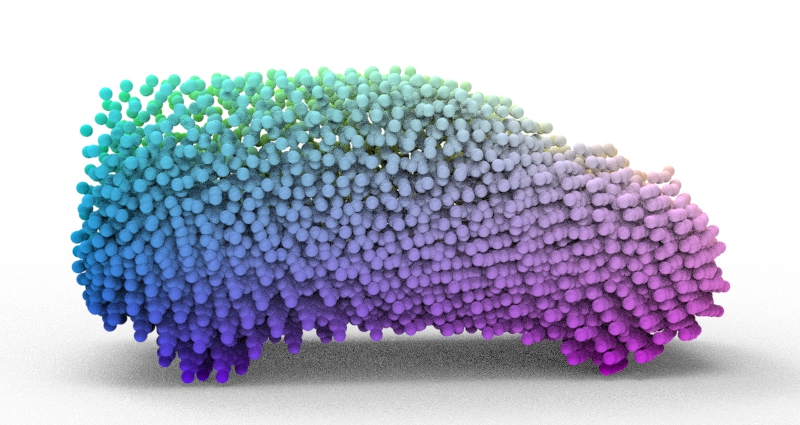}}
	\subfigure{
		\includegraphics[width=0.095\linewidth]{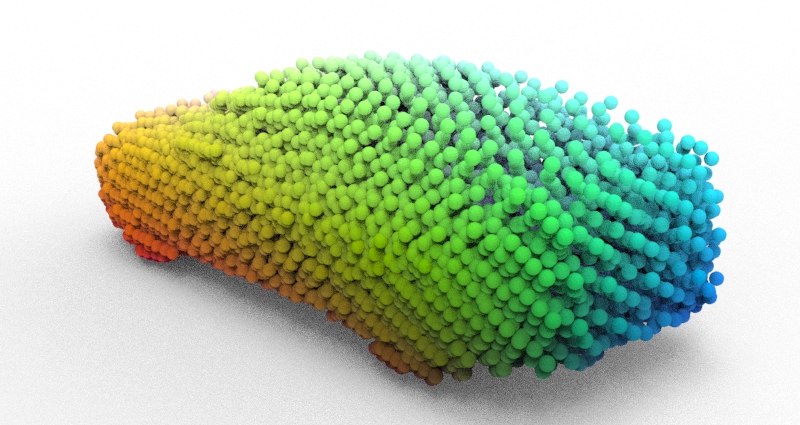}}
	\subfigure{
		\includegraphics[width=0.095\linewidth]{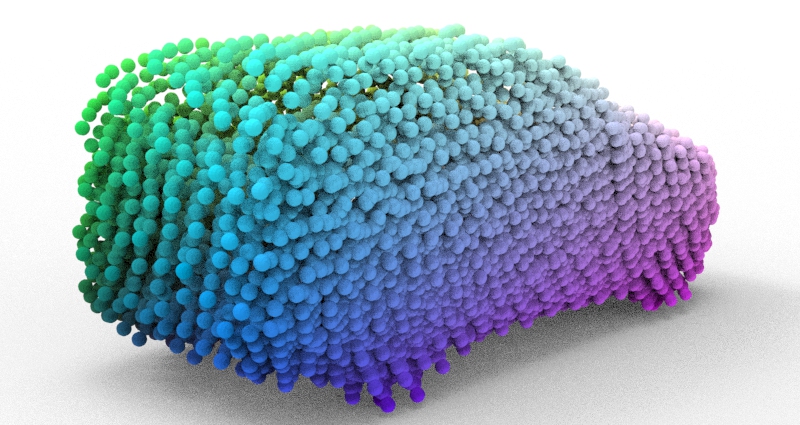}}
	\subfigure{
		\includegraphics[width=0.095\linewidth]{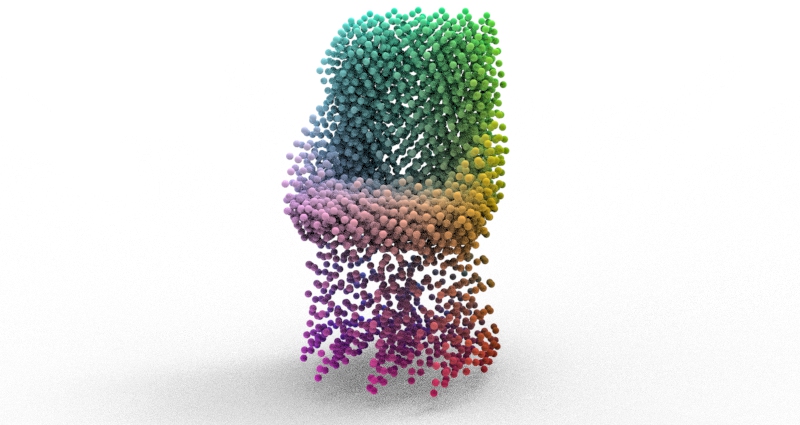}}
	\subfigure{
		\includegraphics[width=0.095\linewidth]{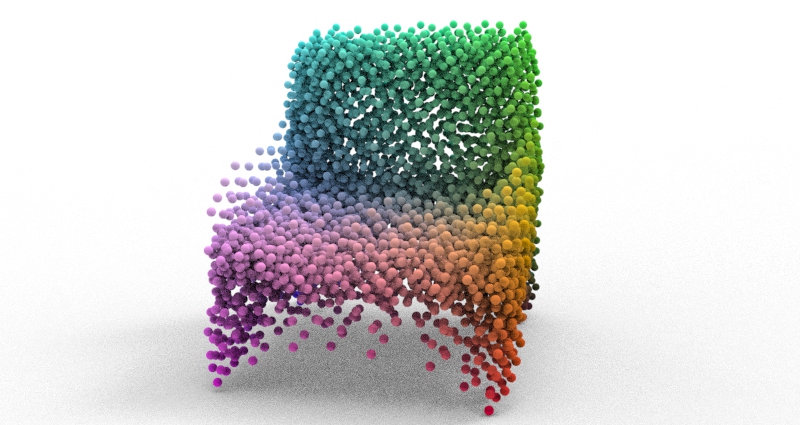}}
	\subfigure{
		\includegraphics[width=0.095\linewidth]{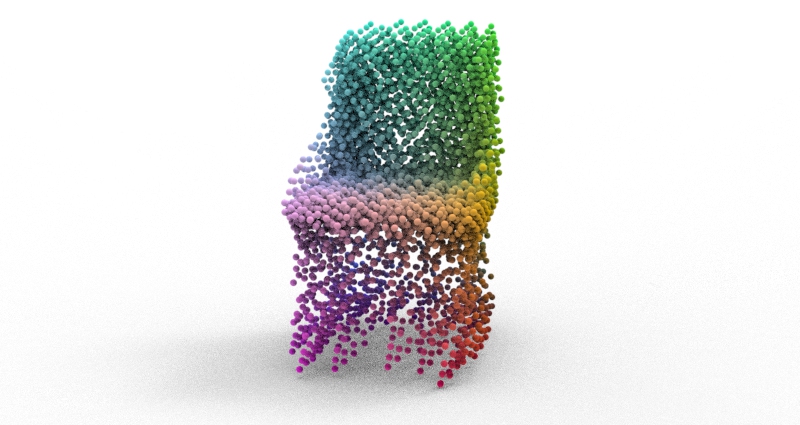}}
	\subfigure{
		\includegraphics[width=0.095\linewidth]{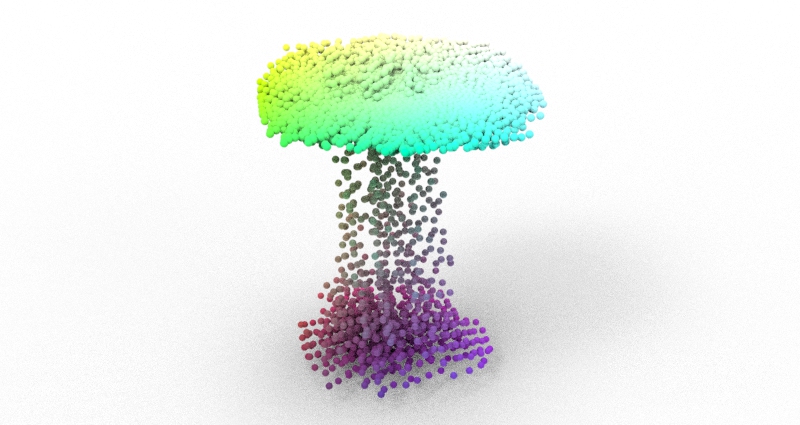}}
	\subfigure{
		\includegraphics[width=0.095\linewidth]{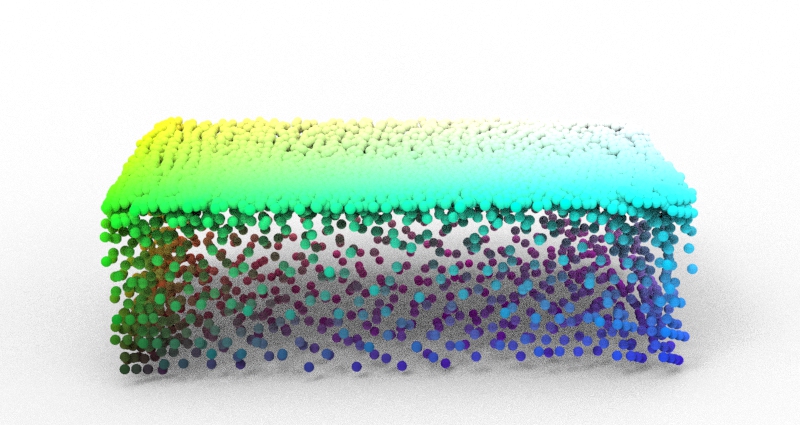}}
	\subfigure{
		\includegraphics[width=0.095\linewidth]{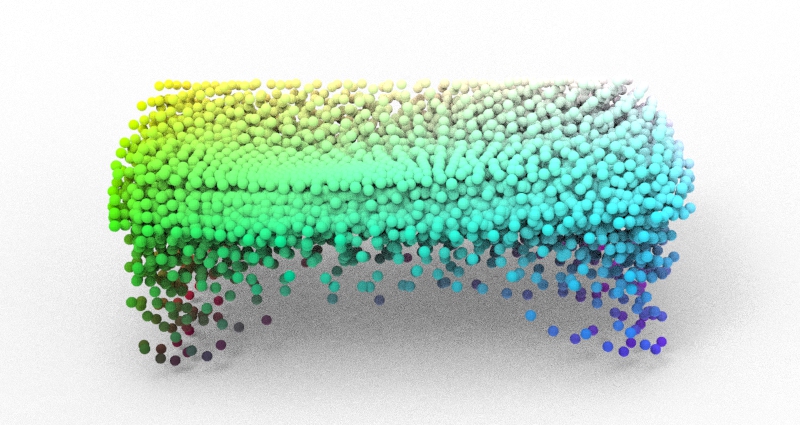}}
	\vspace{5mm}
        \\
	\subfigure{
		\includegraphics[width=0.095\linewidth]{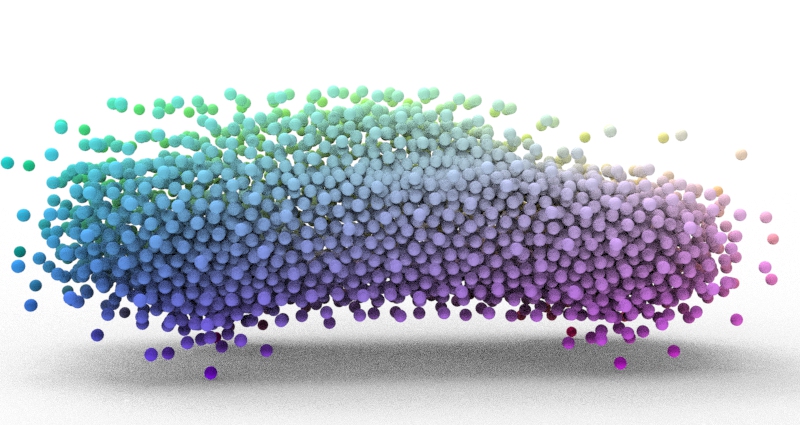}}
	\subfigure{
		\includegraphics[width=0.095\linewidth]{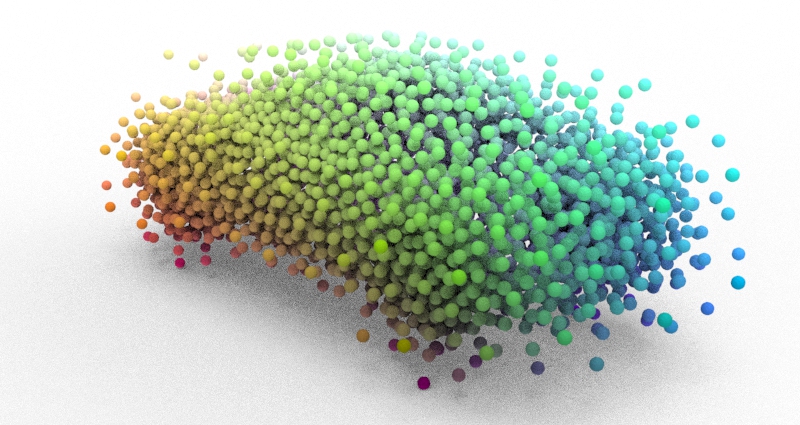}}
	\subfigure{
		\includegraphics[width=0.095\linewidth]{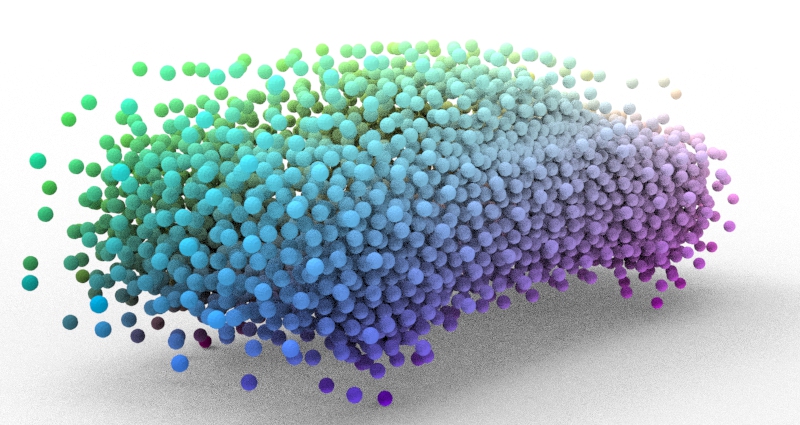}}
	\subfigure{
		\includegraphics[width=0.095\linewidth]{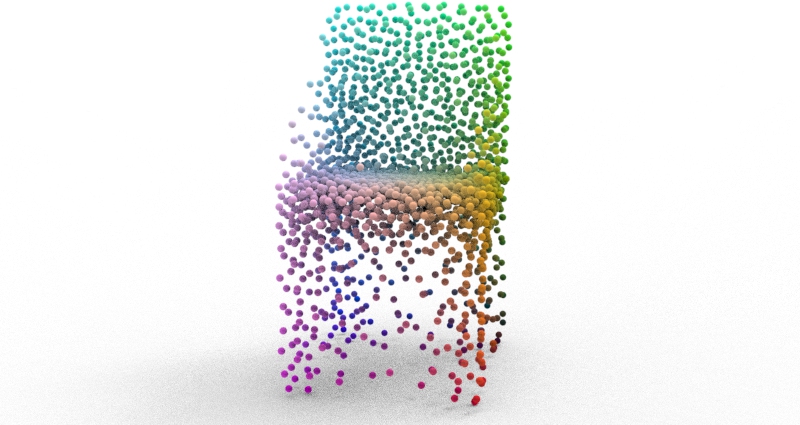}}
	\subfigure{
		\includegraphics[width=0.095\linewidth]{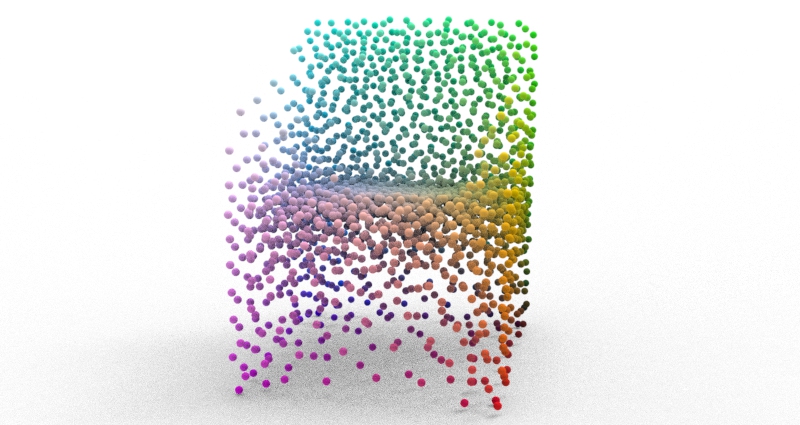}}
	\subfigure{
		\includegraphics[width=0.095\linewidth]{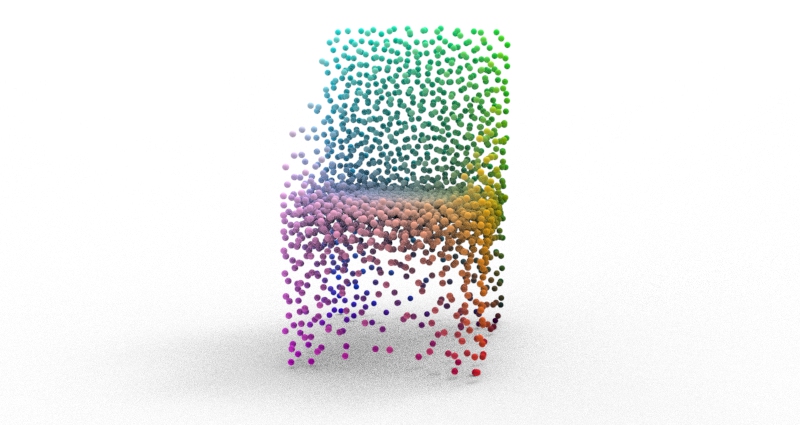}}
	\subfigure{
		\includegraphics[width=0.095\linewidth]{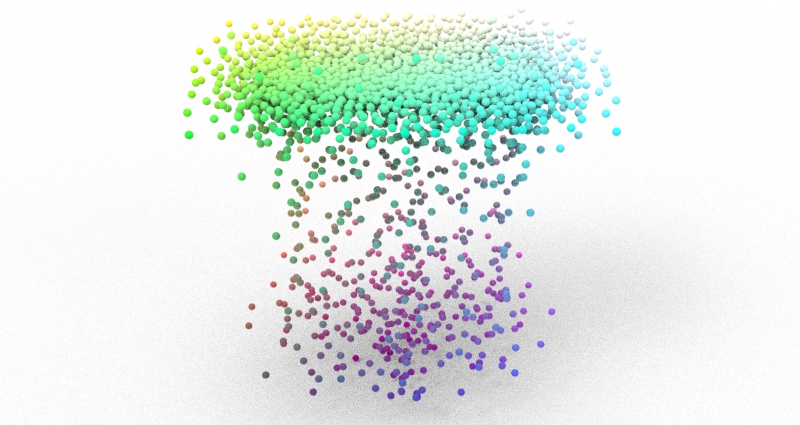}}
	\subfigure{
		\includegraphics[width=0.095\linewidth]{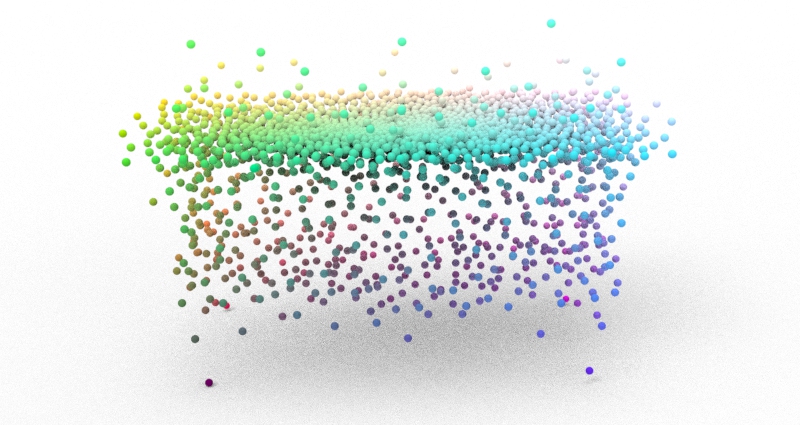}}
	\subfigure{
		\includegraphics[width=0.095\linewidth]{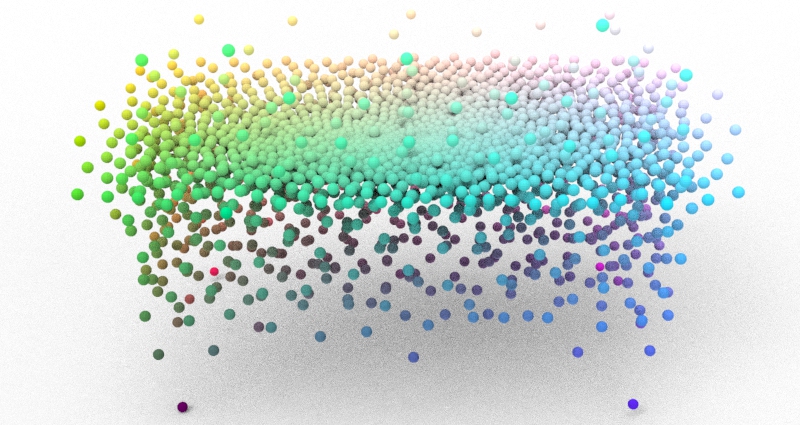}}
	\vspace{5mm}
        \\
	\subfigure{
		\includegraphics[width=0.095\linewidth]{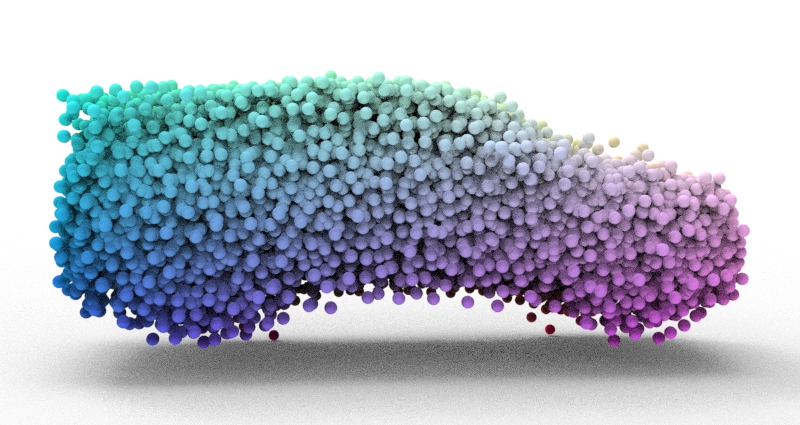}}
	\subfigure{
		\includegraphics[width=0.095\linewidth]{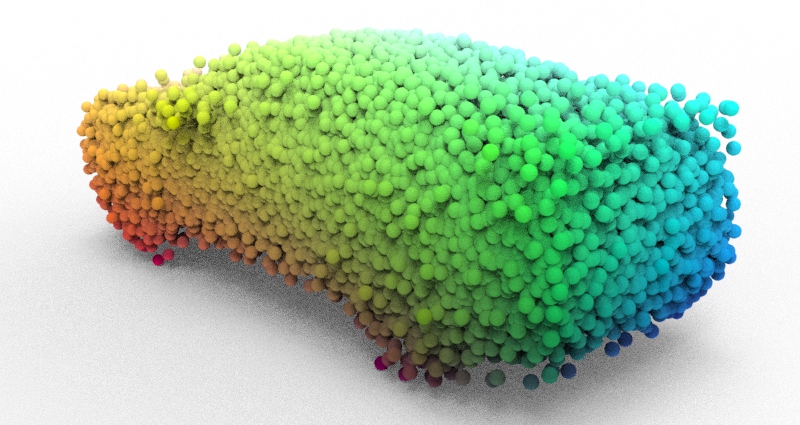}}
	\subfigure{
		\includegraphics[width=0.095\linewidth]{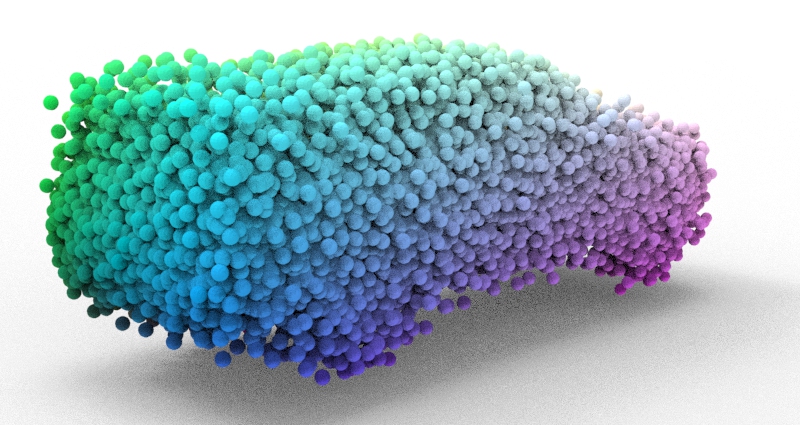}}
	\subfigure{
		\includegraphics[width=0.095\linewidth]{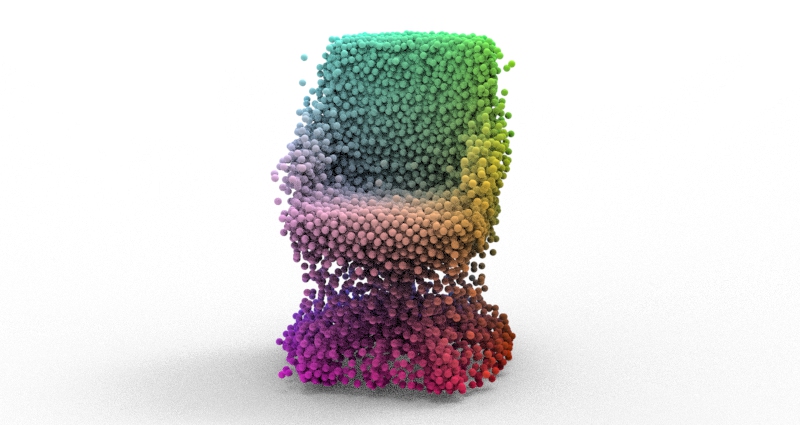}}
	\subfigure{
		\includegraphics[width=0.095\linewidth]{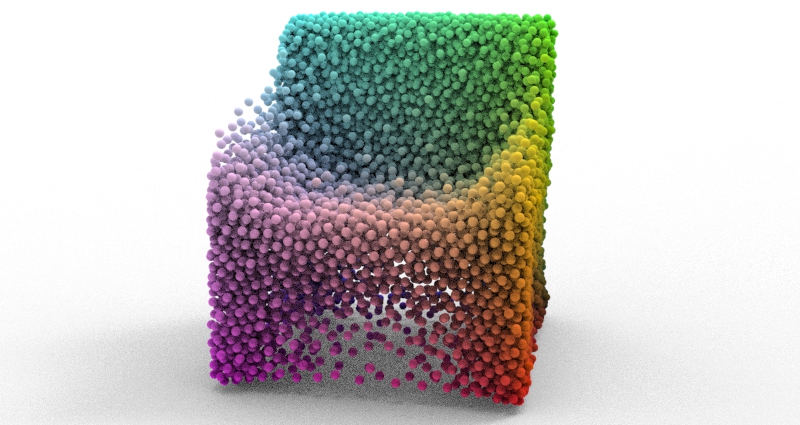}}
	\subfigure{
		\includegraphics[width=0.095\linewidth]{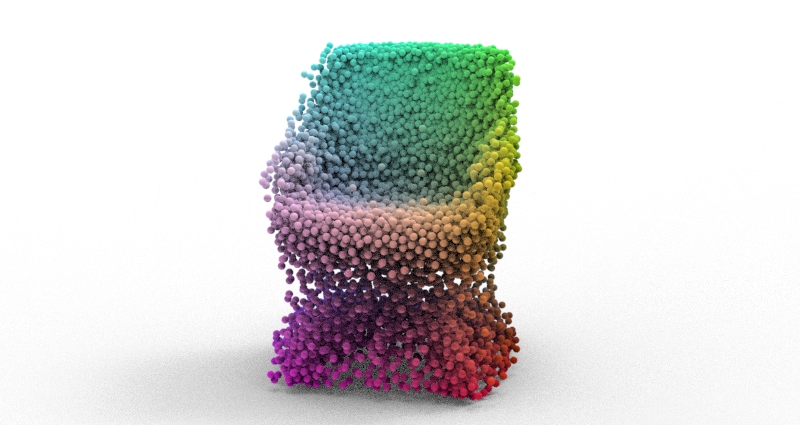}}
	\subfigure{
		\includegraphics[width=0.095\linewidth]{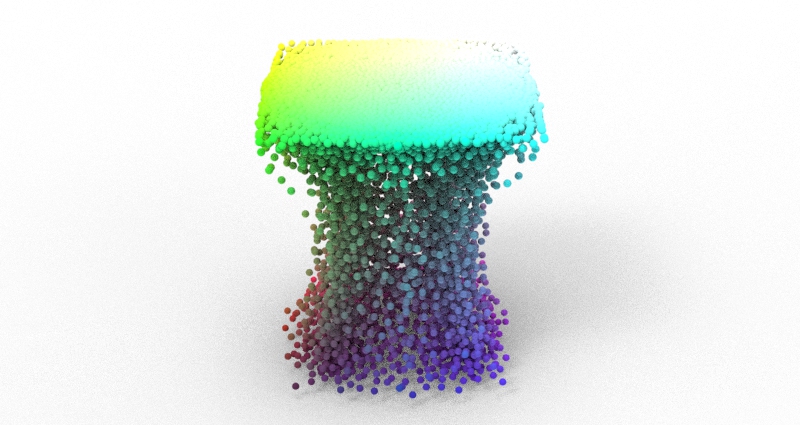}}
	\subfigure{
		\includegraphics[width=0.095\linewidth]{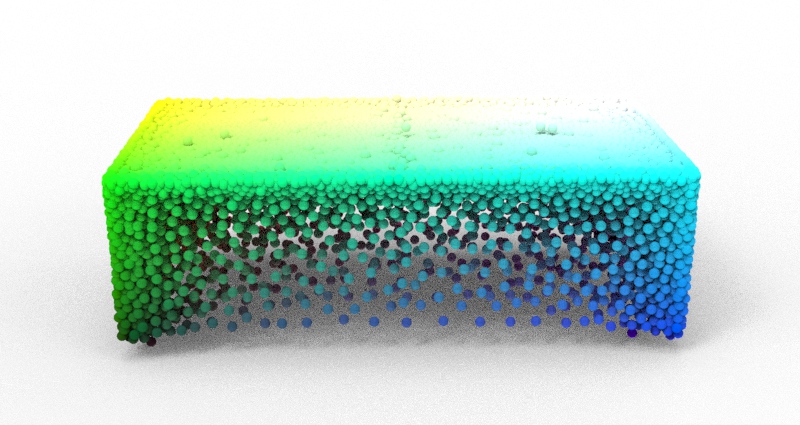}}
	\subfigure{
		\includegraphics[width=0.095\linewidth]{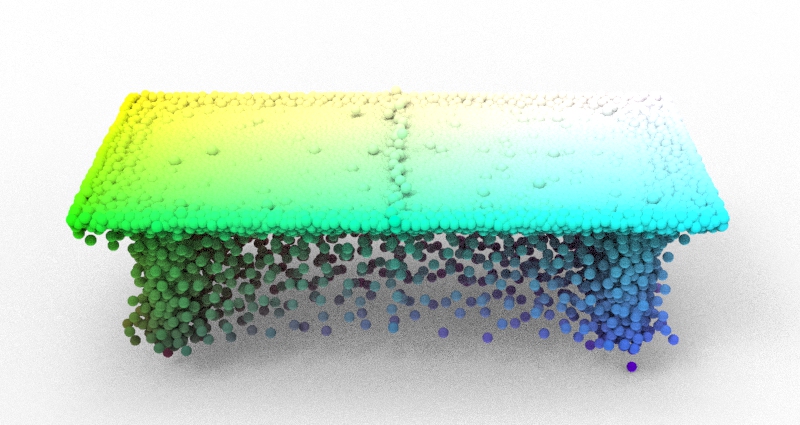}}
	\vspace{5mm}
        \\
        \subfigure{
		\includegraphics[width=0.095\linewidth]{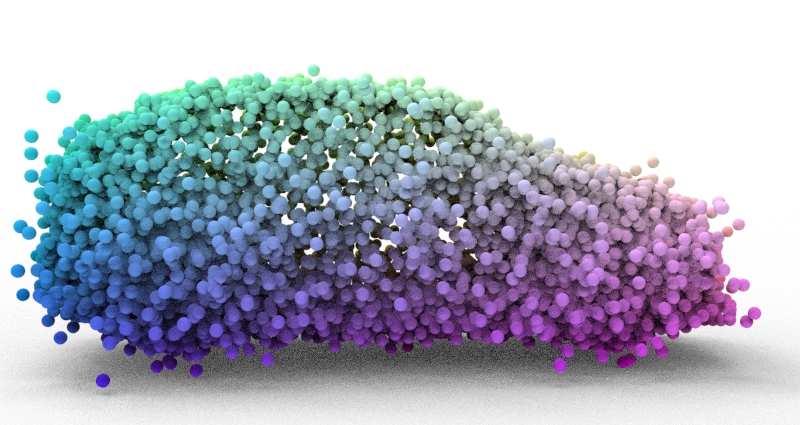}}
	\subfigure{
		\includegraphics[width=0.095\linewidth]{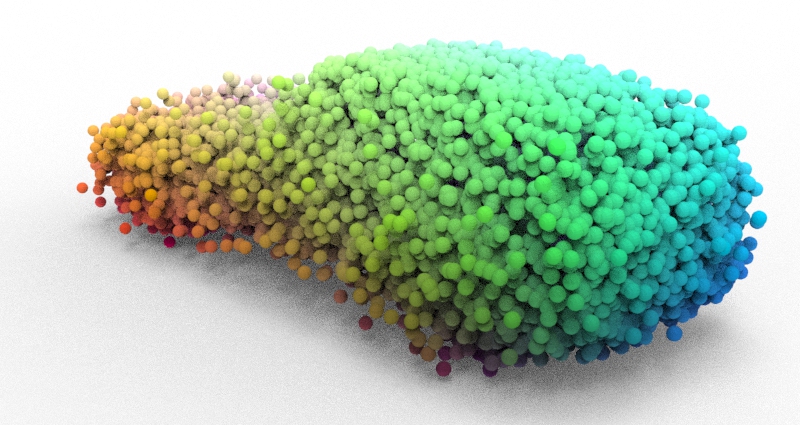}}
	\subfigure{
		\includegraphics[width=0.095\linewidth]{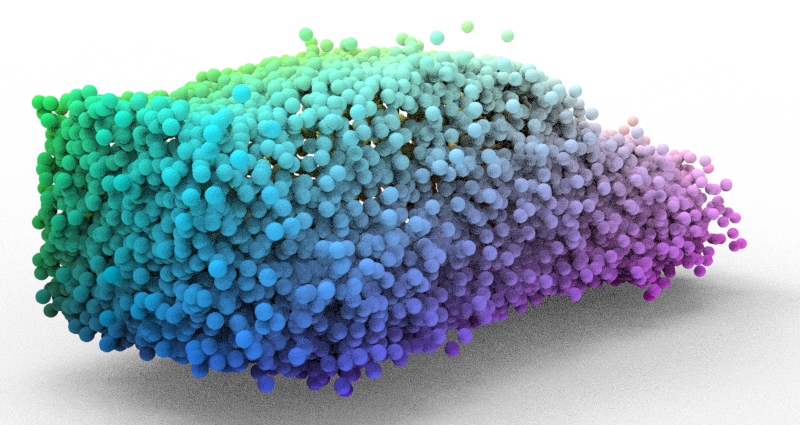}}
	\subfigure{
		\includegraphics[width=0.095\linewidth]{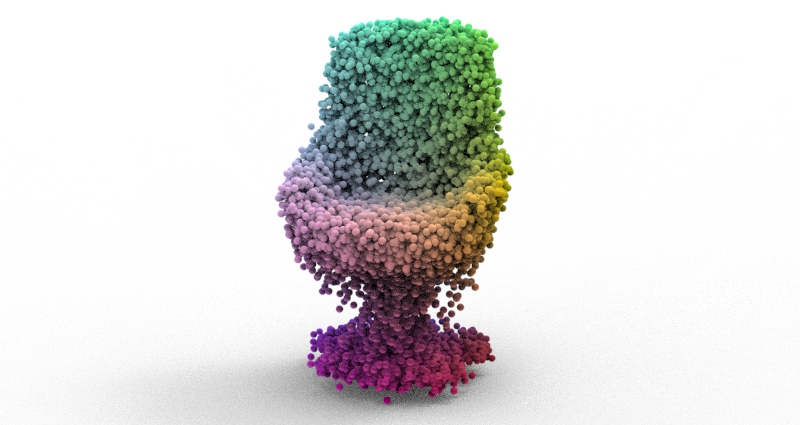}}
	\subfigure{
		\includegraphics[width=0.095\linewidth]{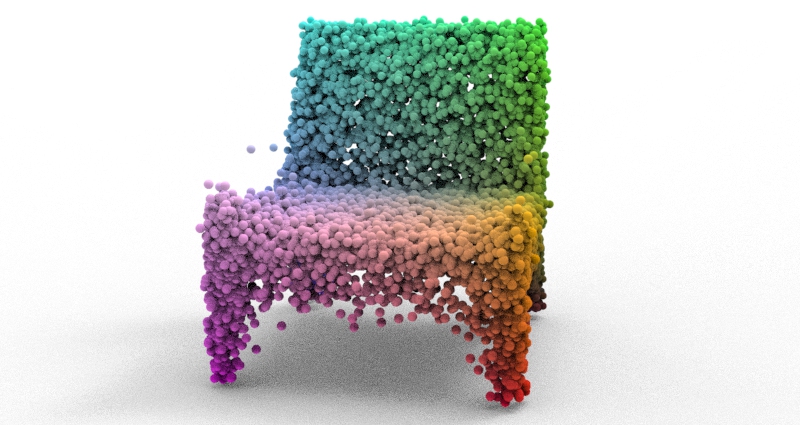}}
	\subfigure{
		\includegraphics[width=0.095\linewidth]{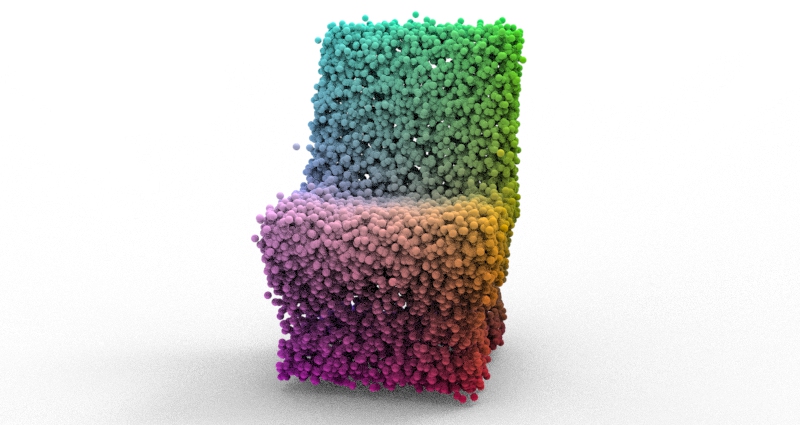}}
	\subfigure{
		\includegraphics[width=0.095\linewidth]{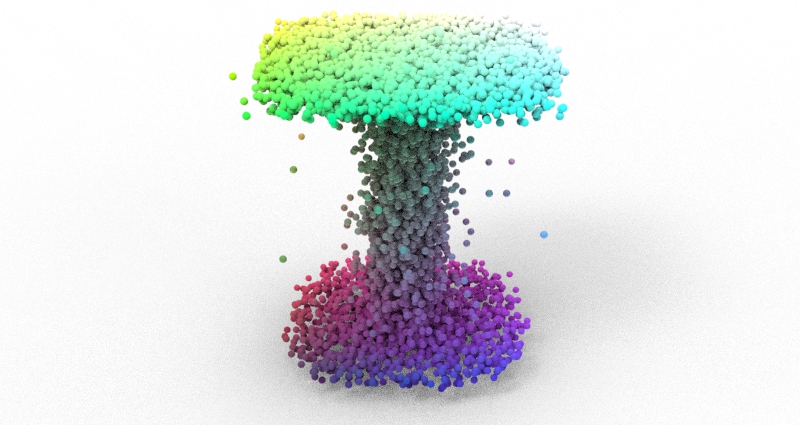}}
	\subfigure{
		\includegraphics[width=0.095\linewidth]{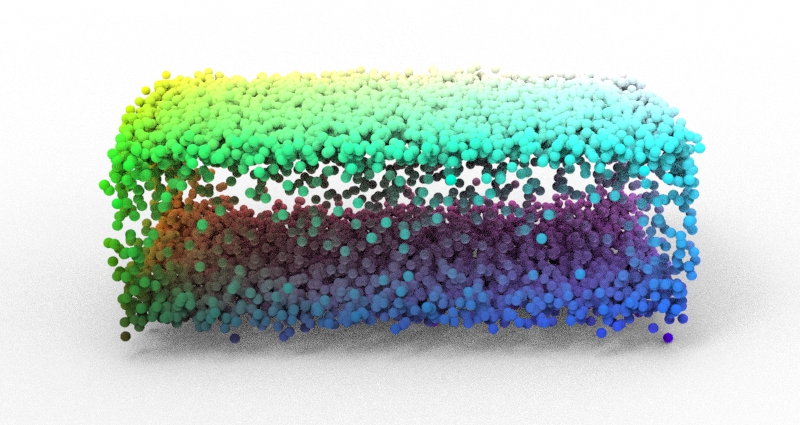}}
	\subfigure{
		\includegraphics[width=0.095\linewidth]{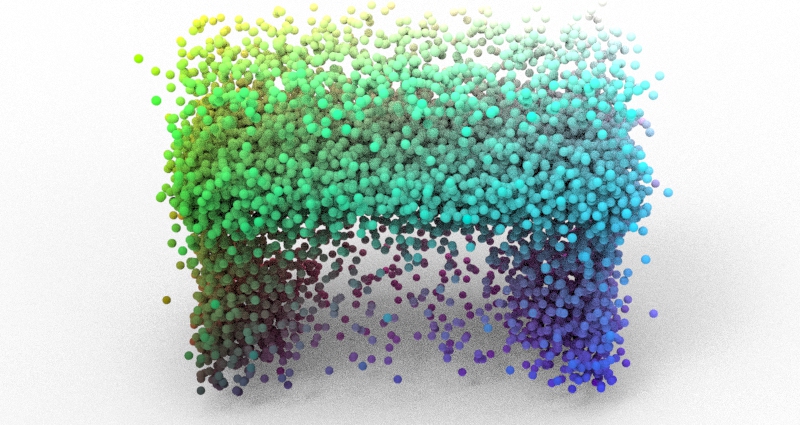}}
	\vspace{5mm}
        \\
        \setcounter{subfigure}{0}
	\subfigure[]{
		\includegraphics[width=0.095\linewidth]{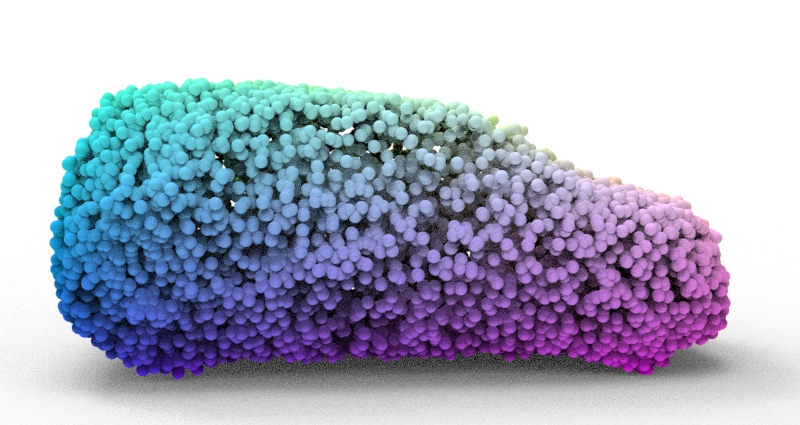}}
	\subfigure[]{
		\includegraphics[width=0.095\linewidth]{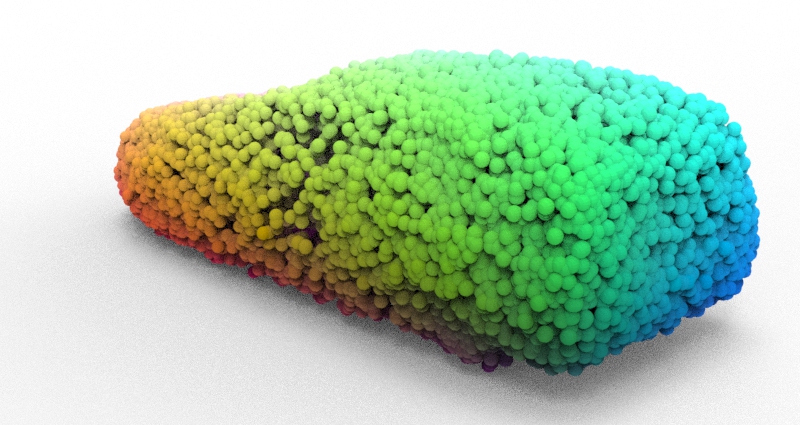}}
	\subfigure[]{
		\includegraphics[width=0.095\linewidth]{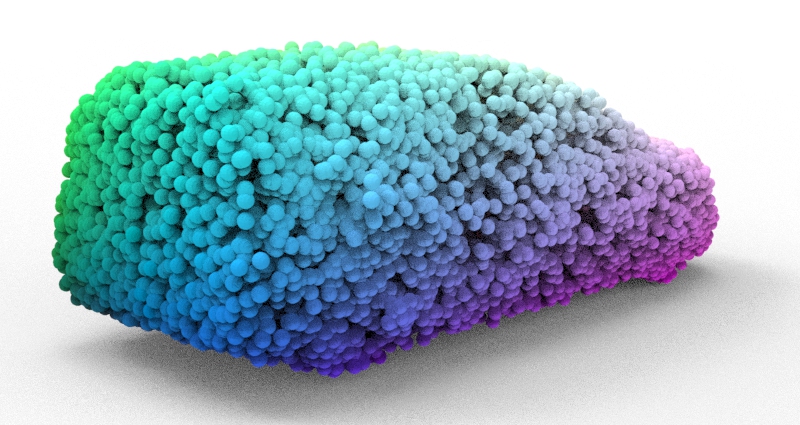}}
	\subfigure[]{
		\includegraphics[width=0.095\linewidth]{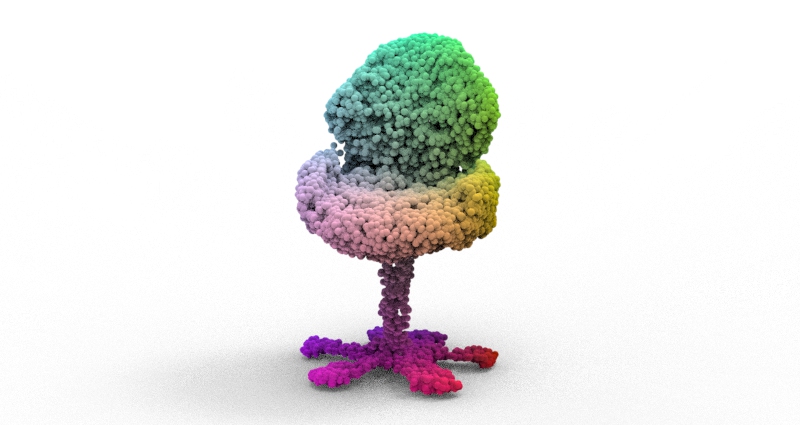}}
	\subfigure[]{
		\includegraphics[width=0.095\linewidth]{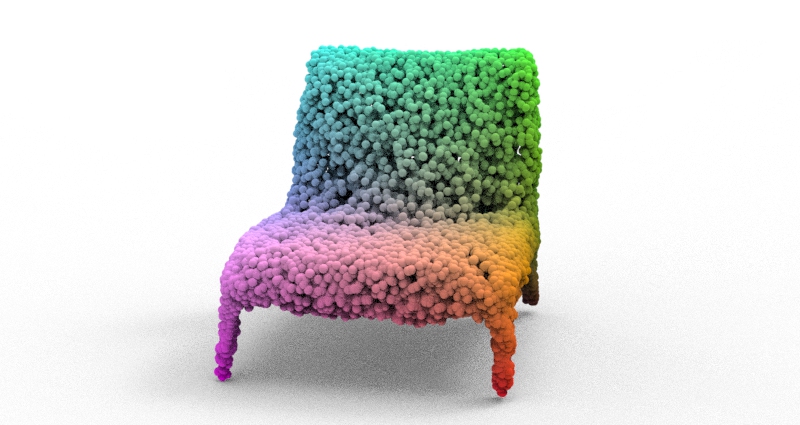}}
	\subfigure[]{
		\includegraphics[width=0.095\linewidth]{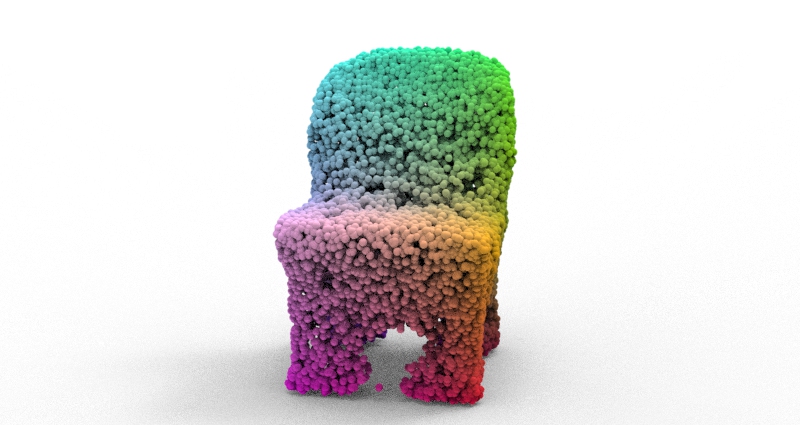}}
	\subfigure[]{
		\includegraphics[width=0.095\linewidth]{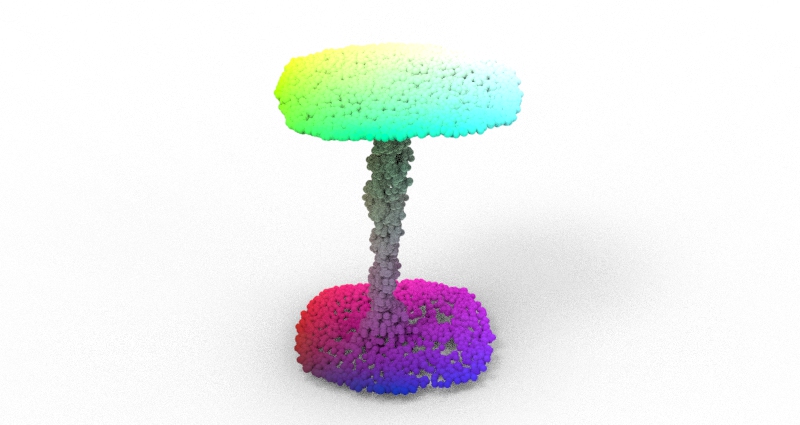}}
	\subfigure[]{
		\includegraphics[width=0.095\linewidth]{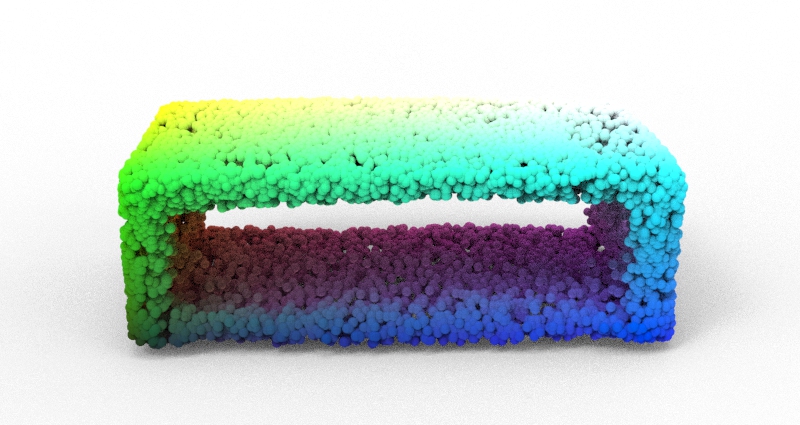}}
	\subfigure[]{
		\includegraphics[width=0.095\linewidth]{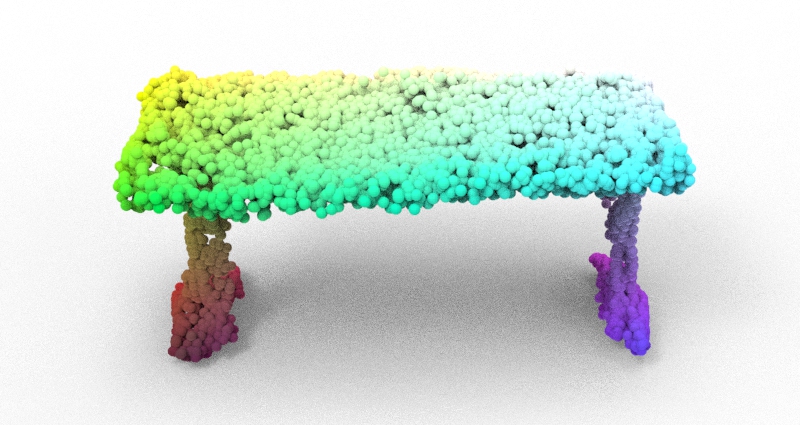}}
	\caption{Qualitative results on the real-world dataset. The first raw represents the input incomplete point cloud, while the second to the last rows correspond to the results of PCN \cite{pcn}, PCL2PCL \cite{pcl2plc}, ACL-SPC \cite{hong2023acl}, P2C \cite{cui2023p2c}, and our method, respectively. (a)$\sim$(c) Car. (d)$\sim$(f) Chair. (g)$\sim$(i) Table.}
	\label{visual_comparison_real}
\end{figure*}

\begin{table*}[htbp]
	\centering
	\renewcommand\arraystretch{1.25}
	\caption{\wltrev{Results on severely occluded data in terms of L2 CD\(\downarrow\) (scaled by \(10^{4}\)). The \textbf{bold} numbers represent the best results.}}
	\label{severe_occlusion}
	\begin{tabular}{l|c|c|c|c|c|c|c|c|c}
		\toprule[1.2pt]
		Methods             & Average & Plane & Cabinet & Car & Chair & Lamp	& Couch	& Table	& Watercraft \\ \hline 
             ACL-SPC \cite{hong2023acl}   & 25.04 & 10.47 & 25.63 & 11.33 & 25.35 & 61.68 & 26.93 & 23.71  & 15.29 \\
		P2C \cite{cui2023p2c}    & 18.67 & 5.62  & 16.05 & 16.00 & \textbf{15.68} & \textbf{36.29} & 25.81  &  19.06  & 14.83 \\ \hline
		Ours          & \textbf{16.93} & \textbf{4.34}  & \textbf{12.82} & \textbf{8.07} & 19.20  & 44.09 & \textbf{15.85} &  \textbf{18.84} & \textbf{12.28} \\ \bottomrule[1.2pt]
	\end{tabular}
\end{table*}

\begin{table*}[htbp]
	\centering
	\renewcommand\arraystretch{1.25}
	\caption{\wltrev{Results on long-tail data in terms of L2 CD\(\downarrow\) (scaled by \(10^{4}\)). The reported values to the left and right of the slash ``/" are CD on the common set and long-tail set, respectively.}}
	\label{longtail}
	\begin{tabular}{l|c|c|c|c|c|c|c|c}
		\toprule[1.2pt]
		Methods              & Plane & Cabinet & Car & Chair & Lamp	& Couch	& Table	& Watercraft \\ \hline 
             ACL-SPC \cite{hong2023acl}  & 0.74/10.84 & 17.16/44.00 & 6.40/8.99 & 7.46/14.71 & 12.23/50.15 & 9.08/37.38 & 14.45/60.62  & 6.48/23.54 \\
		P2C \cite{cui2023p2c}    & 0.53/4.91 & 8.75/16.58 & 4.01/7.54 & 5.60/35.13 & 6.16/21.83 & 9.09/104.86 & 3.98/57.88  & 3.67/13.15 \\ \hline
		Ours           & 0.87/5.97 & 11.17/12.88 & 6.78/14.39 & 5.39/16.82 & 9.36/13.82 & 9.22/11.46 & 7.67/41.41  & 3.57/11.23 \\ \bottomrule[1.2pt]
	\end{tabular}
\end{table*}

\wltrev{As shown in the last two rows of Table \ref{synthetic_results}, the results of our method trained separately for each category are generally better than training them together. However, the latter also outperforms P2C and ACL-SPC.} 
\wltrev{Table \ref{pre_cov} shows that our method achieves the best Precision across most categories, and outperforms self-supervised methods in terms of average Precision. Although it falls slightly behind ACL-SPC in Coverage, the difference remains minimal. \wltrevRe{Specifically, our Coverage is slightly higher than ACL-SPC by 0.93 (51.38\%), while ACL-SPC's Precision is higher than ours by 7.1 (112.16\%)}. \wltrevReRe{In addition, relying only on Coverage or Precision is insufficient to comprehensively measure the quality of predicted results. For example, if the predicted points fill the whole 3D space, the Coverage value would be zero, and if all the predicted points are concentrated at the position of a single point in the ground truth, the Precision value would be zero. In both cases, the predicted results would be extremely poor. Thus, a comprehensive quantitative evaluation requires combining Coverage and Precision. Therefore, we believe that the slight disadvantage of Coverage does not undermine the overall superiority of our method.}}

\wltrev{Table \ref{other_results} indicates that our method leads on the PCN and CRN results,  with the average CD surpassing the second place by 6.04 and 6.09, respectively. However, the performance of our method and ACL-SPC shows a significant decline on the 3D-EPN dataset. To simulate the human perspective, we set the elevation angle for scanning objects between -30 and 30 degrees when generating our dataset, whereas the partial point clouds in 3D-EPN are scanned from arbitrary angles. This difference in viewpoints introduces a domain gap between the datasets, resulting in the underperformance of our method and ACL-SPC. Note that the performance of P2C increases on the 3D-EPN dataset, indicating that it is more robust to viewpoint variations of partial shape scan.}

\wltrev{Moreover, as shown in Table \ref{synthetic_results}, our method achieves favorable results in the unpaired case, with the average CD significantly outperforming PCL2PCL and USSPA. However, the results slightly lag behind Ours-S model. We believe that, on the one hand, our PRN effectively learns the common 3D structural features within each object and category, thereby reducing the impact of the unpaired complete point cloud. On the other hand, unlike unpaired methods, our MAN leverages a 2D network as the discriminator. It is well-known that 2D neural networks generally possess stronger learning capabilities than 3D MLPs, and our method is able to effectively utilize this advantage.}

\wltrevRe{Furthermore, Table \ref{synthetic_results} shows that after adding our rendering loss and MAN module to the P2C model, the average CD decreases by 1.2 and the CD values across six categories also decline. This demonstrates the versatility of our depth map-based strategies, which is also a significant contribution of our work. On the other hand, the average CD of this method is still higher than that of our method by 2.0, further illustrating the superiority of our complete framework over the self-supervised methods.}

Figure \ref{visual_comparison} illustrates a comparison of the visualization results between our method and other benchmarks. We can see that MAL-UPC can infer a more reasonable global shape while better preserving the incomplete parts. For example, the four legs of the input chair point cloud are all thin cylindrical shapes, while self-supervised or even supervised methods reconstruct chair legs as thick rectangular prisms, whereas MAL-UPC preserves its original shape well. In addition, it is obvious that the noise points generated by our method are significantly fewer than those produced by other methods. Self-supervised benchmarks and most supervised approaches generate a large number of outlier points outside the edges of objects, whereas our method has almost no outlier points.

\subsubsection{Results on real-world scans}

In order to validate the robustness and effectiveness of our method in real-world environments, we evaluate its performance on two challenging datasets: ScanNet and KITTI. As demonstrated in Table \ref{real_table}, compared to ACL-SPC and P2C, our MAL-UPC yields the best UHD on all three categories and the best UCD on the car and table categories. Our method only slightly falls behind the second-best approach in terms of UCD on the chair category. Moreover, regarding the results on the car category, MAL-UPC even surpasses the unpaired and supervised methods by a large margin. 
Figure \ref{visual_comparison_real} shows the qualitative results of our model and other state-of-the-art benchmarks. Similar to its superior performance on synthetic data, our method predicts more reasonable shapes and better preserves the original inputs, while generating significantly fewer noise points.

\begin{figure}[htbp]
	\centering  
	\subfigbottomskip=1pt 
	\subfigure{
		\includegraphics[width=0.19\linewidth]{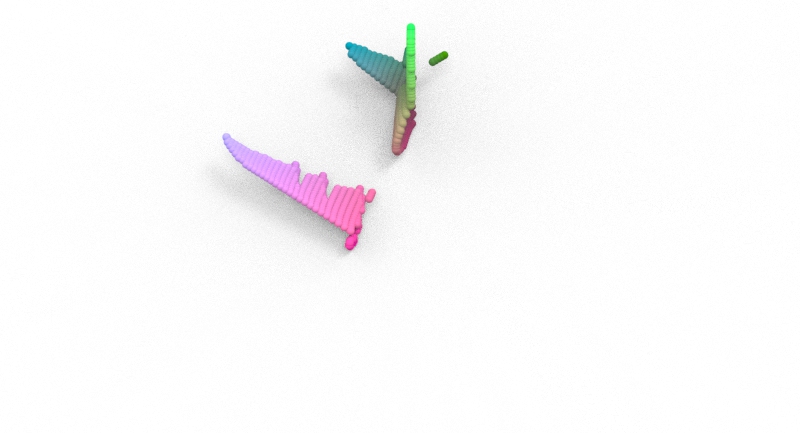}}
	\hspace{-3mm}
	\subfigure{
		\includegraphics[width=0.19\linewidth]{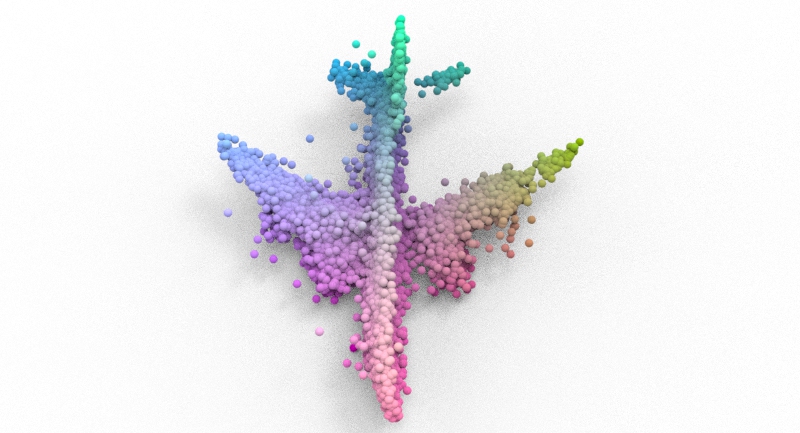}}
	\hspace{-3mm}
	\subfigure{
		\includegraphics[width=0.19\linewidth]{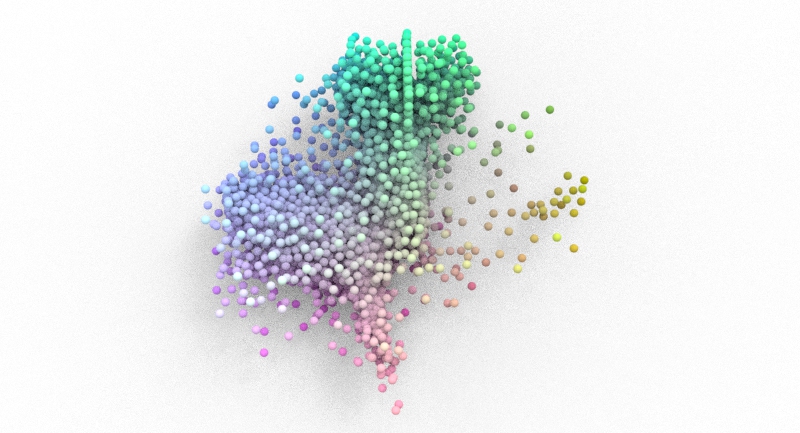}}
	\hspace{-3mm}
	\subfigure{
		\includegraphics[width=0.19\linewidth]{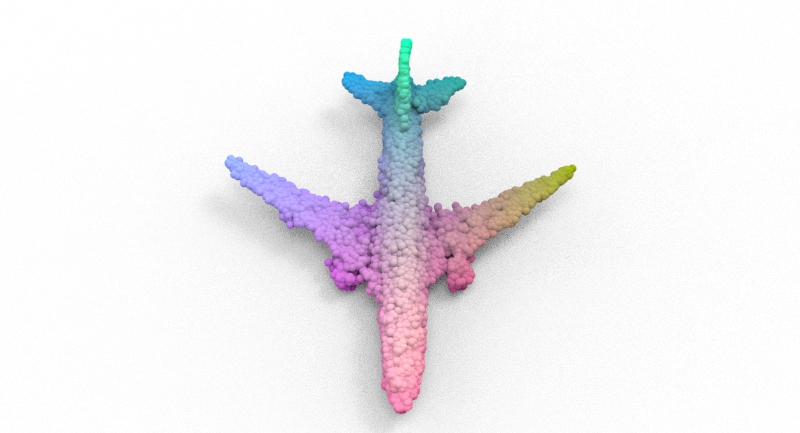}}
	\hspace{-3mm}
	\subfigure{
		\includegraphics[width=0.19\linewidth]{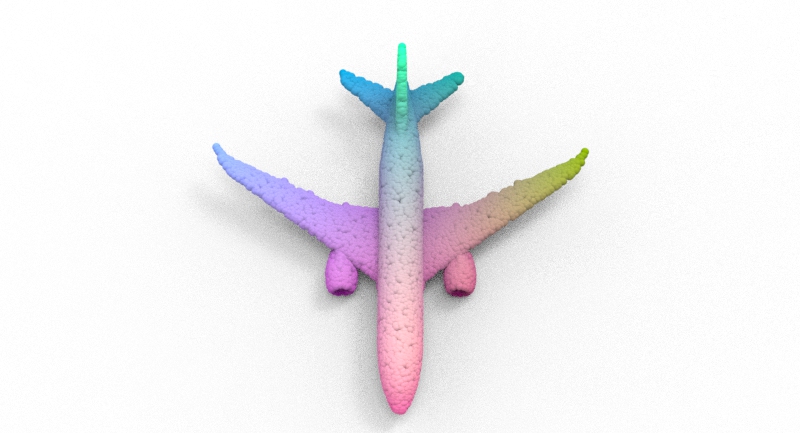}}
	\\
	\subfigure{
		\includegraphics[width=0.19\linewidth]{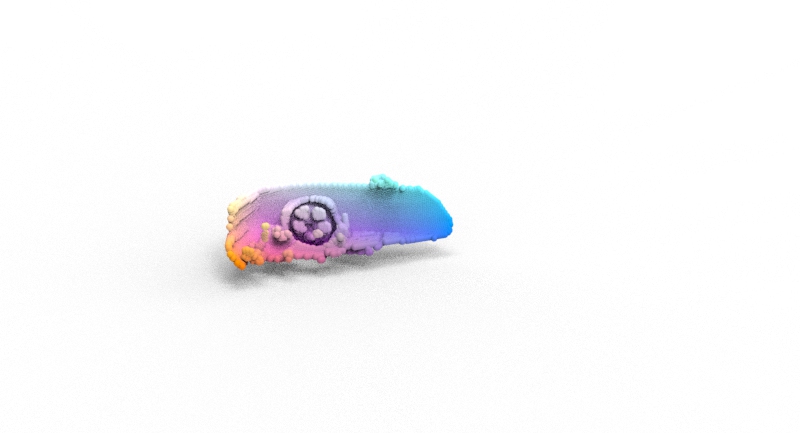}}
	\hspace{-3mm}
	\subfigure{
		\includegraphics[width=0.19\linewidth]{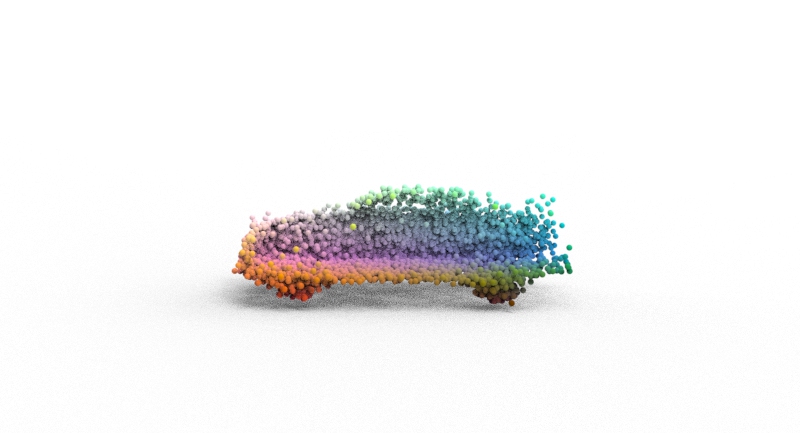}}
	\hspace{-3mm}
	\subfigure{
		\includegraphics[width=0.19\linewidth]{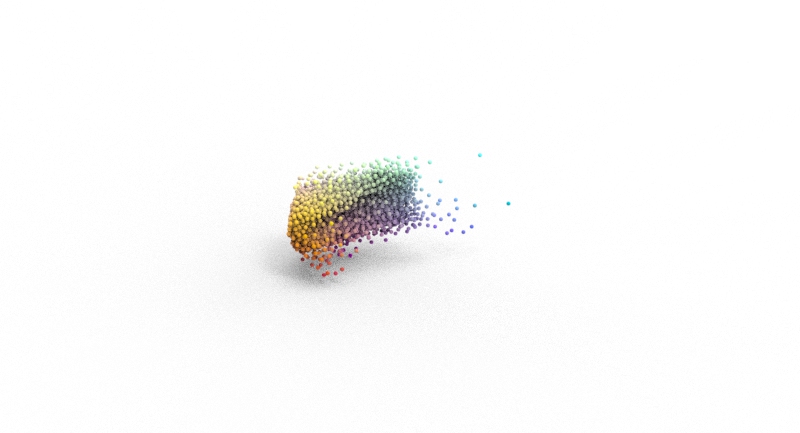}}
	\hspace{-3mm}
	\subfigure{
		\includegraphics[width=0.19\linewidth]{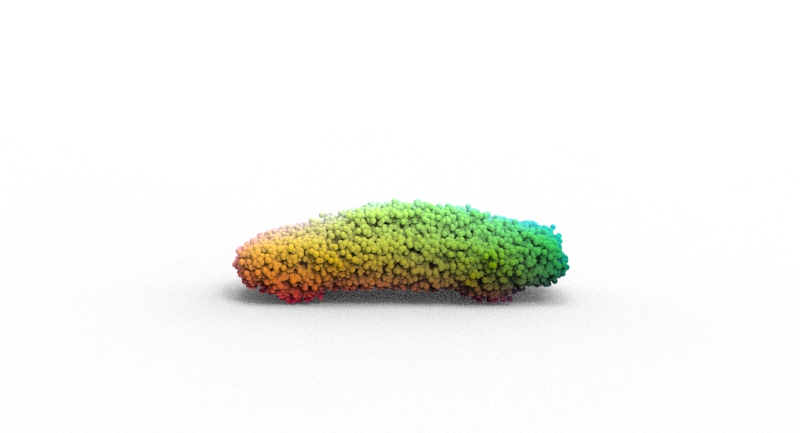}}
	\hspace{-3mm}
	\subfigure{
		\includegraphics[width=0.19\linewidth]{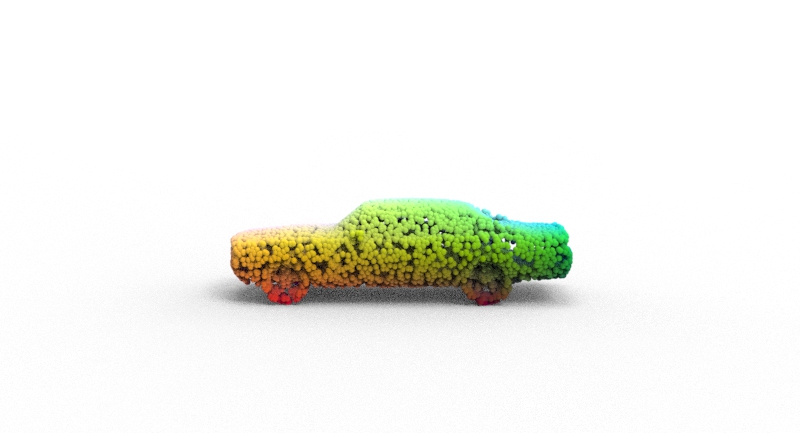}}
	\\
	\setcounter{subfigure}{0}
	\subfigure[]{
		\includegraphics[width=0.19\linewidth]{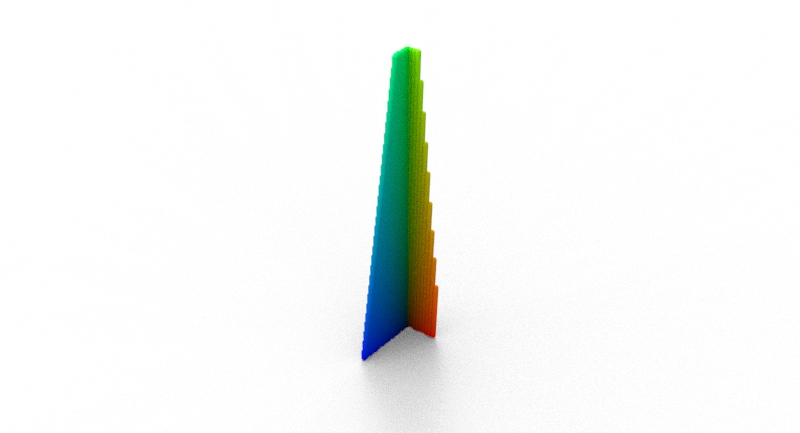}}
	\hspace{-3mm}
	\subfigure[]{
		\includegraphics[width=0.19\linewidth]{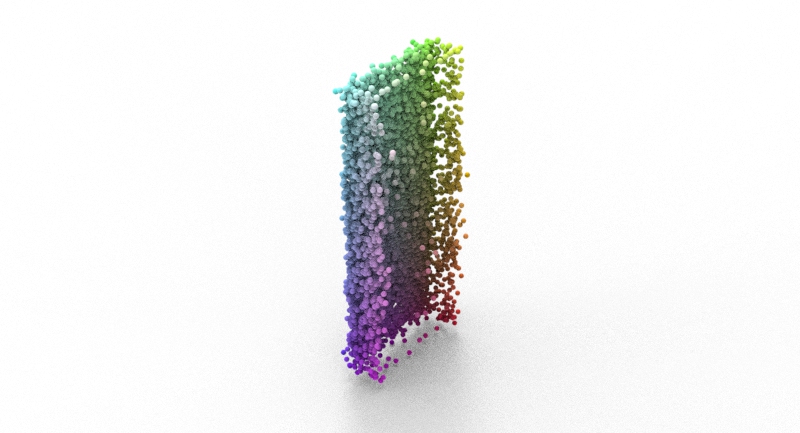}}
	\hspace{-3mm}
	\subfigure[]{
		\includegraphics[width=0.19\linewidth]{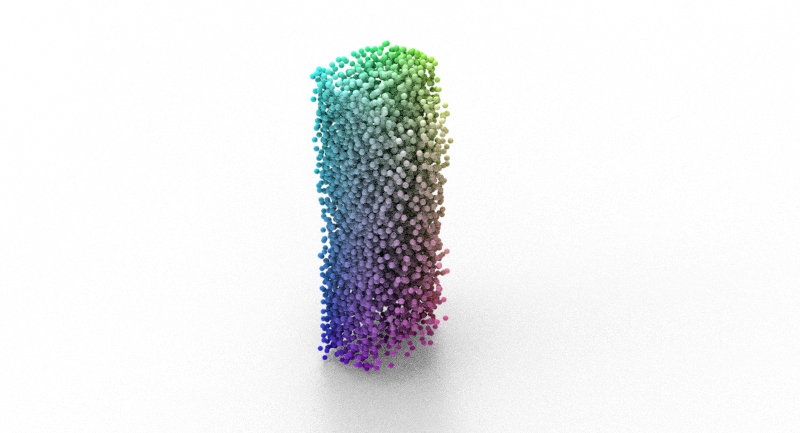}}
	\hspace{-3mm}
	\subfigure[]{
		\includegraphics[width=0.19\linewidth]{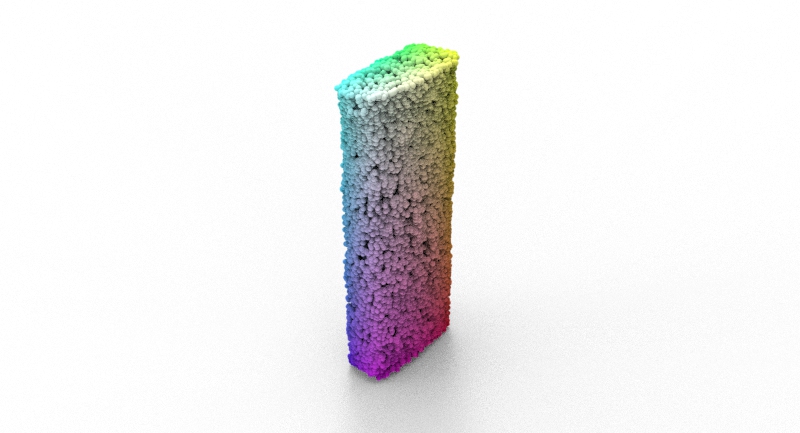}}
	\hspace{-3mm}
	\subfigure[]{
		\includegraphics[width=0.19\linewidth]{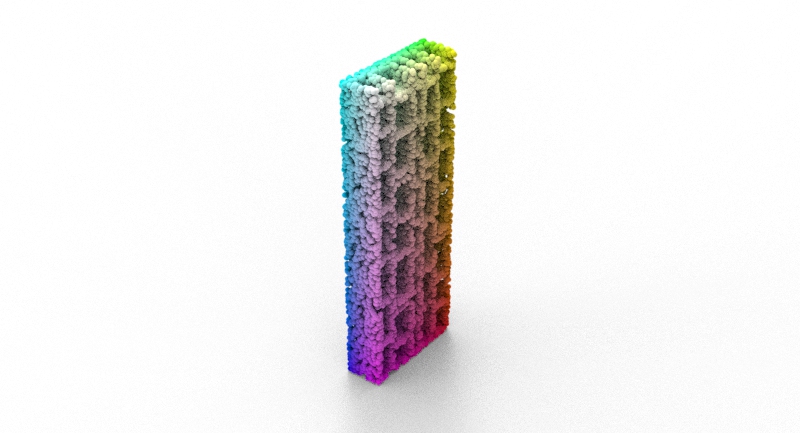}}

	\caption{Comparison of visual results on the severely occluded dataset. (a) Input partial points. (b) P2C \cite{cui2023p2c}. (c) ACL-SPC \cite{hong2023acl}. (d) Ours. (e) Ground truth.}
	\label{hard_visual}
\end{figure}

\subsubsection{\wltrev{Results on severely occluded objects}}
\wltrev{To assess our model's ability to complete objects with severe occlusions, following \cite{pointr}, we remove half of the points that are furthest from the camera in the partial scans and then use these data to train and test our model. As indicated in Table \ref{severe_occlusion}, the performance of the three methods significantly decreases when completing objects with severe occlusions. However, our method still leads the others, achieving the best results in terms of average CD and the CD across six categories. The visual results in Figure \ref{hard_visual} demonstrate that our method can still infer the complete shape under extreme occlusion conditions. For example, even when only parts of an airplane's wing and tail are visible, our method can still reconstruct the fuselage.  We believe this is because our model learns category-specific geometric patterns to complete missing parts.}

\subsubsection{\wltrev{Results on long-tail objects}}
\wltrev{Our model takes advantage of the common geometries within each category. To test its ability to reconstruct shapes with rare structures, we select long-tail objects of each category for experiments. Specifically, we use a pre-trained point cloud reconstruction model to generate a global feature for each object, and then calculate the cosine similarity between each global feature and others within the same category. The ten samples with the lowest average similarity are selected as the long-tail set for that category, while the ten with the highest average similarity form the common set, representing objects with the most typical geometric structures.  As Table \ref{longtail} demonstrates, the performance of the three methods drops significantly on the long-tail set, as they rely on common geometric patterns to infer missing structures without supervision from complete point clouds. In addition, our method demonstrates stable performance on the cabinet and couch categories, while on the table category, the performance difference between the two sets is the largest, reaching nearly 34.}

\begin{table}[htbp]
	\centering
	\renewcommand\arraystretch{1.25}
	\caption{\wltrev{Quantitative results on unseen categories. ``Major Categ." and ``Minor Categ." represent the two categories in ShapeNet-34 with the largest and smallest number of objects, respectively. The \textbf{bold} numbers represent the best results.} }
	\label{unseen_res}
	\begin{tabular}{c|c|c c|c c}
		\toprule[1.2pt]
		 \multirow{2}*{Methods} & \multirow{2}*{Average} & \multicolumn{2}{c|}{Major Categ.} & \multicolumn{2}{c}{Minor Categ.}                   \\  \cline{3-6}
           ~ & ~ & bowl & washer & cap & keyboard \\ \hline
		ACL-SPC \cite{hong2023acl}	& 230.54 & 59.79 & 431.94 & 132.45 & 240.59  \\ 
            P2C \cite{cui2023p2c}	& 119.96 & 52.81 & 249.57 & \textbf{83.03} & 20.66  \\ \hline
            Ours	& \textbf{98.89} & \textbf{49.84} & \textbf{196.15} & 83.84 & \textbf{13.02}   \\
        \bottomrule[1.2pt]
	\end{tabular}
\end{table}

\subsubsection{\wltrev{Results on novel categories}}
\wltrev{To evaluate the generalization ability of our model, we train it on all eight categories and then test it on novel categories. Specifically, the testing data are derived from ShapeNet-34 \cite{pointr}, which comprises 34 categories from ShapeNet. As shown in Table \ref{unseen_res}, the unsupervised methods generally perform poorly on the unseen dataset, indicating their limitations in generalization. This is because, without complete shapes as supervision, these approaches can only capture some typical geometric structures within the training categories. Even so, our method outperforms ACL-SPC and P2C and demonstrates particularly good performance on the keyboard category.}

\begin{table}[htbp]
	\centering
    \renewcommand\arraystretch{1.25}
	\caption{ Comparison of FLOPs and model size.}
	\label{flops_comparison}
	\begin{tabular}{l|c|c|c|c}
		\toprule[1.2pt]
		Methods             & USSPA \cite{ma2023symmetric} & ACL-SPC \cite{hong2023acl}& P2C \cite{cui2023p2c}  & Ours \\ \hline  
		FLOPs (G)      &  45.28  & 0.59 & 1.79 & 15.10 \\   
		Model size (M)    & 75.06 & 198.87 & 427.35 & 94.79 \\ \bottomrule[1.2pt]
	\end{tabular}
\end{table}

\subsubsection{Comparison of computational complexity and model size} 
Table \ref{flops_comparison} provides a comprehensive comparison of the computational complexity, measured in Floating Point Operations (FLOPs), and the model size for our proposed method against several state-of-the-art approaches: USSPA, ACL-SPC, and P2C. Our model demonstrates a well-balanced approach in terms of both computational complexity and model size. With 15.10G FLOPs, it is considerably more efficient than USSPA. In terms of model size, our method, with 94.79M parameters, remains manageable and smaller than both ACL-SPC and P2C. This balance makes our model an attractive option for practical applications where both computational resources and memory capacity are constrained.

\begin{figure}[htbp]
	\centering  
	\subfigbottomskip=1pt 
	\subfigure{
		\includegraphics[width=0.26\linewidth]{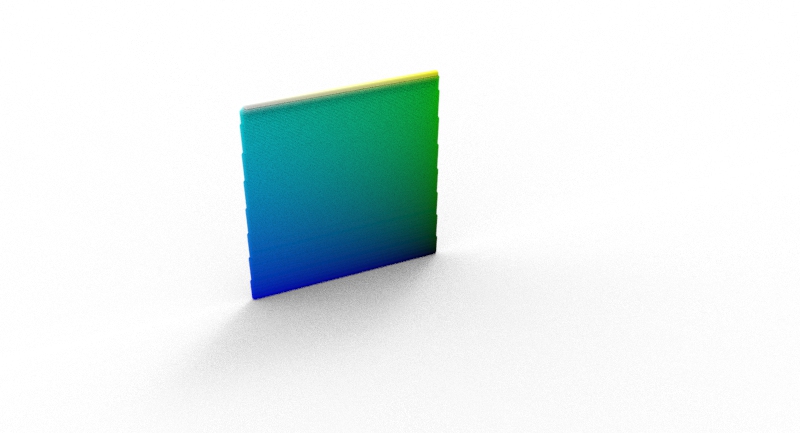}}
        \subfigure{
		\includegraphics[width=0.26\linewidth]{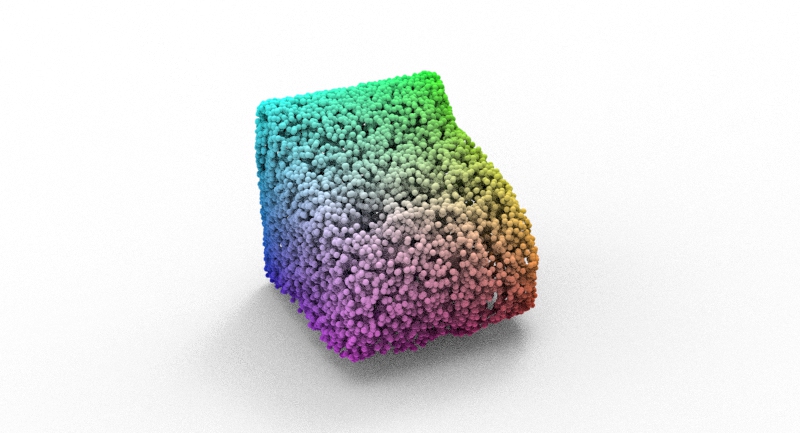}}
        \hspace{2mm}
        \subfigure{
		\includegraphics[width=0.26\linewidth]{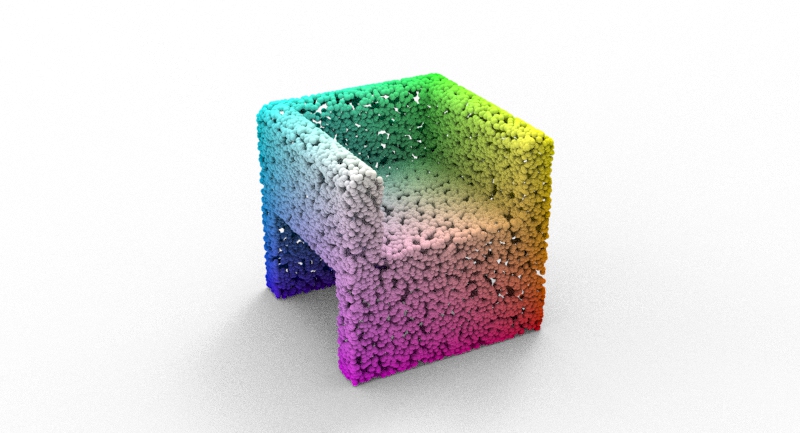}}
        \\
        \subfigure{
		\includegraphics[width=0.26\linewidth]{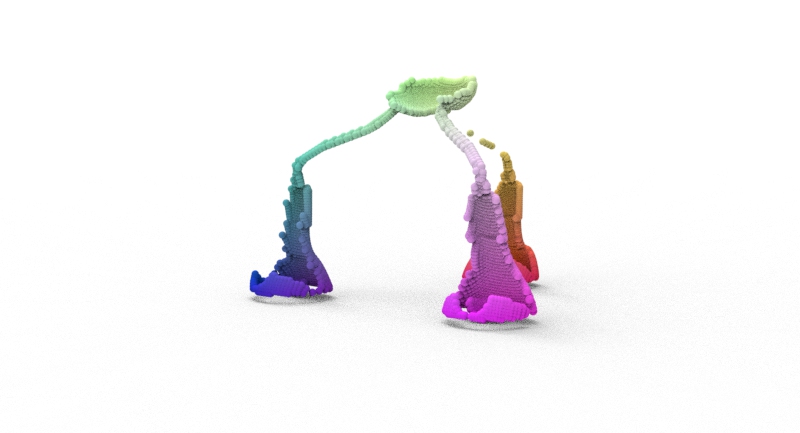}}
        \subfigure{
		\includegraphics[width=0.26\linewidth]{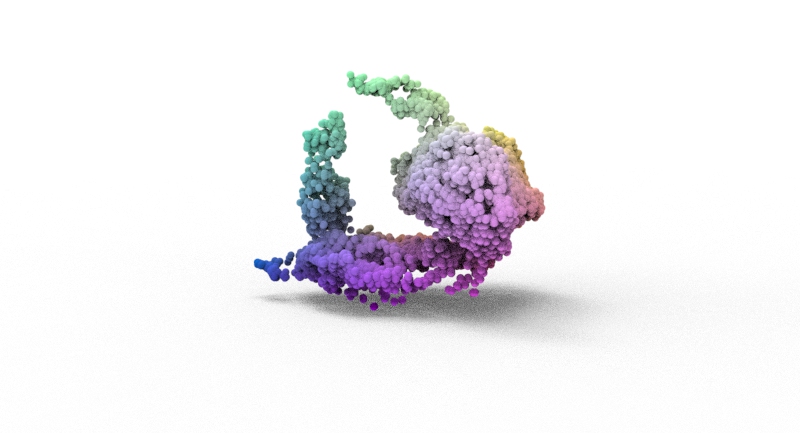}}
        \hspace{2mm}
        \subfigure{
		\includegraphics[width=0.26\linewidth]{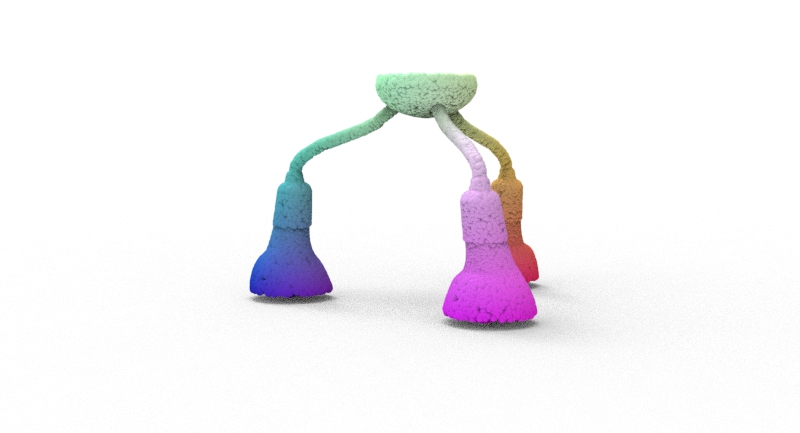}}
        \\
        \setcounter{subfigure}{0}
	\subfigure[]{
		\includegraphics[width=0.26\linewidth]{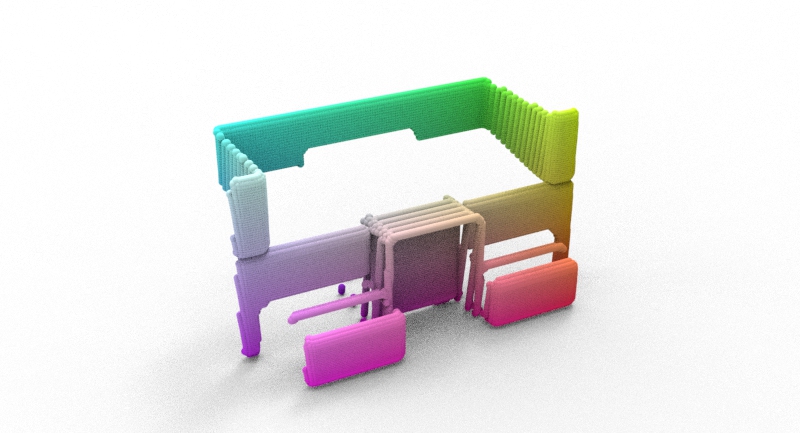}}
        \subfigure[]{
		\includegraphics[width=0.26\linewidth]{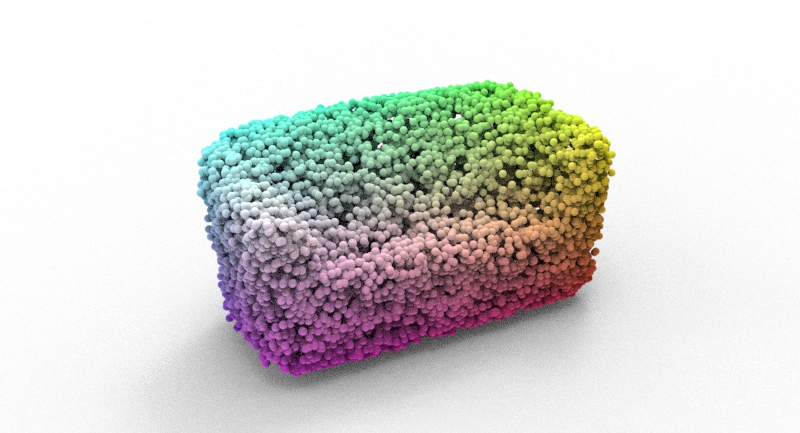}}
        \hspace{2mm}
        \subfigure[]{
		\includegraphics[width=0.26\linewidth]{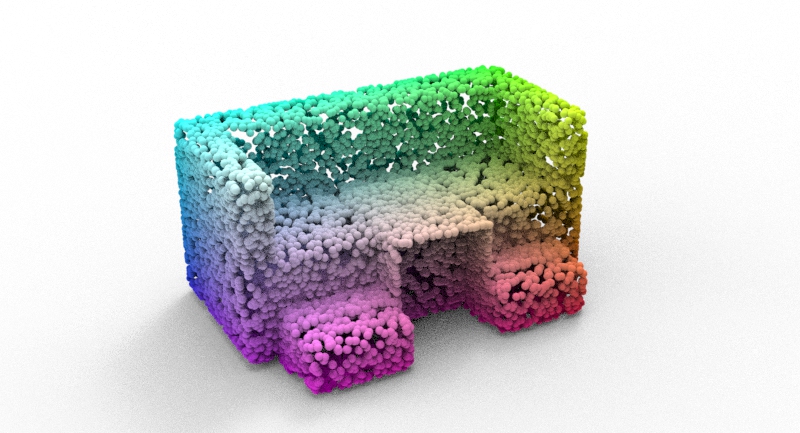}}
	\caption{Examples of the failure cases. (a) Partial inputs. (b) Ours. (c) Ground truth.}
	\label{failurecase}
\end{figure}

\subsubsection{Failure cases} 
\wltrev{In Figure \ref{failurecase}, we present some examples of failure cases. When the shape of the target object is rare within its category or the occlusion is severe, our method struggles to reconstruct the accurate structures, instead erroneously generating more common structures. This is because, without full supervision, our method heavily relies on learning the common geometric features within the category to reconstruct missing structures, which is a constraint shared by current self-supervised methods.}

\subsection{Ablation Study}

\subsubsection{Effects of each module}

\begin{table*}[htbp]
	\centering
	\renewcommand\arraystretch{1.25}
	\caption{Quantitative comparison of the effectiveness of each module in our framework. ``\ding{51}" and ``\ding{55}" represent the model with or without the component, respectively. ``PE" and ``CE" denote the position encoding and the curvature encoding, respectively. ``MAN" means the multi-view adversarial network, and ``DARE" represents the density-aware radius estimation algorithm. The \textbf{bold} numbers represent the best results. }
	\label{component_ablation}
	\begin{tabular}{c|c|c|c|c|c|c|c|c|c|c|c|c}
		\toprule[1.2pt]
		\multicolumn{4}{c|}{Components} & \multicolumn{9}{c}{Categories}                   \\ \hline
		PE         & CE	 & MAN	& DARE     & Average & Plane & Cabinet & Car & Chair & Lamp	& Couch	& Table	& Watercraft \\ \hline
		\ding{55}    & \ding{55} & \ding{55}	& \ding{55}    & 11.97  & 3.89 & 13.22 & 7.34  & 12.83 & 24.22  & 11.15 	& 14.43 & 8.68 \\
		\ding{51}    & \ding{55}	& \ding{55}	& \ding{55}    &  10.76  &3.85 & 12.73 & 6.68 & 12.40 & 16.89 & 11.34 & 14.12 & 8.11 \\
		\ding{51}    & \ding{51} & \ding{55}	& \ding{55}    &  9.87 & 3.44 & 10.53 & \textbf{5.83} & 11.75 & 16.64 & 10.23 & 12.73 & 7.78  \\
		\ding{51}    & \ding{51}	& \ding{51} & \ding{55} & 9.32  & \textbf{2.25} & \textbf{10.34}  & 6.13 & 10.97 & 14.67 & 10.25	& 12.52 & 7.49 \\
		\ding{51}    & \ding{51}	& \ding{51}	& \ding{51} & \textbf{9.08}	& 2.35 & 10.64  & 5.92 & \textbf{10.14} & \textbf{14.51} & \textbf{10.15}  & \textbf{11.65}  & \textbf{7.28} \\

        \bottomrule[1.2pt]
	\end{tabular}
\end{table*}

\begin{figure}[htbp]
	\centering  
	\subfigbottomskip=1pt 
	\subfigure{
		\includegraphics[width=0.21\linewidth]{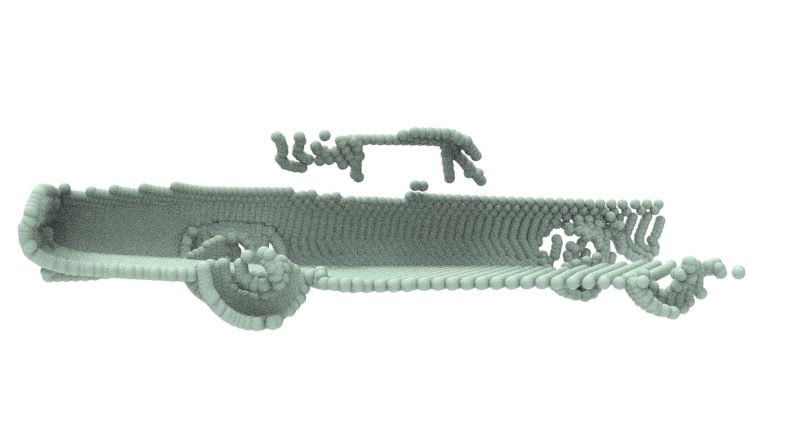}}
        \subfigure{
		\includegraphics[width=0.21\linewidth]{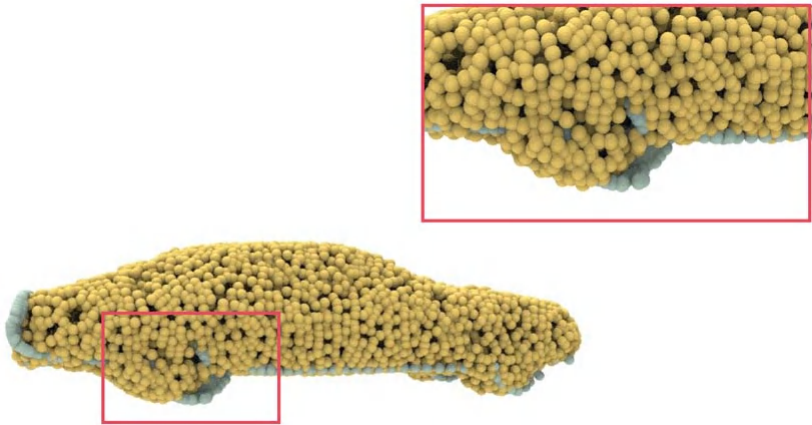}}
        \hspace{2mm}
        \subfigure{
		\includegraphics[width=0.21\linewidth]{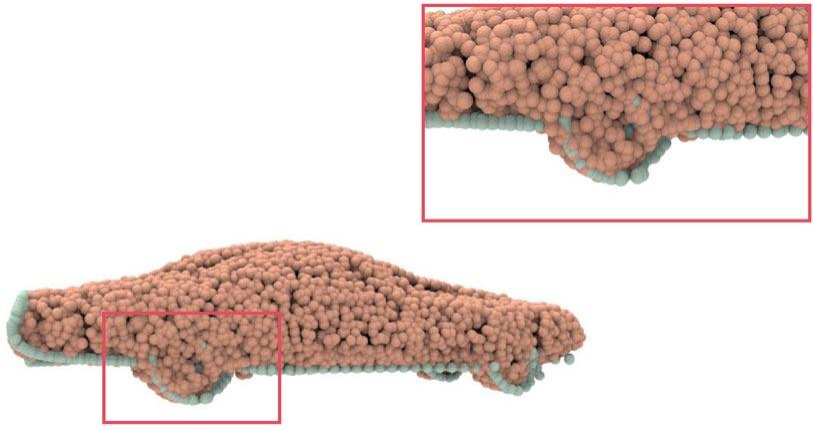}}
        \subfigure{
		\includegraphics[width=0.21\linewidth]{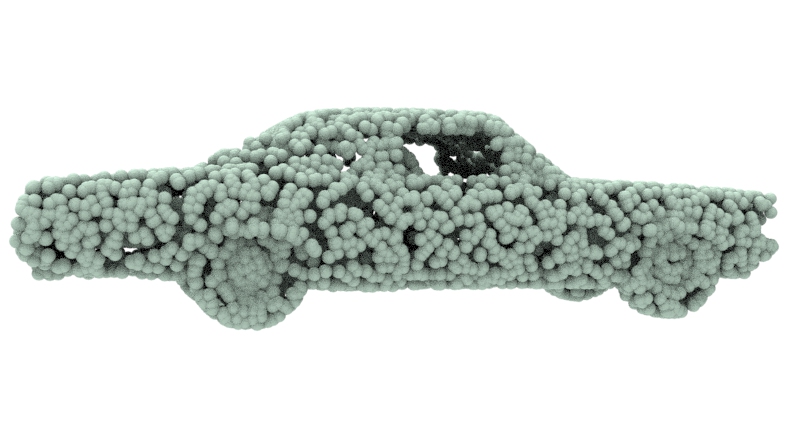}}
        \\
        \setcounter{subfigure}{0}
	\subfigure[]{
		\includegraphics[width=0.21\linewidth]{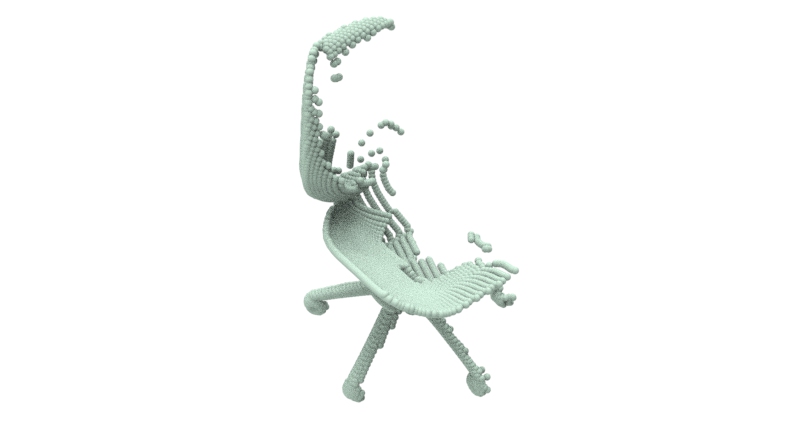}}
        \subfigure[]{
		\includegraphics[width=0.21\linewidth]{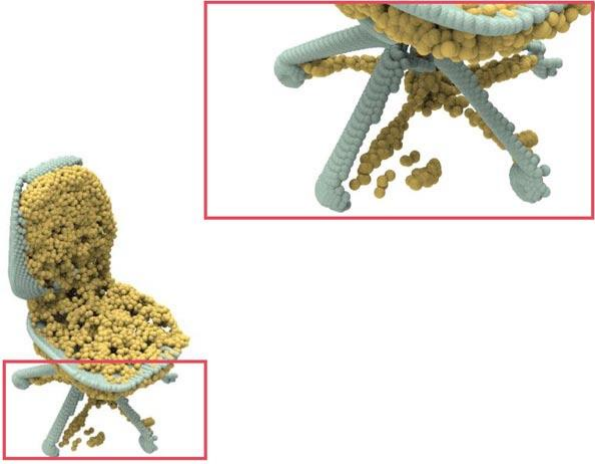}}
        \hspace{2mm}
        \subfigure[]{
		\includegraphics[width=0.21\linewidth]{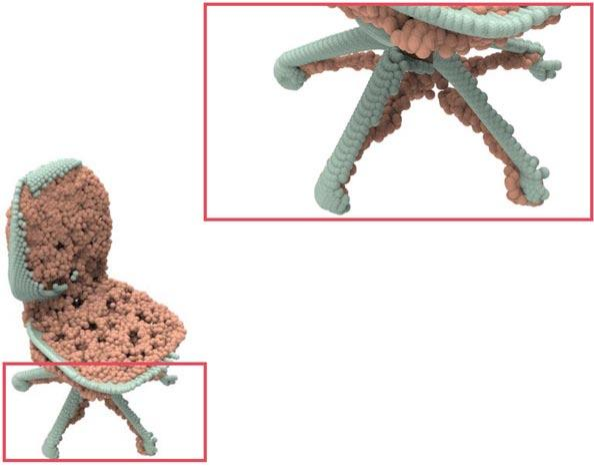}}
        \subfigure[]{
		\includegraphics[width=0.21\linewidth]{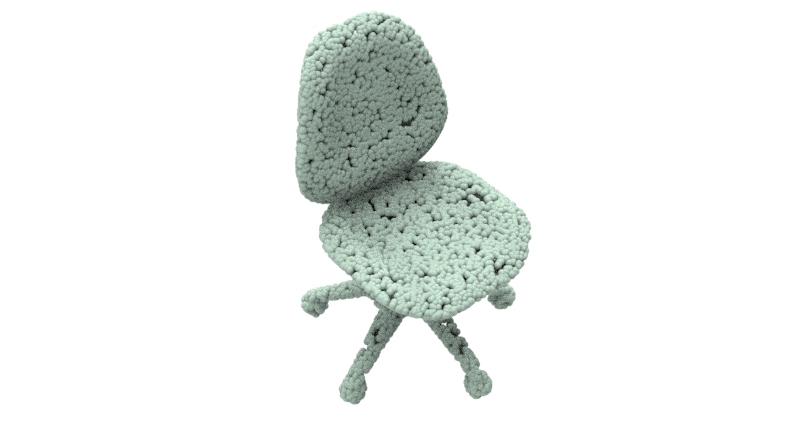}}
	\caption{Visual comparison of reconstructed results with and without the position encodings. With PE, the positions of the predicted point are more accurate, resulting in fewer noisy points. (a) Partial inputs. (b) Point clouds predicted without PE (yellow points) and partial inputs (green points). (c) Point clouds predicted with PE (red points) and partial inputs (green points). (d) Ground truth.}
	\label{visual_pe}
\end{figure}

To analyze the effect of each innovative component in our framework, including position encoding, curvature encoding, multi-view adversarial network, and density-aware radius estimation algorithm, we conduct ablation experiments on synthetic data. As Table \ref{component_ablation} shows, the inclusion of all modules (PE, CE, MAN, and DARE) yields the best average performance, achieving the lowest average CD of 9.08. In particular, adding only PE results in a noticeable reduction in the average CD to 10.76, especially a substantial gain in the lamp category (from 24.22 to 16.89). 
We also compare the visualization results with and without using PE to observe the impact of PE. \wltrev{As Figure \ref{visual_pe} illustrates, without PE, the model predicts some inaccurate points, including outliers that significantly deviate from the partial input and the ground truth. After incorporating PE, however, these inaccurate points are noticeably reduced. We attribute this effect to the following reasons: PE captures the local patterns of the coarse shape \(\textbf{P}_c\) and the partial input $\textbf{P}_{in}$. As discussed in Section \ref{sec_prn}, the PRN retrieves the most relevant PE from $\textbf{P}_{in}$ for \(\textbf{P}_c\) and uses them to predict an offset for \(\textbf{P}_c\). Consequently, even though the local regions in \(\textbf{P}_c\) may be noisy, the PRN leverages the structured and noise-free local patterns from $\textbf{P}_{in}$ to calibrate the positions of noisy local points in \(\textbf{P}_c\).}

\begin{figure*}[htbp]
	\centering  
	\subfigbottomskip=1pt 
	\subfigure{
		\includegraphics[width=0.21\linewidth]{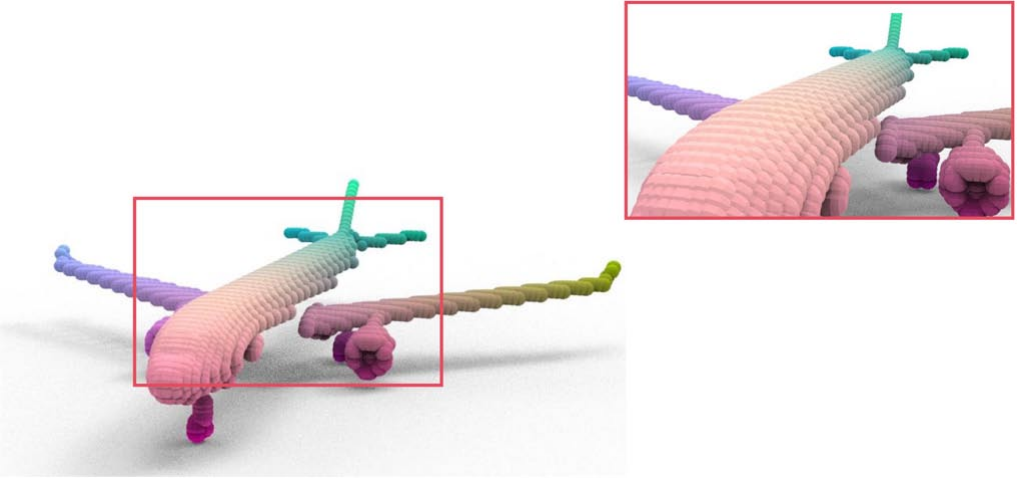}}
	\hspace{3mm}
        \subfigure{
		\includegraphics[width=0.21\linewidth]{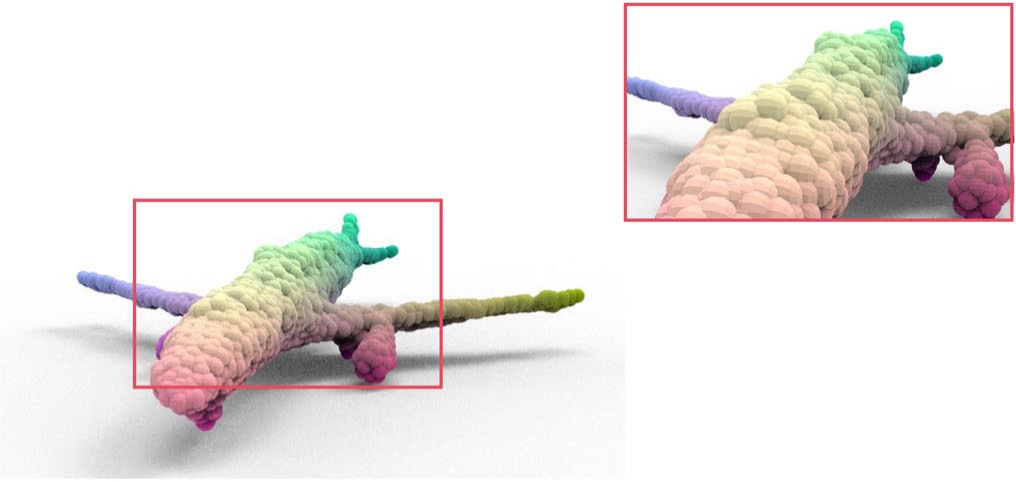}}
        \hspace{3mm}
        \subfigure{
		\includegraphics[width=0.21\linewidth]{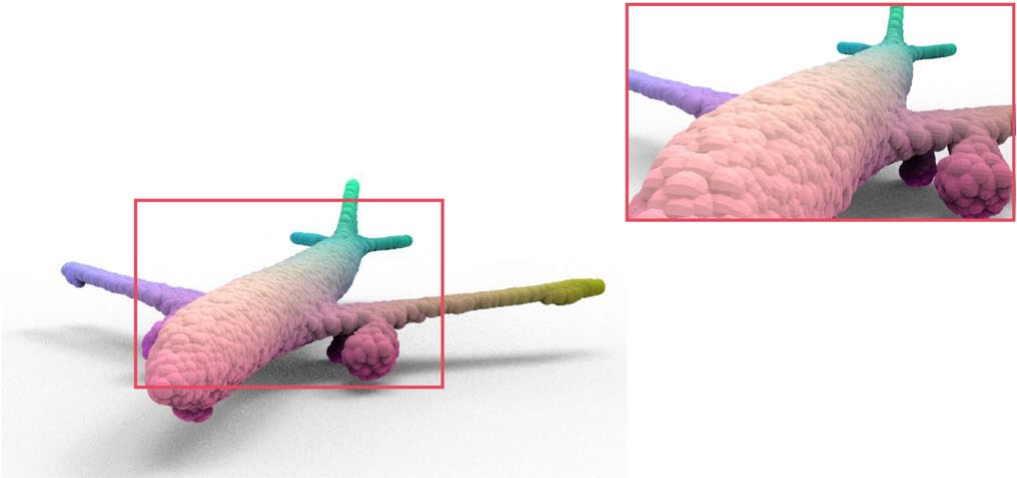}}
	\hspace{3mm}
        \subfigure{
		\includegraphics[width=0.21\linewidth]{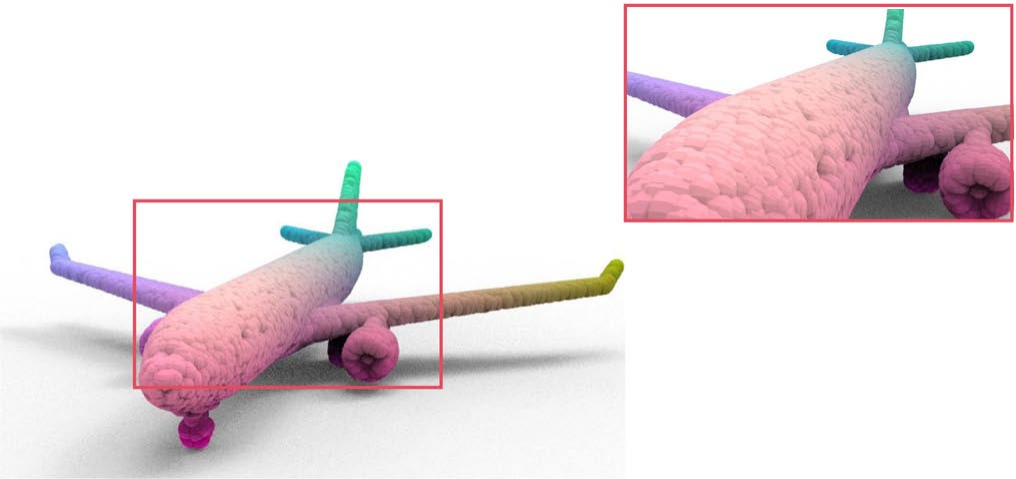}}
        \\
        \setcounter{subfigure}{0}
	\subfigure[]{
		\includegraphics[width=0.21\linewidth]{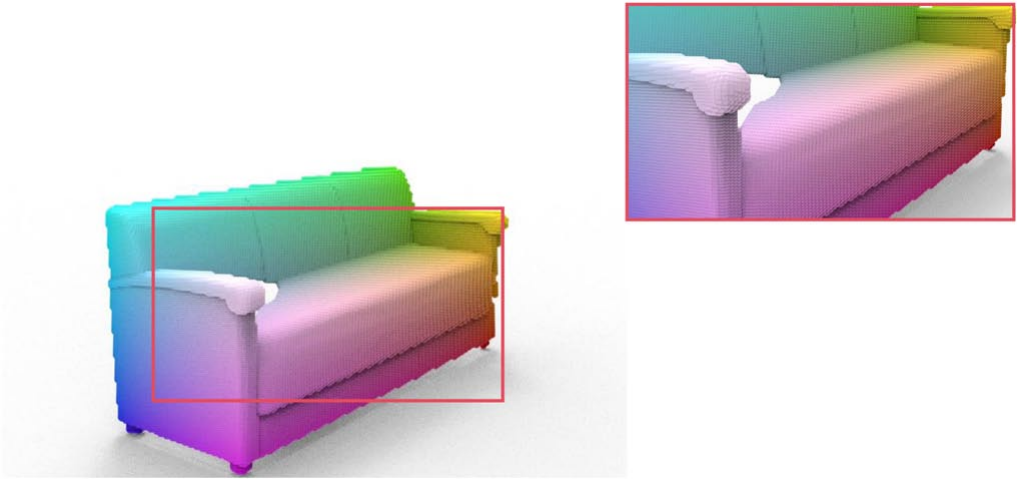}}
	\hspace{3mm}
        \subfigure[]{
		\includegraphics[width=0.21\linewidth]{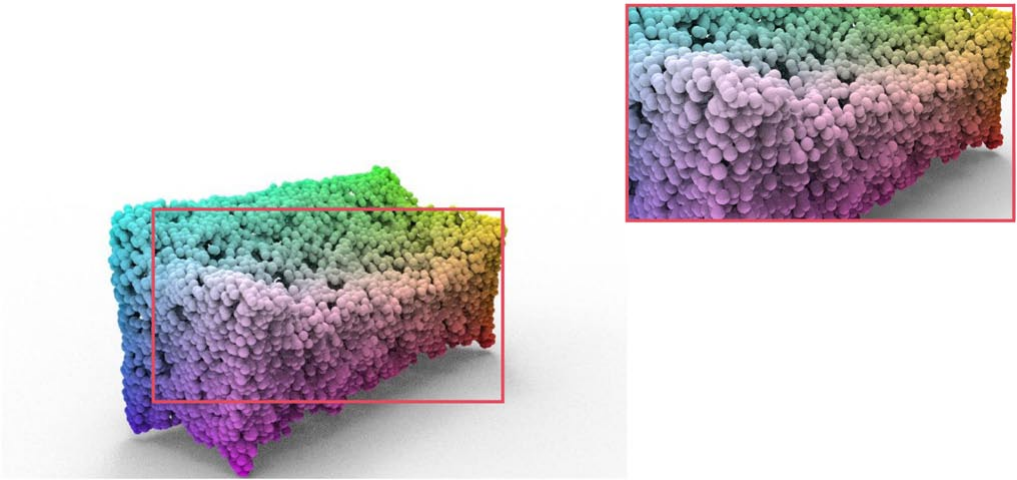}}
	\hspace{3mm}
        \subfigure[]{
		\includegraphics[width=0.21\linewidth]{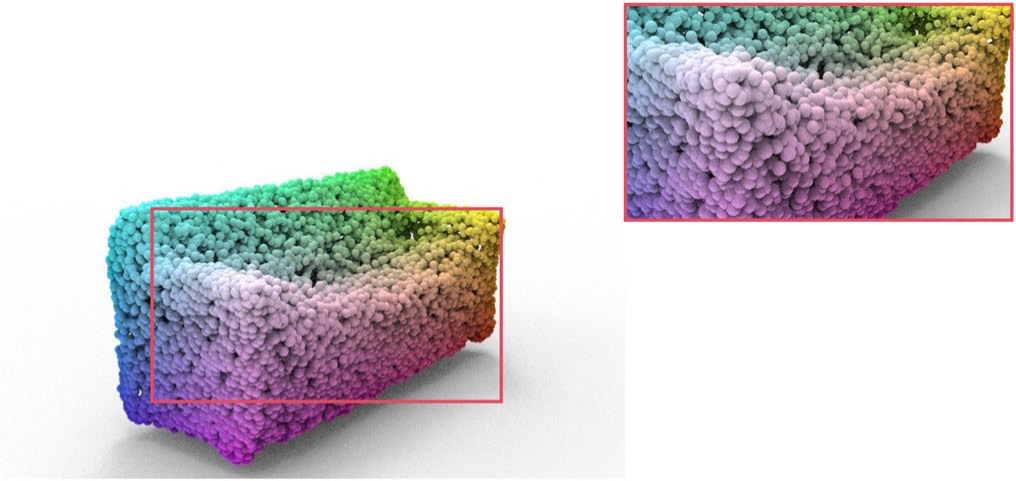}}
	\hspace{3mm}
        \subfigure[]{
		\includegraphics[width=0.21\linewidth]{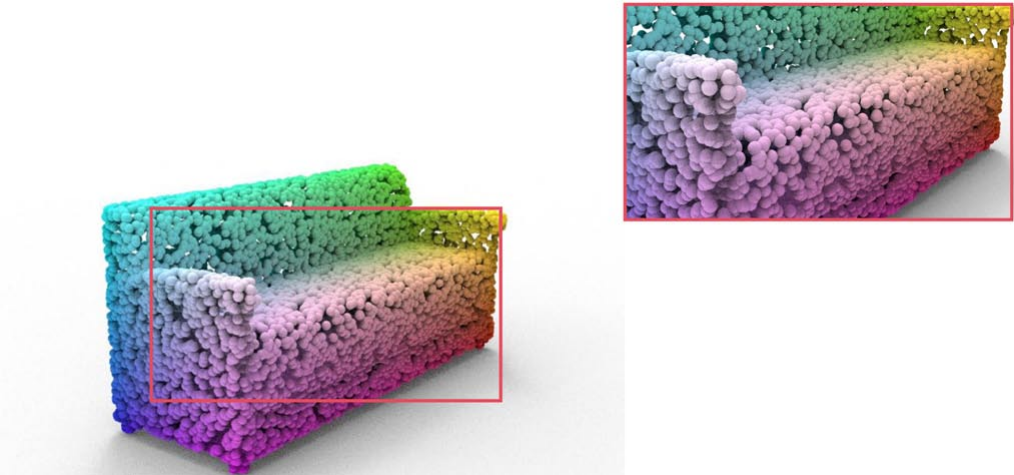}}
	\caption{Visual comparison of reconstructed results with and without the curvature encodings. After using CE, the surfaces of the predicted point clouds are smoother, similar to the input incomplete point cloud. (a) Partial inputs. (b) Point clouds predicted without CE. (c) Point clouds predicted with CE. (d) Ground truth.}
	\label{visual_ce}
\end{figure*}

\begin{figure*}[htbp]
	\centering  
	\subfigbottomskip=1pt 
	\subfigure{
		\includegraphics[width=0.11\linewidth]{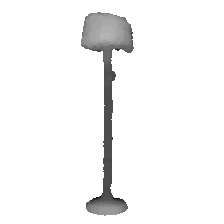}}
	\subfigure{
		\includegraphics[width=0.11\linewidth]{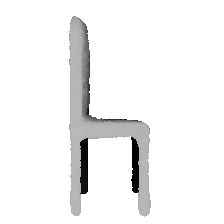}}
        \subfigure{
		\includegraphics[width=0.11\linewidth]{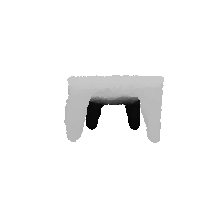}}
	\subfigure{
		\includegraphics[width=0.11\linewidth]{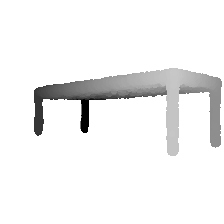}}
	\subfigure{
		\includegraphics[width=0.11\linewidth]{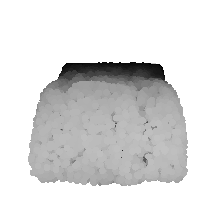}}
	\subfigure{
		\includegraphics[width=0.11\linewidth]{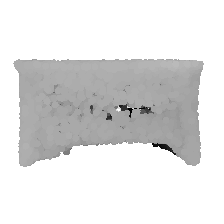}}
	\subfigure{
		\includegraphics[width=0.11\linewidth]{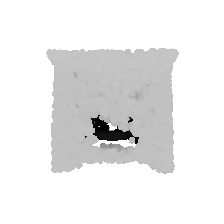}}
	\subfigure{
		\includegraphics[width=0.11\linewidth]{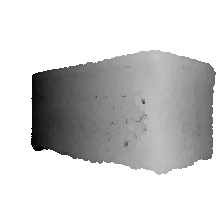}}
        \vspace{5mm}
        \\
        \setcounter{subfigure}{0}
	\subfigure[]{
		\includegraphics[width=0.11\linewidth]{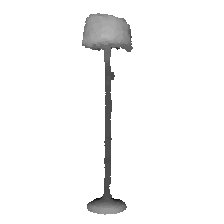}}
	\subfigure[]{
		\includegraphics[width=0.11\linewidth]{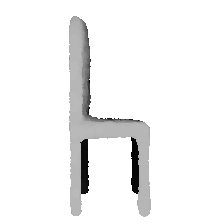}}
	\subfigure[]{
		\includegraphics[width=0.11\linewidth]{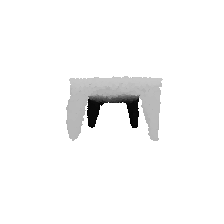}}
	\subfigure[]{
		\includegraphics[width=0.11\linewidth]{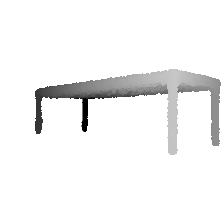}}
	\subfigure[]{
		\includegraphics[width=0.11\linewidth]{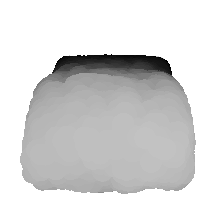}}
	\subfigure[]{
		\includegraphics[width=0.11\linewidth]{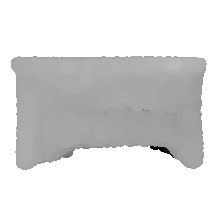}}
	\subfigure[]{
		\includegraphics[width=0.11\linewidth]{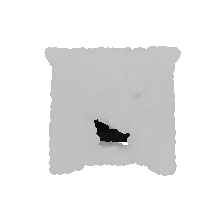}}
	\subfigure[]{
		\includegraphics[width=0.11\linewidth]{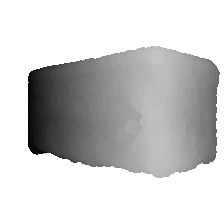}}
	\caption{\wltrevRe{Comparison of rendered depth maps with a fixed radius and with the DARE algorithm. The first and second rows show the rendering results before and after applying the DARE algorithm, respectively. The DARE algorithm adaptively adjusts the rendering radius based on the density of projected points, enabling smoother foregrounds for low-density depth maps while preventing the expansion of foreground regions in high-density depth maps. (a)$\sim$(d) High-density depth maps. (e)$\sim$(h) Low-density depth maps.}}
	\label{visual_dare}
\end{figure*}

Additionally, it is evident from Table \ref{component_ablation} that incorporating both PE and CE reduces the average CD further to 9.87. CE complements PE by providing additional curvature information, significantly improving performance in categories such as table, couch, and cabinet. 
Moreover, as observed from the visual results in Figure \ref{visual_ce}, embedding CE results in a noticeably smoother reconstructed surface when the incomplete input point cloud surface is compact and smooth. \wltrev{Like PE, CE encodes the curvature patterns of the local surfaces of \(\textbf{P}_c\) and $\textbf{P}_{in}$. Through attention-based retrieval, PRN leverages the most relevant smooth surface knowledge from $\textbf{P}_{in}$ to refine the rough local surfaces of \(\textbf{P}_c\).}

In addition to PE and CE, the inclusion of MAN further enhances the performance as demonstrated in Table \ref{component_ablation}, with an average improvement of 0.55. Note that the density-aware radius estimation algorithm also contributes positively to the model's performance, with significant improvements in categories like chair and table. 

To more intuitively demonstrate the effect of the DARE algorithm, we compare the depth maps rendered with a fixed radius and with the DARE algorithm. As shown in Figure \ref{visual_dare}, when a fixed radius is used for rendering, high-density depth maps appear smoother, but low-density depth maps often contain many holes. However, after applying the DARE algorithm, the foreground of low-density depth maps becomes smoother, while the edges of foreground pixels in high-density depth maps do not expand. This proves that the proposed DARE algorithm can adaptively select an appropriate point radius based on the density of projected points, thereby rendering higher-quality depth maps.

\subsubsection{Ablation on supervisory signals} 
Our solution employs two types of direct supervisory signals: an incomplete 3D point cloud and a 2D silhouette map. Although these signals have different data representations, they both originate from the same source, i.e., the incomplete point cloud is derived by back-projecting the depth map into 3D space, and the silhouette map is binarized from the depth map. We do not incorporate 2D supervision from other viewpoints.

To evaluate the impact of these two supervisory signals on the model's performance, we conduct ablation studies on the synthetic dataset. We use our complete model and perform experiments by separately removing \(L_{part}\) and \(L_{rend}\). 
The results show that non-converge occurs in the absence of either \(L_{part}\) or \(L_{rend}\). For the case of lacking \(L_{rend}\), the CD values achieved on the cabinet, car, couch, and table categories are 36.56, 18.19, 19.68, and 74.83, respectively, while convergence is not achieved on other categories. On the other hand, when \(L_{part}\) is removed, the CD values on cabinet, chair, car, couch, table, and watercraft categories are 29.13, 21.12, 7.36, 12.00, 12.53, and 9.78, respectively, while convergence failed for plane and lamp categories. We believe that the reasons for the non-convergence are twofold. First, the training of GAN is inherently unstable \cite{gan_goodfellow, mao2017least}. Second, the inclusion of DARE makes training more challenging. This is because DARE requires a relatively accurate initial depth map rendered from the predicted results before estimating a more precise point radius. If the initial depth map has significant deviations, the estimated point radius will also deviate accordingly, leading to significant errors in the second-round rendering. Consequently, the GAN module might collapse, ultimately preventing the model from converging. In conclusion, both \(L_{rend}\) and \(L_{part}\) are required to predict accurate shapes. However, based on the results from the converged categories, \(L_{rend}\) exhibits stronger supervisory capabilities compared to \(L_{part}\).

\section{Limitation and Future Work}
\label{limit_future}
\wltrev{Although our method achieves outstanding results, it also has some limitations. \wltrevRe{First, since MAL-UPC relies on the geometric similarity of objects within the same category, it requires category labels to allocate the raw depth maps of the partial point clouds to their respective category-specific image banks.} Besides, our model performs poorly on unseen categories and rare objects, despite outperforming self-supervised approaches. In the future,  we will attempt to eliminate the dependency on category labels using methods such as clustering and contrastive learning. Additionally, we plan to leverage large quantities of 2D images and current Large Vision Models to provide 2D priors for partial point clouds, so as to improve the generalization capability of unsupervised methods.}
\section{Conclusion}
\label{conclusion}
\wltrev{In the scenario of unsupervised point cloud completion, it is critical to fully utilize the prior geometries from partial observations due to the lack of complete shapes as supervision}. To this end, we have presented a novel approach that effectively exploits the objects' intrinsic region-level and category-specific geometric similarity to infer complete shapes without the need for multi-view partial scans and masked operation. Our method demonstrates strong capabilities in both inferring complete shapes and preserving incomplete input, resulting in reconstructed shapes that are both reasonable and accurate, with minimal noise. Quantitatively and visually, our approach surpasses current state-of-the-art self-supervised methods, as well as a portion of unpaired methods. We hope this work inspires future research in 3D object and scene completion, reconstruction, and understanding tasks. 

\ifCLASSOPTIONcaptionsoff
  \newpage
\fi

\bibliographystyle{IEEEtran}

\begin{thebibliography}{10}
\providecommand{\url}[1]{#1}
\csname url@samestyle\endcsname
\providecommand{\newblock}{\relax}
\providecommand{\bibinfo}[2]{#2}
\providecommand{\BIBentrySTDinterwordspacing}{\spaceskip=0pt\relax}
\providecommand{\BIBentryALTinterwordstretchfactor}{4}
\providecommand{\BIBentryALTinterwordspacing}{\spaceskip=\fontdimen2\font plus
\BIBentryALTinterwordstretchfactor\fontdimen3\font minus \fontdimen4\font\relax}
\providecommand{\BIBforeignlanguage}[2]{{%
\expandafter\ifx\csname l@#1\endcsname\relax
\typeout{** WARNING: IEEEtran.bst: No hyphenation pattern has been}%
\typeout{** loaded for the language `#1'. Using the pattern for}%
\typeout{** the default language instead.}%
\else
\language=\csname l@#1\endcsname
\fi
#2}}
\providecommand{\BIBdecl}{\relax}
\BIBdecl

\bibitem{zhang2024text2nerf}
J.~Zhang, X.~Li, Z.~Wan, C.~Wang, and J.~Liao, ``Text2nerf: Text-driven 3d scene generation with neural radiance fields,'' \emph{IEEE Trans. Vis. Comput. Graph.}, 2024.

\bibitem{wang2022and}
J.~Wang, Y.~Li, Z.~Zhou, C.~Wang, Y.~Hou, L.~Zhang, X.~Xue, M.~Kamp, X.~L. Zhang, and S.~Chen, ``When, where and how does it fail? a spatial-temporal visual analytics approach for interpretable object detection in autonomous driving,'' \emph{IEEE Trans. Vis. Comput. Graph.}, vol.~29, no.~12, pp. 5033--5049, 2022.

\bibitem{cao2022monoscene}
A.-Q. Cao and R.~De~Charette, ``Monoscene: Monocular 3d semantic scene completion,'' in \emph{Proc. IEEE Conf. Comput. Vis. Pattern Recog.}, 2022, pp. 3991--4001.

\bibitem{li2023voxformer}
Y.~Li, Z.~Yu, C.~Choy, C.~Xiao, J.~M. Alvarez, S.~Fidler, C.~Feng, and A.~Anandkumar, ``Voxformer: Sparse voxel transformer for camera-based 3d semantic scene completion,'' in \emph{Proc. IEEE Conf. Comput. Vis. Pattern Recog.}, 2023, pp. 9087--9098.

\bibitem{cui2021deep}
Y.~Cui, R.~Chen, W.~Chu, L.~Chen, D.~Tian, Y.~Li, and D.~Cao, ``Deep learning for image and point cloud fusion in autonomous driving: A review,'' \emph{IEEE Trans. Intell. Transp. Syst.}, vol.~23, no.~2, pp. 722--739, 2021.

\bibitem{xu2019mo}
W.~Xu, A.~Chatterjee, M.~Zollhoefer, H.~Rhodin, P.~Fua, H.-P. Seidel, and C.~Theobalt, ``Mo 2 cap 2: Real-time mobile 3d motion capture with a cap-mounted fisheye camera,'' \emph{IEEE Trans. Vis. Comput. Graph.}, vol.~25, no.~5, pp. 2093--2101, 2019.

\bibitem{park20203d}
G.~Park, A.~Argyros, J.~Lee, and W.~Woo, ``3d hand tracking in the presence of excessive motion blur,'' \emph{IEEE Trans. Vis. Comput. Graph.}, vol.~26, no.~5, pp. 1891--1901, 2020.

\bibitem{han2020live}
L.~Han, T.~Zheng, Y.~Zhu, L.~Xu, and L.~Fang, ``Live semantic 3d perception for immersive augmented reality,'' \emph{IEEE Trans. Vis. Comput. Graph.}, vol.~26, no.~5, pp. 2012--2022, 2020.

\bibitem{ren2023geoudf}
S.~Ren, J.~Hou, X.~Chen, Y.~He, and W.~Wang, ``Geoudf: Surface reconstruction from 3d point clouds via geometry-guided distance representation,'' in \emph{Proc. IEEE Int. Conf. Comput. Vis.}, 2023, pp. 14\,214--14\,224.

\bibitem{zhang2023unleash}
Y.~Zhang, Q.~Zhang, J.~Hou, Y.~Yuan, and G.~Xing, ``Unleash the potential of image branch for cross-modal 3d object detection,'' in \emph{Proc. Adv. Neural Inf. Process. Syst.}, 2023.

\bibitem{qian20223d}
R.~Qian, X.~Lai, and X.~Li, ``3d object detection for autonomous driving: A survey,'' \emph{Pattern Recog.}, vol. 130, p. 108796, 2022.

\bibitem{mao20233d}
J.~Mao, S.~Shi, X.~Wang, and H.~Li, ``3d object detection for autonomous driving: A comprehensive survey,'' \emph{Int. J. Comput. Vision}, vol. 131, no.~8, pp. 1909--1963, 2023.

\bibitem{xia2023scpnet}
Z.~Xia, Y.~Liu, X.~Li, X.~Zhu, Y.~Ma, Y.~Li, Y.~Hou, and Y.~Qiao, ``Scpnet: Semantic scene completion on point cloud,'' in \emph{Proc. IEEE Conf. Comput. Vis. Pattern Recog.}, 2023, pp. 17\,642--17\,651.

\bibitem{hou2021exploring}
J.~Hou, B.~Graham, M.~Nie{\ss}ner, and S.~Xie, ``Exploring data-efficient 3d scene understanding with contrastive scene contexts,'' in \emph{Proc. IEEE Conf. Comput. Vis. Pattern Recog.}, 2021, pp. 15\,587--15\,597.

\bibitem{pcn}
W.~Yuan, T.~Khot, D.~Held, C.~Mertz, and M.~Hebert, ``{PCN:} point completion network,'' in \emph{Int. Conf. 3D Vis.}, 2018, pp. 728--737.

\bibitem{pointr}
X.~Yu, Y.~Rao, Z.~Wang, Z.~Liu, J.~Lu, and J.~Zhou, ``Pointr: Diverse point cloud completion with geometry-aware transformers,'' in \emph{Proc. IEEE Int. Conf. Comput. Vis.}, 2021, pp. 12\,478--12\,487.

\bibitem{chen2023anchorformer}
Z.~Chen, F.~Long, Z.~Qiu, T.~Yao, W.~Zhou, J.~Luo, and T.~Mei, ``Anchorformer: Point cloud completion from discriminative nodes,'' in \emph{Proc. IEEE Conf. Comput. Vis. Pattern Recog.}, 2023, pp. 13\,581--13\,590.

\bibitem{snowflake}
P.~Xiang, X.~Wen, Y.~Liu, Y.~Cao, P.~Wan, W.~Zheng, and Z.~Han, ``Snowflakenet: Point cloud completion by snowflake point deconvolution with skip-transformer,'' in \emph{Proc. IEEE Int. Conf. Comput. Vis.}, 2021, pp. 5479--5489.

\bibitem{wen2020point}
X.~Wen, T.~Li, Z.~Han, and Y.-S. Liu, ``Point cloud completion by skip-attention network with hierarchical folding,'' in \emph{Proc. IEEE Conf. Comput. Vis. Pattern Recog.}, 2020, pp. 1939--1948.

\bibitem{wang2024pointattn}
J.~Wang, Y.~Cui, D.~Guo, J.~Li, Q.~Liu, and C.~Shen, ``Pointattn: You only need attention for point cloud completion,'' in \emph{Proc. AAAI Conf. Artif. Intell.}, vol.~38, no.~6, 2024, pp. 5472--5480.

\bibitem{kasten2024point}
Y.~Kasten, O.~Rahamim, and G.~Chechik, ``Point cloud completion with pretrained text-to-image diffusion models,'' \emph{Proc. Adv. Neural Inf. Process. Syst.}, vol.~36, 2024.

\bibitem{cai2024orthogonal}
P.~Cai, D.~Scott, X.~Li, and S.~Wang, ``Orthogonal dictionary guided shape completion network for point cloud,'' in \emph{Proc. AAAI Conf. Artif. Intell.}, vol.~38, no.~2, 2024, pp. 864--872.

\bibitem{huang2020pf}
Z.~Huang, Y.~Yu, J.~Xu, F.~Ni, and X.~Le, ``Pf-net: Point fractal network for 3d point cloud completion,'' in \emph{Proc. IEEE Conf. Comput. Vis. Pattern Recog.}, 2020, pp. 7662--7670.

\bibitem{wang2020cascaded}
X.~Wang, M.~H. Ang~Jr, and G.~H. Lee, ``Cascaded refinement network for point cloud completion,'' in \emph{Proc. IEEE Conf. Comput. Vis. Pattern Recog.}, 2020, pp. 790--799.

\bibitem{pcl2plc}
X.~Chen, B.~Chen, and N.~J. Mitra, ``Unpaired point cloud completion on real scans using adversarial training,'' in \emph{Proc. Int. Conf. Learn. Represent.}, 2020.

\bibitem{cycle}
X.~Wen, Z.~Han, Y.~Cao, P.~Wan, W.~Zheng, and Y.~Liu, ``Cycle4completion: Unpaired point cloud completion using cycle transformation with missing region coding,'' in \emph{Proc. IEEE Conf. Comput. Vis. Pattern Recog.}, 2021, pp. 13\,080--13\,089.

\bibitem{ganinverse}
J.~Zhang, X.~Chen, Z.~Cai, L.~Pan, H.~Zhao, S.~Yi, C.~K. Yeo, B.~Dai, and C.~C. Loy, ``Unsupervised 3d shape completion through {GAN} inversion,'' in \emph{Proc. IEEE Conf. Comput. Vis. Pattern Recog.}, 2021, pp. 1768--1777.

\bibitem{lsls}
Y.~Cai, K.-Y. Lin, C.~Zhang, Q.~Wang, X.~Wang, and H.~Li, ``Learning a structured latent space for unsupervised point cloud completion,'' in \emph{Proc. IEEE Conf. Comput. Vis. Pattern Recog.}, 2022, pp. 5543--5553.

\bibitem{cao2023kt}
Z.~Cao, W.~Zhang, X.~Wen, Z.~Dong, Y.-S. Liu, X.~Xiao, and B.~Yang, ``Kt-net: knowledge transfer for unpaired 3d shape completion,'' in \emph{Proc. AAAI Conf. Artif. Intell.}, vol.~37, no.~1, 2023, pp. 286--294.

\bibitem{ma2023symmetric}
C.~Ma, Y.~Chen, P.~Guo, J.~Guo, C.~Wang, and Y.~Guo, ``Symmetric shape-preserving autoencoder for unsupervised real scene point cloud completion,'' in \emph{Proc. IEEE Conf. Comput. Vis. Pattern Recog.}, 2023, pp. 13\,560--13\,569.

\bibitem{liu2024cloudmix}
F.~Liu, J.~Gong, Q.~Zhou, X.~Lu, R.~Yi, Y.~Xie, and L.~Ma, ``Cloudmix: Dual mixup consistency for unpaired point cloud completion,'' \emph{IEEE Trans. Vis. Comput. Graph.}, 2024.

\bibitem{wu2023leveraging}
L.~Wu, Q.~Zhang, J.~Hou, and Y.~Xu, ``Leveraging single-view images for unsupervised 3d point cloud completion,'' \emph{IEEE Trans. Multimedia}, 2023.

\bibitem{aiello2022cross}
E.~Aiello, D.~Valsesia, and E.~Magli, ``Cross-modal learning for image-guided point cloud shape completion,'' \emph{Proc. Adv. Neural Inf. Process. Syst.}, vol.~35, pp. 37\,349--37\,362, 2022.

\bibitem{3depn}
A.~Dai, C.~R. Qi, and M.~Nie{\ss}ner, ``Shape completion using 3d-encoder-predictor cnns and shape synthesis,'' in \emph{Proc. IEEE Conf. Comput. Vis. Pattern Recog.}, 2017, pp. 6545--6554.

\bibitem{grnet}
H.~Xie, H.~Yao, S.~Zhou, J.~Mao, S.~Zhang, and W.~Sun, ``Grnet: Gridding residual network for dense point cloud completion,'' in \emph{Proc. European Conf. Comput. Vis.}, vol. 12354, 2020, pp. 365--381.

\bibitem{sun2022patchrd}
B.~Sun, V.~G. Kim, N.~Aigerman, Q.~Huang, and S.~Chaudhuri, ``Patchrd: Detail-preserving shape completion by learning patch retrieval and deformation,'' in \emph{Proc. European Conf. Comput. Vis.}, 2022, pp. 503--522.

\bibitem{yan2022shapeformer}
X.~Yan, L.~Lin, N.~J. Mitra, D.~Lischinski, D.~Cohen-Or, and H.~Huang, ``Shapeformer: Transformer-based shape completion via sparse representation,'' in \emph{Proc. IEEE Conf. Comput. Vis. Pattern Recog.}, 2022, pp. 6239--6249.

\bibitem{chibane2020implicit}
J.~Chibane, T.~Alldieck, and G.~Pons-Moll, ``Implicit functions in feature space for 3d shape reconstruction and completion,'' in \emph{Proc. IEEE Conf. Comput. Vis. Pattern Recog.}, 2020, pp. 6970--6981.

\bibitem{mittal2022autosdf}
P.~Mittal, Y.-C. Cheng, M.~Singh, and S.~Tulsiani, ``Autosdf: Shape priors for 3d completion, reconstruction and generation,'' in \emph{Proc. IEEE Conf. Comput. Vis. Pattern Recog.}, 2022, pp. 306--315.

\bibitem{wuwssc}
L.~Wu, J.~Hou, L.~Song, and Y.~Xu, ``3d shape completion on unseen categories: A weakly-supervised approach,'' \emph{IEEE Trans. Vis. Comput. Graph.}, pp. 1--15, 2024.

\bibitem{foldingnet}
Y.~Yang, C.~Feng, Y.~Shen, and D.~Tian, ``Foldingnet: Point cloud auto-encoder via deep grid deformation,'' in \emph{Proc. IEEE Conf. Comput. Vis. Pattern Recog.}, 2018, pp. 206--215.

\bibitem{topnet}
L.~P. Tchapmi, V.~Kosaraju, H.~Rezatofighi, I.~D. Reid, and S.~Savarese, ``Topnet: Structural point cloud decoder,'' in \emph{Proc. IEEE Conf. Comput. Vis. Pattern Recog.}, 2019, pp. 383--392.

\bibitem{li2023proxyformer}
S.~Li, P.~Gao, X.~Tan, and M.~Wei, ``Proxyformer: Proxy alignment assisted point cloud completion with missing part sensitive transformer,'' in \emph{Proc. IEEE Conf. Comput. Vis. Pattern Recog.}, 2023, pp. 9466--9475.

\bibitem{wang2022learning}
Y.~Wang, D.~J. Tan, N.~Navab, and F.~Tombari, ``Learning local displacements for point cloud completion,'' in \emph{Proc. IEEE Conf. Comput. Vis. Pattern Recog.}, 2022, pp. 1568--1577.

\bibitem{zhang2020detail}
W.~Zhang, Q.~Yan, and C.~Xiao, ``Detail preserved point cloud completion via separated feature aggregation,'' in \emph{Proc. European Conf. Comput. Vis.}\hskip 1em plus 0.5em minus 0.4em\relax Springer, 2020, pp. 512--528.

\bibitem{pan2021variational}
L.~Pan, X.~Chen, Z.~Cai, J.~Zhang, H.~Zhao, S.~Yi, and Z.~Liu, ``Variational relational point completion network,'' in \emph{Proc. IEEE Conf. Comput. Vis. Pattern Recog.}, 2021, pp. 8524--8533.

\bibitem{atlasnet}
T.~Groueix, M.~Fisher, V.~G. Kim, B.~C. Russell, and M.~Aubry, ``A papier-m{\^{a}}ch{\'{e}} approach to learning 3d surface generation,'' in \emph{Proc. IEEE Conf. Comput. Vis. Pattern Recog.}, 2018, pp. 216--224.

\bibitem{msnpcc}
M.~Liu, L.~Sheng, S.~Yang, J.~Shao, and S.~Hu, ``Morphing and sampling network for dense point cloud completion,'' in \emph{Proc. AAAI Conf. Artif. Intell.}, 2020, pp. 11\,596--11\,603.

\bibitem{psgaupcl}
J.~Yang, P.~Ahn, D.~Kim, H.~Lee, and J.~Kim, ``Progressive seed generation auto-encoder for unsupervised point cloud learning,'' in \emph{Proc. IEEE Int. Conf. Comput. Vis.}, 2021, pp. 6393--6402.

\bibitem{tearingnet}
J.~Pang, D.~Li, and D.~Tian, ``Tearingnet: Point cloud autoencoder to learn topology-friendly representations,'' in \emph{Proc. IEEE Conf. Comput. Vis. Pattern Recog.}, 2021, pp. 7453--7462.

\bibitem{vaswani2017attention}
A.~Vaswani, N.~Shazeer, N.~Parmar, J.~Uszkoreit, L.~Jones, A.~N. Gomez, {\L}.~Kaiser, and I.~Polosukhin, ``Attention is all you need,'' \emph{Proc. Adv. Neural Inf. Process. Syst.}, vol.~30, 2017.

\bibitem{tang2022lake}
J.~Tang, Z.~Gong, R.~Yi, Y.~Xie, and L.~Ma, ``Lake-net: Topology-aware point cloud completion by localizing aligned keypoints,'' in \emph{Proc. IEEE Conf. Comput. Vis. Pattern Recog.}, 2022, pp. 1726--1735.

\bibitem{zhou2022seedformer}
H.~Zhou, Y.~Cao, W.~Chu, J.~Zhu, T.~Lu, Y.~Tai, and C.~Wang, ``Seedformer: Patch seeds based point cloud completion with upsample transformer,'' in \emph{Proc. European Conf. Comput. Vis.}\hskip 1em plus 0.5em minus 0.4em\relax Springer, 2022, pp. 416--432.

\bibitem{zhu2023svdformer}
Z.~Zhu, H.~Chen, X.~He, W.~Wang, J.~Qin, and M.~Wei, ``Svdformer: Complementing point cloud via self-view augmentation and self-structure dual-generator,'' in \emph{Proc. IEEE Int. Conf. Comput. Vis.}, 2023, pp. 14\,508--14\,518.

\bibitem{zhao2021point}
H.~Zhao, L.~Jiang, J.~Jia, P.~H. Torr, and V.~Koltun, ``Point transformer,'' in \emph{Proc. IEEE Int. Conf. Comput. Vis.}, 2021, pp. 16\,259--16\,268.

\bibitem{guo2021pct}
M.-H. Guo, J.-X. Cai, Z.-N. Liu, T.-J. Mu, R.~R. Martin, and S.-M. Hu, ``Pct: Point cloud transformer,'' \emph{Comput. Visual Media}, vol.~7, no.~2, pp. 187--199, 2021.

\bibitem{du2024cdpnet}
Z.~Du, J.~Dou, Z.~Liu, J.~Wei, G.~Wang, N.~Xie, and Y.~Yang, ``Cdpnet: Cross-modal dual phases network for point cloud completion,'' in \emph{Proc. AAAI Conf. Artif. Intell.}, vol.~38, no.~2, 2024, pp. 1635--1643.

\bibitem{vipc}
X.~Zhang, Y.~Feng, S.~Li, C.~Zou, H.~Wan, X.~Zhao, Y.~Guo, and Y.~Gao, ``View-guided point cloud completion,'' in \emph{Proc. IEEE Conf. Comput. Vis. Pattern Recog.}, 2021, pp. 15\,890--15\,899.

\bibitem{zhu2023csdn}
Z.~Zhu, L.~Nan, H.~Xie, H.~Chen, J.~Wang, M.~Wei, and J.~Qin, ``Csdn: Cross-modal shape-transfer dual-refinement network for point cloud completion,'' \emph{IEEE Trans. Vis. Comput. Graph.}, 2023.

\bibitem{hu2019render4completion}
T.~Hu, Z.~Han, A.~Shrivastava, and M.~Zwicker, ``Render4completion: Synthesizing multi-view depth maps for 3d shape completion,'' in \emph{Proc. IEEE Int. Conf. Comput. Vis.}, 2019, pp. 0--0.

\bibitem{hu20203d}
T.~Hu, Z.~Han, and M.~Zwicker, ``3d shape completion with multi-view consistent inference,'' in \emph{Proc. AAAI Conf. Artif. Intell.}, vol.~34, no.~07, 2020, pp. 10\,997--11\,004.

\bibitem{wu2020multimodal}
R.~Wu, X.~Chen, Y.~Zhuang, and B.~Chen, ``Multimodal shape completion via conditional generative adversarial networks,'' in \emph{Proc. European Conf. Comput. Vis.}, 2020, pp. 281--296.

\bibitem{cui2022energy}
R.~Cui, S.~Qiu, S.~Anwar, J.~Zhang, and N.~Barnes, ``Energy-based residual latent transport for unsupervised point cloud completion,'' in \emph{Proc. British Mach. Vis. Conf.}, 2022.

\bibitem{gan_goodfellow}
I.~Goodfellow, J.~Pouget-Abadie, M.~Mirza, B.~Xu, D.~Warde-Farley, S.~Ozair, A.~Courville, and Y.~Bengio, ``Generative adversarial nets,'' in \emph{Proc. Adv. Neural Inf. Process. Syst.}, vol.~27, 2014, pp. 5543--5553.

\bibitem{mittal2021self}
H.~Mittal, B.~Okorn, A.~Jangid, and D.~Held, ``Self-supervised point cloud completion via inpainting,'' in \emph{Proc. British Mach. Vis. Conf.}, 2021.

\bibitem{hong2023acl}
S.~Hong, M.~Yavartanoo, R.~Neshatavar, and K.~M. Lee, ``Acl-spc: Adaptive closed-loop system for self-supervised point cloud completion,'' in \emph{Proc. IEEE Conf. Comput. Vis. Pattern Recog.}, 2023, pp. 9435--9444.

\bibitem{cui2023p2c}
R.~Cui, S.~Qiu, S.~Anwar, J.~Liu, C.~Xing, J.~Zhang, and N.~Barnes, ``P2c: Self-supervised point cloud completion from single partial clouds,'' in \emph{Proc. IEEE Int. Conf. Comput. Vis.}, 2023, pp. 14\,351--14\,360.

\bibitem{kim2023learning}
J.~Kim, H.~Kwon, Y.~Yang, and K.-J. Yoon, ``Learning point cloud completion without complete point clouds: A pose-aware approach,'' in \emph{Proc. IEEE Int. Conf. Comput. Vis.}\hskip 1em plus 0.5em minus 0.4em\relax IEEE, 2023, pp. 14\,157--14\,167.

\bibitem{tombari2010unique}
F.~Tombari, S.~Salti, and L.~Di~Stefano, ``Unique signatures of histograms for local surface description,'' in \emph{Proc. European Conf. Comput. Vis.}\hskip 1em plus 0.5em minus 0.4em\relax Springer, 2010, pp. 356--369.

\bibitem{p3ddocu}
N.~Ravi, J.~Reizenstein, D.~Novotn{\'{y}}, T.~Gordon, W.~Lo, J.~Johnson, and G.~Gkioxari, ``Accelerating 3d deep learning with pytorch3d,'' \emph{CoRR}, vol. abs/2007.08501, 2020.

\bibitem{mao2017least}
X.~Mao, Q.~Li, H.~Xie, R.~Y. Lau, Z.~Wang, and S.~Paul~Smolley, ``Least squares generative adversarial networks,'' in \emph{Proc. IEEE Int. Conf. Comput. Vis.}, 2017, pp. 2794--2802.

\bibitem{chang2015shapenet}
A.~X. Chang, T.~Funkhouser, L.~Guibas, P.~Hanrahan, Q.~Huang, Z.~Li, S.~Savarese, M.~Savva, S.~Song, H.~Su \emph{et~al.}, ``Shapenet: An information-rich 3d model repository,'' \emph{arXiv preprint arXiv:1512.03012}, 2015.

\bibitem{dai2017scannet}
A.~Dai, A.~X. Chang, M.~Savva, M.~Halber, T.~Funkhouser, and M.~Nie{\ss}ner, ``Scannet: Richly-annotated 3d reconstructions of indoor scenes,'' in \emph{Proc. IEEE Conf. Comput. Vis. Pattern Recog.}, 2017, pp. 5828--5839.

\bibitem{geiger2012we}
A.~Geiger, P.~Lenz, and R.~Urtasun, ``Are we ready for autonomous driving? the kitti vision benchmark suite,'' in \emph{Proc. IEEE Conf. Comput. Vis. Pattern Recog.}, 2012, pp. 3354--3361.

\bibitem{gu2020weakly}
J.~Gu, W.-C. Ma, S.~Manivasagam, W.~Zeng, Z.~Wang, Y.~Xiong, H.~Su, and R.~Urtasun, ``Weakly-supervised 3d shape completion in the wild,'' in \emph{Proc. European Conf. Comput. Vis.}\hskip 1em plus 0.5em minus 0.4em\relax Springer, 2020, pp. 283--299.

\end{thebibliography}

\end{document}